\documentclass[11pt]{article}

\usepackage[preprint]{acl}

\usepackage{times}
\usepackage{latexsym}

\usepackage[T1]{fontenc}

\usepackage[utf8]{inputenc}

\usepackage{microtype}

\usepackage{inconsolata}

\usepackage{graphicx}
\usepackage{paralist}
\usepackage{enumitem}
\usepackage{subcaption}
\usepackage{color,colortbl}
\usepackage[table]{xcolor}
\usepackage{booktabs}
\usepackage{diagbox}
\usepackage{multirow}
\usepackage{adjustbox}
\usepackage{tabularx}
\usepackage{array}
\usepackage{amsfonts, amssymb, amsmath}
\usepackage{bbm}
\usepackage{cleveref}
\usepackage[normalem]{ulem}
\usepackage[symbol]{footmisc}

\newcommand{\multirowcell}[1]{\begin{tabular}[c]{@{}c@{}}#1\end{tabular}}
\definecolor{predgreen}{HTML}{147A3D}
\definecolor{predred}{HTML}{B3261E}
\newcommand{\correctclaim}[1]{\textcolor{green!50!black}{\uline{#1}}}
\newcommand{\hallucinatedclaim}[1]{\textcolor{red!70!black}{\uline{#1}}}

\title{From Signals to Transfer: A Factorised Study of Probe-Based Uncertainty Estimation in Large Language Models}

\author{
    Ponhvoan Srey\textsuperscript{1} \quad
    Xiaobao Wu\textsuperscript{2}\thanks{Corresponding Authors.} \quad
    Cong-Duy Nguyen\textsuperscript{3} \\
    \textbf{Quang Minh Nguyen\textsuperscript{4}} \quad
    \textbf{Duc Anh Vu\textsuperscript{1}} \quad
    \textbf{Anh Tuan Luu\textsuperscript{1,3}\footnotemark[1]}\\
    \textsuperscript{1}Nanyang Technological University \quad
    \textsuperscript{2}Shanghai Jiao Tong University\\
    \textsuperscript{3}VinUniversity \quad
    \textsuperscript{4}KAIST\\
    \texttt{\{ponhvoan002, vuducanh001, anhtuan.luu\}@ntu.edu.sg}\\
     \texttt{xiaobaowu@sjtu.edu.cn} \quad \texttt{duy.ntc@vinuni.edu.vn} \quad \texttt{qm.nguyen@kaist.ac.kr}
}

\begin{document}
\maketitle
\begin{abstract}

Probe-based uncertainty estimation (UE) has emerged as a prominent approach to detect hallucinations in Large Language Models (LLMs) by learning uncertainty from internal model signals. Yet, recent methods vary simultaneously across feature design, training data construction, and evaluation setting, obscuring what actually drives performance.
To address this issue, we propose a factorised study of probe-based UE under matched conditions.
Our results show that raw hidden states and attention features are difficult to outperform in-domain.
However, under distribution shift, structured and compressed features are more robust, suggesting that in-domain performance alone is insufficient to measure progress.
Furthermore, prompting and label construction significantly affect probe behaviour.
Building on these empirical design principles, we construct benchmark-pretrained probes that retain useful discriminative signal on open-ended factual generation. Their competitive performance relative to limited-label target setting probes and sampling-based estimators suggests that they can provide a practical starting point for subsequent target-domain adaptation.
More broadly, our work encourages deployment-oriented evaluation of probe-based uncertainty estimators\footnote[2]{The code repository is available at \url{https://github.com/ponhvoan/ProbeUE}.}.

\end{abstract}

\section{Introduction}

Hallucination in Large Language Models (LLMs), the tendency to generate fictitious information, remains a persistent barrier to reliable deployment in real-world applications \citep{sahoo-etal-2024-comprehensive, huang-etal-2025-hallucination, zhang-etal-2025-sirens}. This necessitates the development of robust uncertainty estimation (UE) to accurately flag potentially erroneous generations for users \citep{vashurin-etal-2025-benchmarking}. Recent work suggests that probe-based UE, which leverages internal model states, provides some of the most effective signals for hallucination detection \citep{mahaut-etal-2024-factual, tan-etal-2025-consistent}.
This has led to a growing body of work that engineers progressively more structured and informative internal features and integrates them into more sophisticated optimisation protocols \citep{chuang-etal-2024-lookback, he-etal-2024-llm, vazhentsev-etal-2025-token, shelmanov-etal-2025-head}.
However, despite this progress, two questions remain unresolved: what accounts for the gains reported, and whether these gains translate beyond matched benchmark settings.
First, current evaluations conflate multiple design choices, such as training data acquisition, feature representation, supervision, and probe architecture, making it unclear what actually accounts for the observed gains. This motivates our first central research question: \emph{What truly drives performance in probe-based uncertainty  estimation?}

At the same time, a critical bottleneck of probe-based UE is limited generalisability.
Even though probes are highly effective under matched train-test conditions, their performance often degrades when applied to new domains or generation settings \citep{ch-wang-etal-2024-androids, chuang-etal-2024-lookback}.
This limits their utility in real use cases, where uncertainty estimators must handle open-ended generations, rather than only the benchmark format on which they were trained.
Although some prior work evaluates transfer across datasets, such evaluations are restricted to benchmark-to-benchmark transfer \citep{chuang-etal-2024-lookback}, or remain within comparable long-form and claim-level generation setups \citep{han-etal-2025-simple, shelmanov-etal-2025-head}.
In these settings, probes are tested under distribution shift, but the generation format, answer structure, supervision signal, and evaluation protocol remain relatively constrained.
This leaves open whether probes can generalise to less standardised target settings, where outputs are open-ended, vary significantly in length and style, and contain more diverse factual errors.
Such a setting is practically motivated because target-specific probes require costly construction and annotation of in-domain long-form data, whereas benchmark-pretrained probes provide an immediate baseline that can be adapted with target-domain labels.
This formulates our second research question: \emph{Can probes trained under controlled benchmark settings retain useful factuality signal when transferred to open-ended long-form generation tasks?}

To answer these questions, we conduct a controlled study of probe-based UE across three primary dimensions: feature representations, training data construction, and transfer settings.
Our study covers a wide range of recently proposed feature representations, spanning latent embeddings, output probabilities, attention patterns, and their combinations \citep{azaria-mitchell-2023-internal, chuang-etal-2024-lookback, he-etal-2024-llm, huang2025look, shelmanov-etal-2025-head}, evaluated with different probe architectures, supervision sizes, prompting strategies, and automated correctness labels.
We further study benchmark-to-benchmark transfer and a deployment-oriented setting in which probes pretrained on benchmark data are applied to open-ended long-form factual generation.
Together, these allow us to identify which design choices drive in-domain performance, which remain robust under shift, and which design principles support reusable pretrained factuality probes.

Our findings challenge several common assumptions.
First, simple linear probes over raw hidden states and attention features are surprisingly difficult to outperform, even with limited supervision.
Second, data construction choices strongly shape probe behaviour: reasoning-eliciting prompts and lexical matching-based labels can often degrade probe discriminability.
Finally, structured and compressed features offer better trade-offs under distribution shift.
These findings yield a practical recipe: use simple probe architectures, concise generations with semantic correctness labels, and transfer-robust feature representations.
Building on these recommendations, we show that structured benchmark-pretrained probes can remain competitive with limited-label target setting probes and training-free estimators, including expensive sampling-based methods, under open-ended transfer, although performance varies across settings.

Collectively, our work pushes probe-based UE beyond in-domain benchmark comparison toward deployment-oriented practice. Rather than pursuing increasingly complex internal state representation features in isolation, the field should prioritise simple and transferable probe configurations that maintain reliability beyond benchmarks. In summary, our contributions are threefold:
\begin{itemize}[leftmargin=*, itemsep=0pt]
    \item We propose a factorised evaluation framework to disentangle the design factors behind probe-based UE performance.
    \item We identify empirical design recommendations to train lightweight uncertainty probes under different constraints.
    \item We stress test probes pretrained with these practices under benchmark-to-open-ended transfer, characterising when reusable uncertainty signals persist. 
\end{itemize}

\begin{figure*}[!ht]
    \centering
    \begin{subfigure}[b]{0.95\textwidth}
        \includegraphics[width=\linewidth]{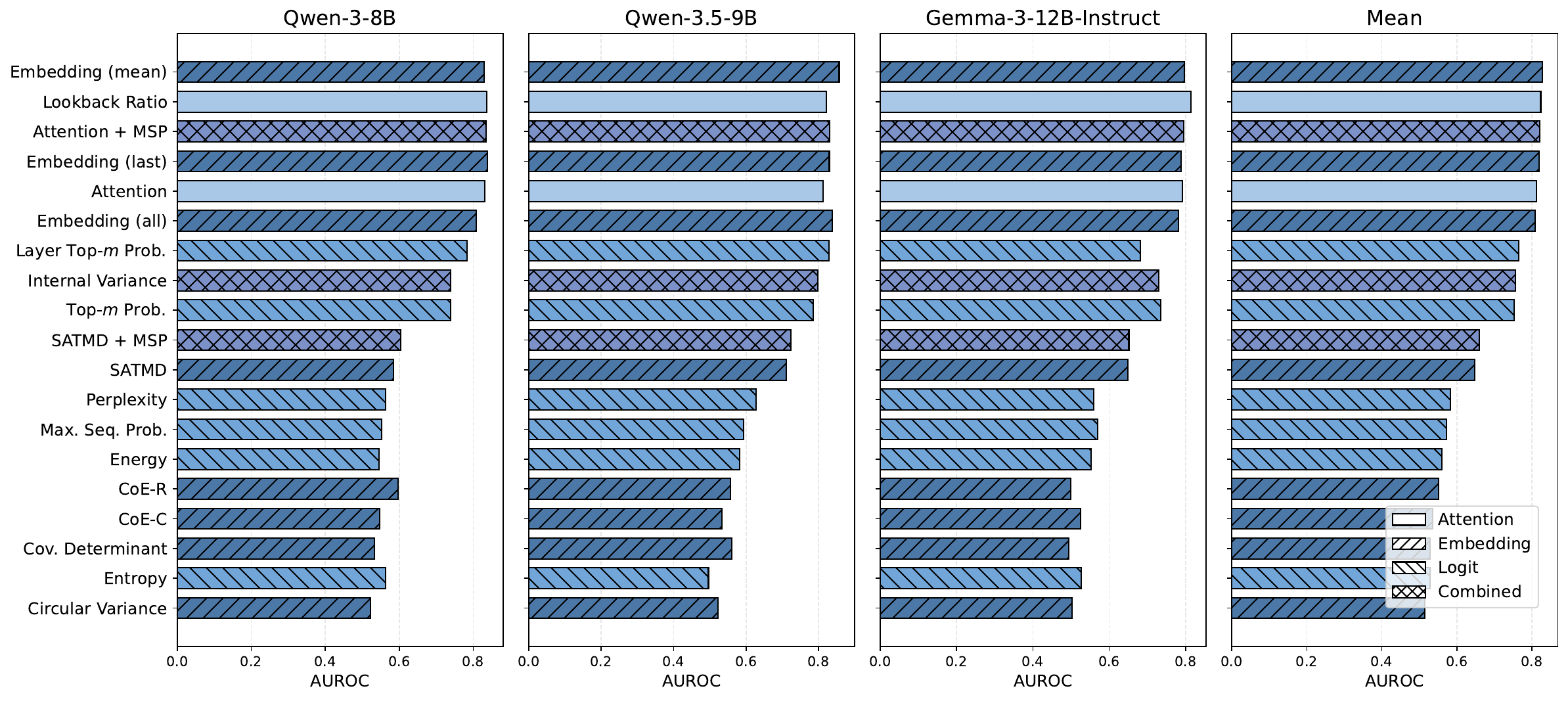}
        \caption{Average AUROC ($\uparrow$)}
    \end{subfigure}

    \begin{subfigure}[b]{0.95\textwidth}
        \includegraphics[width=\linewidth]{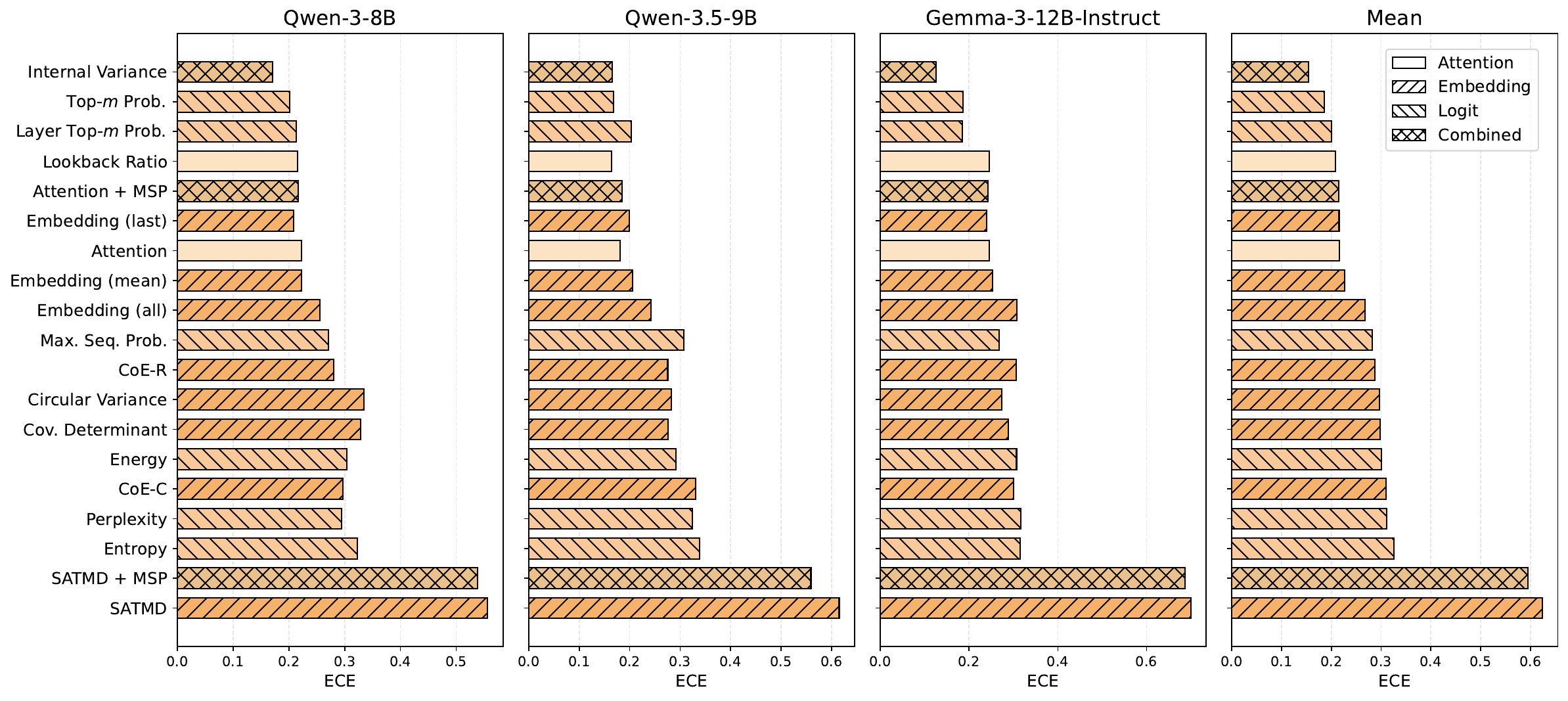}
        \caption{Average ECE ($\downarrow$)}
    \end{subfigure}
    \caption{Main results: In-domain performance averaged over all benchmarks and ranked by mean (rightmost).}
    \label{fig:main_results}
\end{figure*}

\section{Related Work}

In this section, we outline key research threads related to our work. For a more detailed review, please refer to \Cref{sec:extended_related_work}.

\paragraph{Probe-based Uncertainty Estimation (UE)} trains lightweight probes on top of LLM internal signals to predict response correctness or, conversely,  hallucination risk. Early work shows that truth-related information is linearly accessible from final-layer hidden states \citep{azaria-mitchell-2023-internal, burns2022discovering, marks2023geometry}. Subsequent methods use richer representations, including all-layer or token-sequence hidden states \citep{ch-wang-etal-2024-androids, shelmanov-etal-2025-head, zhu-etal-2024-pollmgraph}, hidden-state geometry and density \citep{srey-etal-2026-learning, vazhentsev-etal-2025-token, wang2025latent}, attention patterns \citep{chuang-etal-2024-lookback}, intermediate-layer and output probabilities \citep{huang2025look, he-etal-2024-llm}, and a combination of these signals \citep{he-etal-2024-llm, shelmanov-etal-2025-head}. However, these approaches frequently vary feature representations, data construction, prompting, probe architecture, and evaluation setting simultaneously, obscuring which choices account for their gains.

\paragraph{Benchmarking and Evaluation.}
LM-Polygraph \citep{fadeeva-etal-2023-lm} and a follow-up benchmarking work \citep{vashurin-etal-2025-benchmarking} standardise the comparison of diverse UE methods across tasks, while other work examines robustness to input or dataset shifts \citep{mahaut-etal-2024-factual, wang2025measuring}. \citet{han-etal-2025-simple} establish that linear hidden-states probes can detect factual errors in long-form generation effectively. Similarly, \citet{shelmanov-etal-2025-head} investigated long-form tasks with more sophisticated pretrained uncertainty heads.
Separately, \citet{janiak-etal-2025-illusion} show that lexical correctness metrics can substantially misrepresent hallucination detection performance.
Likewise, \citet{santilli-etal-2025-revisiting} corroborate this finding, and additionally, include supervised final-token probes. We further test how broadly this phenomenon extends across substantially more diverse probe representations.

\paragraph{Positioning of Our Study.} Our contribution is a unifying, probe-specific factorisation under matched conditions. In \Cref{sec:probe_performance}, we jointly vary feature representation, probe architecture, supervision size, prompting, annotation, and transfer setting to identify robust design practices under different constraints.
In \Cref{sec:pretrained}, we then stress test whether these practices support probes pretrained only on short-form benchmarks and deployed off-the-shelf on atomic claims from open-ended long-form generations.
Unlike prior long-form probing work, this setting introduces simultaneous shifts in task, response format, domain, and annotation granularity.
Our work therefore provides a controlled account of what makes probe-based UE work, where it fails, and which design choices better support benchmark-to-open-ended transfer.

\section{What Drives Probe Performance?}
\label{sec:probe_performance}

In this section, we answer our first research question: \emph{What truly drives performance in probe-based uncertainty  estimation?}
To this end, we perform a factorised analysis by varying important design choices, %
namely feature representation, probe architecture, and supervision budget (\Cref{subsec:feature}), and training data and annotation construction (\Cref{subsec:data_construction}). %
We then scrutinise a benchmark-to-benchmark transfer setting (\Cref{subsec:transfer}), where training and test data differ, but they are both derived from the benchmark pool.
We find that 
\begin{inparaenum}[(i)]
\item raw hidden state and attention features are strong in-domain, even with simple linear probes and limited supervision;
\item response elicitation and ground truth annotation strategy choices can substantially affect probe quality; and
\item structured features are more robust under transfer.
\end{inparaenum}

\subsection{Experimental Setup}
\label{subsec:setup}

\paragraph{Datasets.}
We evaluate on seven datasets spanning three response-format groups:
\begin{inparaenum}[(i)]
\item \textbf{Question Answering (QA):} we use TriviaQA \citep{joshi-etal-2017-triviaqa}, SciQ \citep{welbl-etal-2017-crowdsourcing}, and PopQA \citep{mallen-etal-2023-trust} as fact-heavy short-answer QA datasets, encompassing general trivia, science knowledge, and long-tail entity knowledge, respectively;
\item \textbf{Verification}: we include BoolQ \citep{clark-etal-2019-boolq} and StrategyQA \citep{geva-etal-2021-aristotle} as factual and logical verification tasks that often require understanding and implicit reasoning;
\item \textbf{Multiple-Choice Questions (MCQ)}: we use CommonsenseQA \citep[CSQA;][]{talmor-etal-2019-commonsenseqa} and ARC \citep{clark2018think}.
\end{inparaenum}
These datasets cover diverse topics, answer formats, and reasoning demands.
For compactness, we refer to these response-format groups as tasks when describing within- and cross-group transfer below.
We present our prompts in \Cref{subsec:prompts}.
To obtain correctness labels of LLM generations for the QA datasets, we utilise Gemini-3.1-Flash-Lite \citep{google2026gemini31flashlite} as the default LLM-as-a-judge ground truth annotator.
We discuss the effects of different annotation strategies in \Cref{subsec:data_construction}.

\paragraph{Language Models.}
For the main experiment, we evaluate with five popular LLMs across three model families: Llama-3.1-8B \citep{grattafiori2024llama}, Qwen-3-4B, Qwen-3-8B \citep{yang2025qwen3}, Qwen-3.5-9B \citep{qwen3.5_2026}, Gemma-3-12B-Instruct \citep{kamath2025gemma}. We use instruction-tuned versions of all models except for Qwen-3-8B as its official implementation is unavailable.
Llama and Qwen-3-4B are deferred to \Cref{app:supp_expt} for better readability.

\paragraph{Evaluation Metrics.}
To evaluate our results, we report two complementary metrics that capture key desiderata of uncertainty estimation:
\begin{inparaenum}[(i)]
    \item Area Under the Receiver-Operating characteristics Curve \citep[AUROC;][]{davis2006relationship}, which measures discriminability, or the probe's ability to separate correct responses from incorrect ones, with $1.0$ indicating perfect discrimination and $0.5$ no better than chance; and
    \item Expected Calibration Error \citep[ECE;][]{guo2017calibration}, which measures calibration, \textit{i.e.} how closely the predicted probabilities align with real-world outcomes, with $0.0$ indicating perfect calibration. We follow the standard implementation with 10 equal-width bins.
\end{inparaenum}
Given the imbalanced dataset, with considerably fewer positive (hallucinated) instances, we additionally report Area Under the Precision-Recall curve (AUPR) in \Cref{app:supp_expt}.

\paragraph{Feature Representations.}
For each query, we run greedy decoding and extract 19 feature representations from the LLM's internal generation trace, grouped into four information levels. First, we use hidden state representation summaries: \textit{Embedding (last)}, \textit{Embedding (mean)}, and \textit{Embedding (all)} \citep{azaria-mitchell-2023-internal, su-etal-2024-unsupervised}; trajectory-based 
\textit{CoE-R} and \textit{CoE-C} \citep{wang2025latent}; dispersion-based \textit{Circular Variance} and \textit{Cov. Determinant} \citep{srey-etal-2026-learning}; and density-based \textit{SATMD} \citep{vazhentsev-etal-2025-token}.
Second, probability features capture output-side uncertainty: \textit{Max. Seq. Prob. (MSP)}, \textit{Entropy}, \textit{Perplexity} \citep{huang2025look}, \textit{Energy} \citep{liu2020energy}, \textit{Top-$m$ Prob.} \citep{he-etal-2024-llm}, with $m=10$ as default.
Third, we include attention features that reflect attention allocation between input and generated tokens, namely recent-token \textit{Attention} \citep{shelmanov-etal-2025-head}, and \textit{Lookback Ratio} \citep{chuang-etal-2024-lookback}.
Finally, combined features concatenate complementary signals into \textit{Layer Top-$m$ Prob} \citep{he-etal-2024-llm}, \textit{Attention + MSP}, \textit{Internal Variance}, and \textit{SATMD + MSP}.
We provide more details on feature representation in \Cref{app:feat_details}.

\paragraph{Probe Training.}
We train lightweight binary probes on top of each feature representation to predict response correctness. Our default is a linear probe, and we evaluate two non-linear variants (see \Cref{subsec:feature}): an MLP probe with one ReLU-activated hidden layer, and a CNN that applies 1D convolutions over the flattened feature vector before pooling and classification.
We adopt the binary cross-entropy loss, optimised using Adam, with feature normalisation and early stopping based on validation AUROC.
We include training details in \Cref{subsec:probe_training}.

\subsection{Simple Features Are Difficult to Outperform}
\label{subsec:feature}

\begin{figure}
    \centering
    \includegraphics[width=\linewidth]{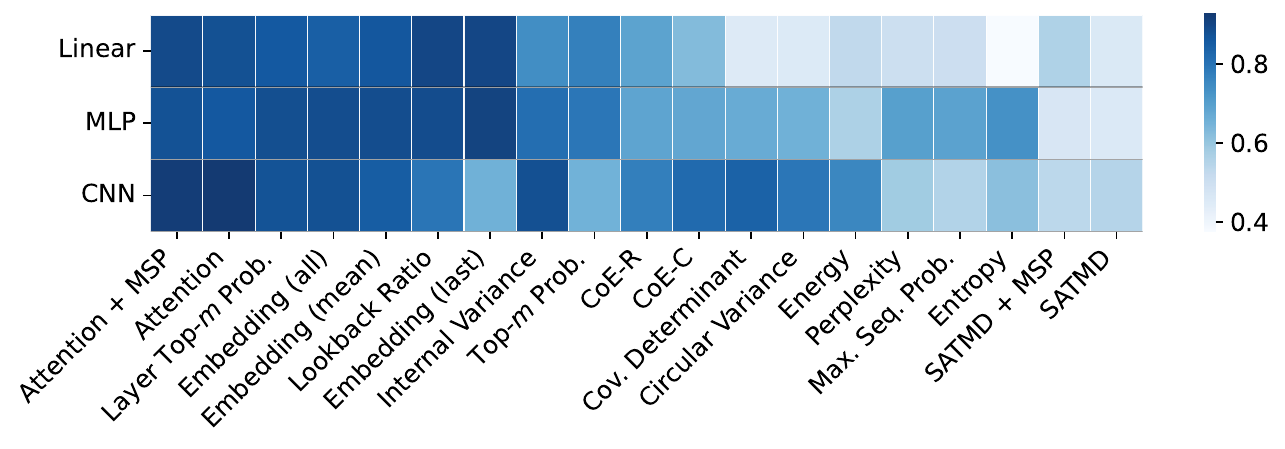}
    \caption{Effect of probe architecture on AUROC.}
    \label{fig:probe_auroc}
\end{figure}

\paragraph{Simple Signals Win.} Under a fixed linear-probe setting, more complex feature engineering does not consistently improve in-domain uncertainty estimation.
As shown in \Cref{fig:main_results,tab:ranking}, simple features, referring to relatively direct representations which do not require fitted density models or substantial geometric transformations, namely Embedding (mean), Embedding (last), Lookback Ratio, and Attention, are consistently competitive across all models, with Lookback Ratio standing out for its smaller input dimension.
Additionally, in line with previous findings, concatenating embeddings from all layers does not always improve performance \citep{chuang-etal-2024-lookback, su-etal-2024-unsupervised}. In contrast, augmenting hidden state features with logit-based signals, such as MSP and Entropy, or vice versa for Top-$m$ Prob., \emph{does} reliably increase discriminability, suggesting that fusing complementary signal types is more useful than simply adding more of the same type of internal features.

\paragraph{Linear Probes Are Stable.} We further test whether stronger probe architectures would affect feature ranking using Qwen-3-8B. \Cref{fig:probe_auroc} shows that MLP and CNN probes can improve weaker feature representations, especially low-dimensional and scalar features, \textit{e.g.} Internal Variance and Energy, but bring limited gains for strong hidden state and attention-based signals.
We observe a similar trend in calibration (\Cref{fig:probe_ece}). However, under the transfer setting (see \Cref{fig:probe_transfer}), higher-capacity probes generally worsen performance, suggesting that they may capture dataset-specific patterns that hurt generalisation. Thus, probe complexity is not uniformly beneficial: once feature representation is sufficiently informative, linear probes are often competitive and more stable.

\begin{figure}[htbp]
    \centering
    \includegraphics[width=\linewidth]{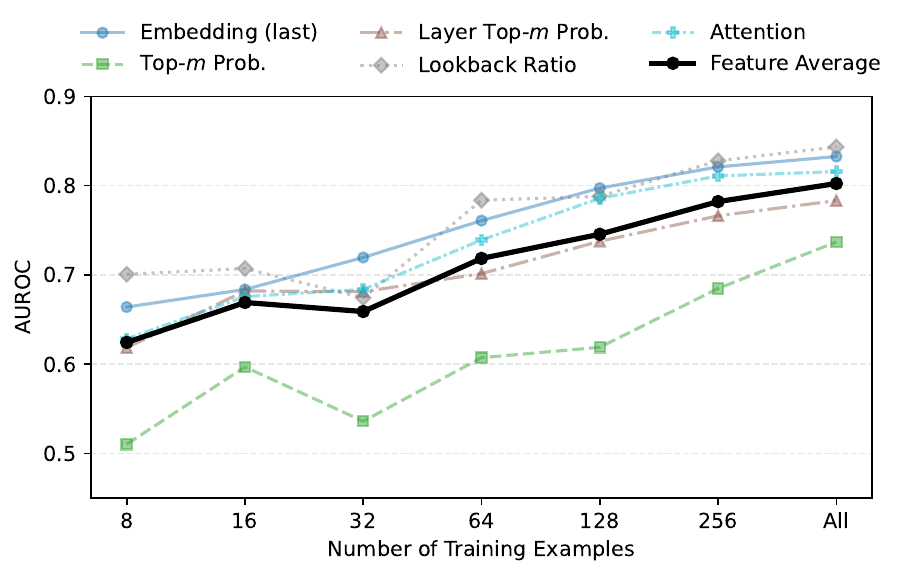}
    \caption{Dependence on number of training examples.}
    \label{fig:trsize}
\end{figure}

\paragraph{Limited Supervision Is Sufficient.} Next, we vary the number of labelled training examples to determine how much supervision is required. \Cref{fig:trsize} visualises AUROC for a strong set of Qwen-3-8B features. Performance steadily improves with more labelled data, but satisfactory performance is recovered with relatively few labels, with average AUROC %
reaching a useful threshold at around 128-256 examples.
Informative representations like Embeddings (last) perform well with few examples and gains beyond 128--256 examples are modest.

Overall, these results suggest that factuality information is largely linearly accessible in the hidden states even with limited supervision.
The main benefit of recent feature engineering may therefore not be stronger in-domain discrimination, but better performance under other constraints, such as calibration and transfer. For instance, ECE results in \Cref{fig:main_results} show that internal variance is relatively well-calibrated despite lower AUROC.
This motivates our further examination in transfer tasks rather than treating in-domain AUROC as the sole indicator of progress.

\subsection{Data Construction Strongly Shapes Performance}
\label{subsec:data_construction}

To obtain training data, there are two crucial design choices: prompting, which directly affects the internal states, and the ground truth annotation strategy.

\begin{figure}
    \centering
    \includegraphics[width=\linewidth]{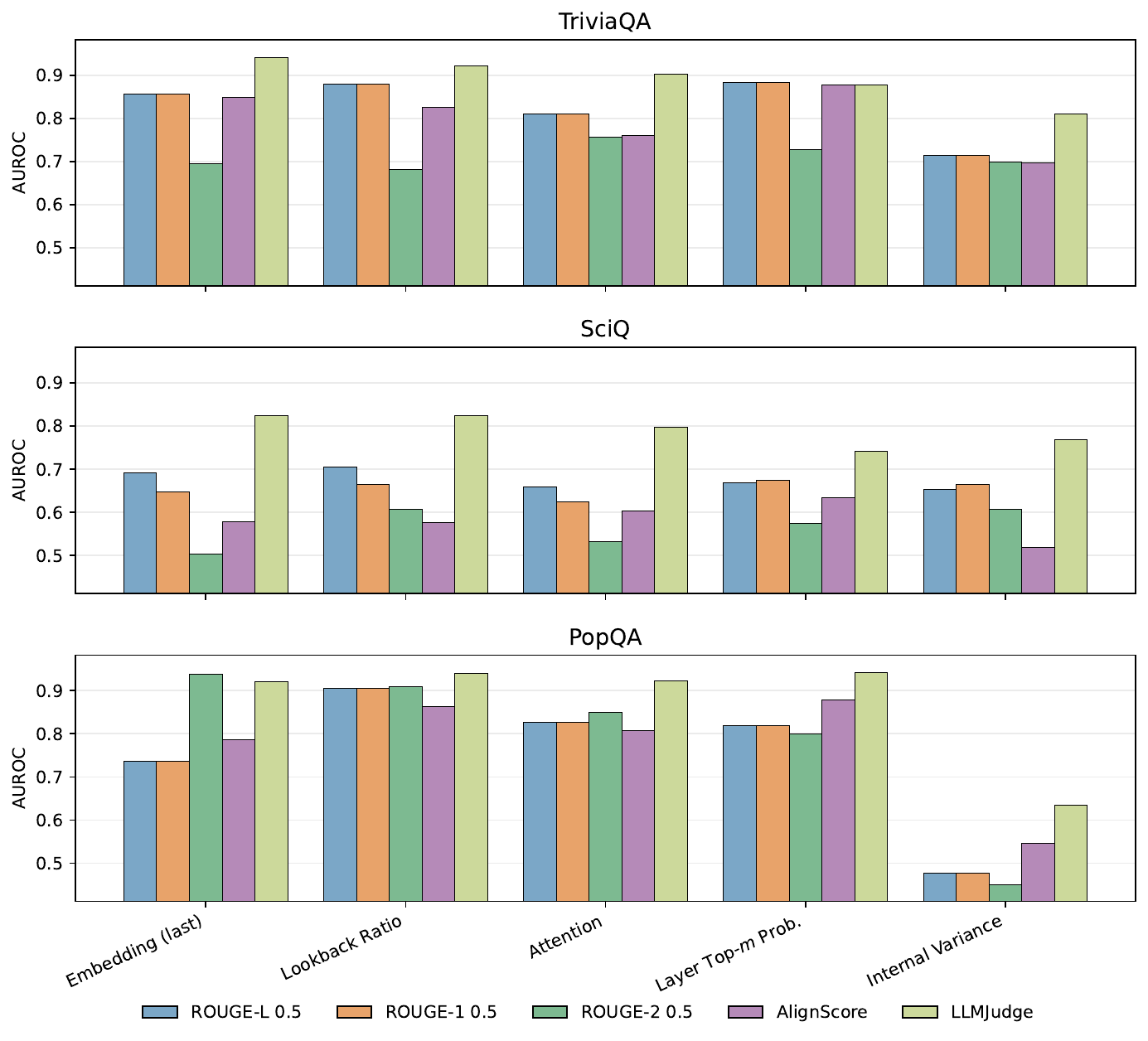}
    \caption{Impact of various automated annotation options on probe performance.}
    \label{fig:labels}
\end{figure}

\begin{figure*}[ht]
    \centering
    \includegraphics[width=\linewidth]{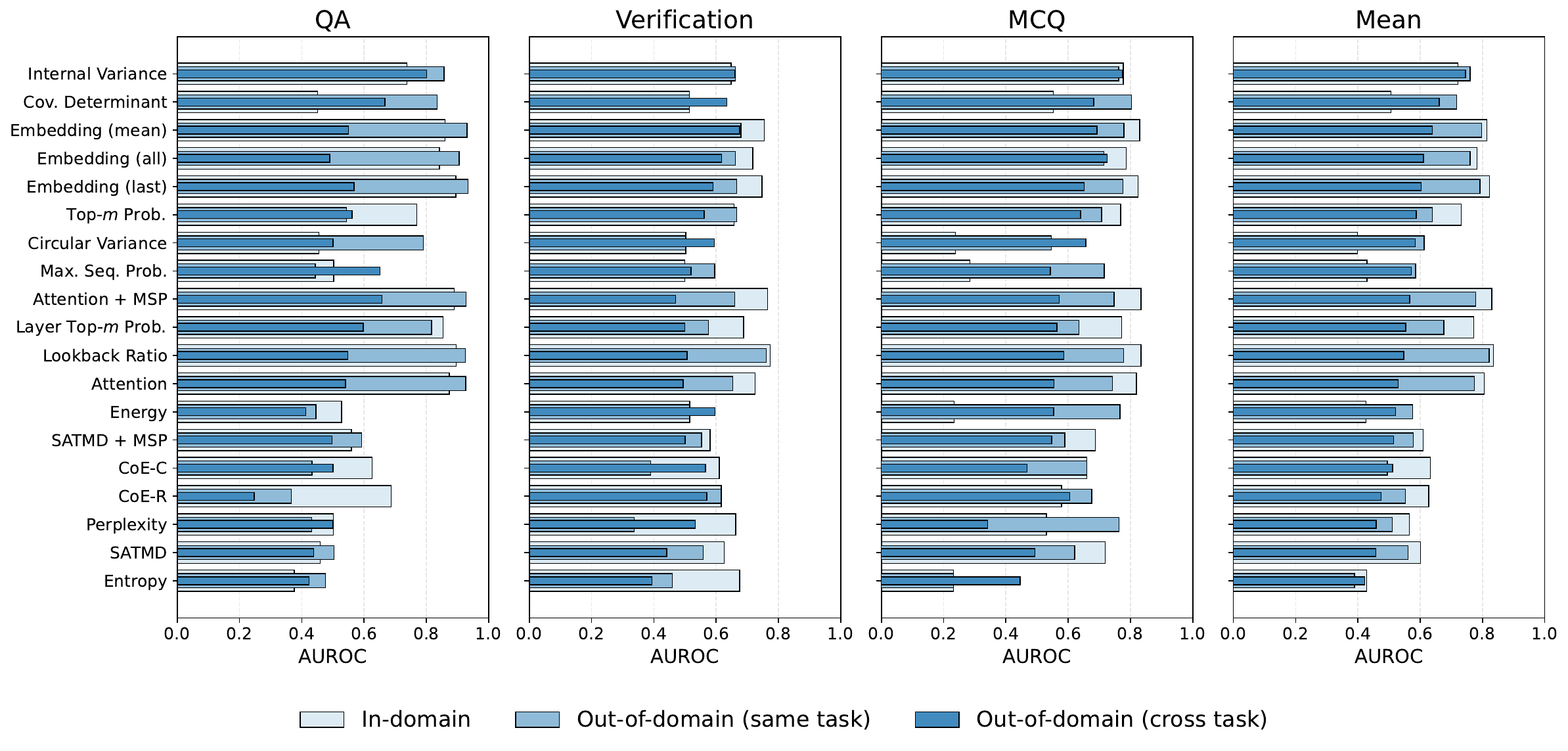}
    \caption{Benchmark-transfer performance. Average AUROC across In-domain, Out-of-domain (same task), and Out-of-domain (cross task) configurations. Methods are ranked by out-of-domain (cross task) performance.}
    \label{fig:task_transfer_auroc}
\end{figure*}

\paragraph{Reasoning-Eliciting Prompts Can Reduce Probe Discriminability.}
We compare no-, medium-, and long-reasoning prompts (\Cref{tab:reasoning_suffix}) on ARC and CSQA using Qwen-3-8B (\Cref{fig:prompt}) and Gemma-3-12B-Instruct (\Cref{fig:prompt_gemma}).
Although reasoning elicitation does not reduce answer accuracy, it can substantially alter probe discriminability. Across both models, the constrained medium-reasoning condition is generally weakest, whereas long reasoning partially restores performance and occasionally matches or exceeds direct answering. We hypothesise that one-sentence explanations introduce relatively stereotyped tokens without adding much correctness-specific information, whereas longer traces contain richer task-relevant content. However, longer traces also shift the answer position and, for sequence-aggregated features, may dilute relevant information, potentially explaining why they may often remain below direct answering. 
We replicate these experiments on StrategyQA and MATH-500 \citep{lightman2024let} (\Cref{fig:prompt_extra}). We observe similar prompt sensitivity trends except when reasoning improves both answer accuracy and probe discriminability in tandem (Gemma on MATH-500).
We present more reasoning-control analysis in \Cref{app:supp_expt}.
Therefore, we recommend direct answering as a stable and computationally efficient default, while emphasising that the effect is feature- and model-dependent.

\paragraph{Ground Truth Annotation Matters.} We study how the choice of correctness annotation affects probe evaluation.
Because obtaining  human annotations at scale is costly, we resort to automated scorers, a common practice in the field \citep{duan-etal-2024-shifting, janiak-etal-2025-illusion, kuhn2023semantic, vazhentsev-etal-2025-token}.
Specifically, we examine three types of labelling strategies on TriviaQA, SciQ, and PopQA with Qwen-3-8B: Rouge \citep{lin-2004-rouge}, AlignScore \citep{zha-etal-2023-alignscore}, and LLM-as-a-judge with Gemini-3.1-Flash-Lite.
We present the detailed validation procedure in \Cref{par:llm_judge_val}.
In \Cref{fig:labels}, we investigate the effect of varying the labelling choice for the training data.
We find that LLM-as-a-judge labels consistently yield the highest AUROC.
These results caution against relying on lexical metrics to obtain ground truth labels, and instead advocate for LLM-as-a-judge labelling, which better captures semantic correctness.

\subsection{Structured Features Are More Robust Under Transfer}
\label{subsec:transfer}

For benchmark transfer, we report average performance for Qwen-3-8B (\Cref{fig:task_transfer_auroc}) and Gemma-3-12B-Instruct (\Cref{fig:task_transfer_auroc_gemma}) across the three groups: QA, Verification, and MCQ. Further, we distinguish between in-domain, out-of-domain (OOD) (same task), and OOD (cross task) performance. By OOD (same task), we mean testing on a different dataset from the same task, \textit{e.g.} TriviaQA $\rightarrow$ SciQ, and OOD (cross task) refers to all remaining OOD train-test configurations.

On average, there is a clear gap between in-domain and transfer performance, especially across tasks. This confirms that probes are brittle, and strong in-domain performance alone is not sufficient evidence they will be useful in deployment.
The in-domain advantage of raw hidden state and attention features, narrows considerably as they may encode task-specific biases that inhibit generalisation.
In comparison, several structured probability and dispersion features, such as Internal Variance and Top-$m$ Prob., retain more of their performance under same task and cross task transfer. As a practical takeaway, pretrained probes should not be selected solely by in-domain performance, but also, by robustness to dataset shift.

\section{Stress-Testing Benchmark-Pretrained Probes on Open-Ended Generation}
\label{sec:pretrained}

In this section, we answer our second research question: \emph{Can probes trained under controlled benchmark settings retain useful factuality signal when transferred to open-ended generation tasks?}
We compare benchmark-pretrained probes against feature-matched probes trained with limited target setting supervision, as well as training-free and sampling baselines.
We find that structured benchmark-pretrained probes retain useful factuality signal under this substantial distribution shift, motivating their use as an immediate baseline when target-domain labels are unavailable.

\subsection{Experimental Setup}

\paragraph{Evaluation Data and Models.}
We sample 100 entities from the dataset used in \citep{min-etal-2023-factscore}, and 50 from each domain in \citep{shelmanov-etal-2025-head}. In total, we have 500 entities spanning nine domains: \textit{Biographies (in-domain; ID)}, \textit{Biographies (OOD)}, \textit{Artworks}, \textit{Books}, \textit{Cities}, \textit{Events}, \textit{Inventions}, \textit{Landmarks}, \textit{Movies}.
Using Qwen-3-8B, Gemma-3-12B-Instruct, and Llama-3.1-8B, we use the same simple prompts as \citet{shelmanov-etal-2025-head} to generate texts for each entity up to 512 tokens.
Similar to \citet{han-etal-2025-simple}, we utilise Gemini-3.1-Flash-Lite to automatically decompose the long-form continuation into atomic factual claims (see \Cref{subsec:prompts} for complete instructions).
For each claim, we obtain the features and labels as in \Cref{sec:probe_performance}, but use GPT-5.4-Mini \citep{gpt5_4} for factuality annotation as it shows substantially higher agreement with human judgements on our validation subset (\Cref{tab:labels_llm}).
Thus, benchmark-pretrained probes additionally face this supervision-source shift as part of the intended benchmark-to-open-ended transfer setting.
This produces \textasciitilde 500--1,000 claim-label pairs per domain; we report the detailed statistics in \Cref{tab:section4_dataset_statistics}.

\begin{figure*}
    \centering
    \includegraphics[width=\linewidth]{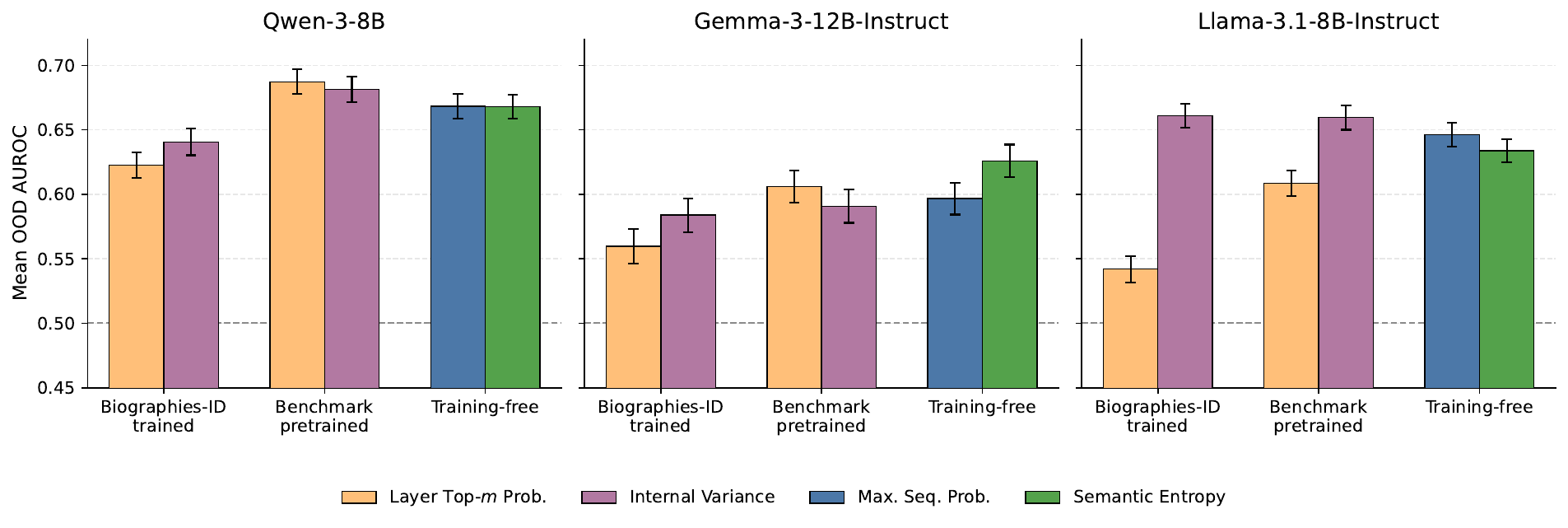}
    \caption{Mean OOD across the eight held-out domains. We compare representative Biographies-ID-trained, benchmark pretrained, and training-free estimators. Benchmark-pretrained structured probes are competitive overall, but their relative performance varies across models; we report full domain results in \Cref{tab:pretrained,tab:pretrained_gemma,tab:pretrained_llama}.}
    \label{fig:sec4_ood_comparison}
\end{figure*}

\paragraph{Pretrained Probes.}
We pretrain linear probes on features extracted from the pooled seven-benchmark dataset. Guided by findings in \Cref{sec:probe_performance}, we use direct-answer generation, LLM-judge labels, and representative transfer-oriented features: Lookback Ratio, Layer Top-$m$ Prob., and Internal Variance, retaining Embedding (last) as a comparison.
Furthermore, we consider two simple ensembles:
Ensemble (probe), which averages from three Internal variance probes trained separately on QA, Verification, and MCQ; and
Ensemble (task), which averages predictions from probes trained on pooled data using the three features.

\paragraph{Biographies-ID-Trained and Training-Free Baselines.}
To isolate the effect of benchmark pretraining from feature choice, we compare benchmark-pretrained probes against feature-matched linear probes trained on 256 claim--label pairs from Biographies (ID).
We refer to these as \emph{Biographies-ID-trained} probes.
For Qwen-3-8B, we additionally report Embedding (last) probes under varying label budgets and probe architectures. We also include training-free logits- (MSP, Entropy, Perplexity, Energy) and sampling-based baselines (Semantic Entropy \citep{kuhn2023semantic}, Eigenscore \citep{chen2024inside}) with five samples per query.

\paragraph{Surrogate-Model Features.}
We additionally test whether probes require the generator's own internal states. We pass generated claims through an accessible surrogate model and evaluate its pretrained probes. We report full results in \Cref{par:surrogate}.

\subsection{Results and Discussion}

\paragraph{Benchmark-Pretrained Probes Retain Useful Zero-Label Signal.}
\Cref{fig:sec4_ood_comparison} summarises the mean OOD results across models.
Structured benchmark-pretrained probes retain useful discriminative signal despite receiving no labels from the open-ended target domains.
Aggregated across OOD domains, the best benchmark-pretrained and training-free methods are broadly competitive, with no method class dominating consistently across models and domains.
Importantly, the benchmark-pretrained probes require only a single greedy generation trace, while sampling-based baselines use five samples per query. We therefore interpret the supported result as a practical and computationally efficient zero-target-label baseline. 

\paragraph{Performance Varies Across Models and Domains.} 
The aggregate picture masks considerable heterogeneity. \Cref{tab:pretrained,tab:pretrained_gemma,tab:pretrained_llama} provide full domain breakdown.
On Qwen, structured benchmark-pretrained probes are among the strongest methods on average, whereas on Gemma and Llama, training-free estimators are often equally strong or stronger. The relative ordering also varies across domains: a feature that transfers well to one domain can approach chance on another.
Furthermore, performance still remains modest with several feature--domain combinations approaching chance, indicating that transferable factuality signal can weaken or vanish under severe shift. Pretrained probes should therefore be viewed as reusable baselines rather than universally superior, deployment-ready detectors.

\paragraph{Toward Reusable Uncertainty Probes.}
Taken together, these results support a modular view of probe-based UE: a broadly trained uncertainty head can offer a reasonable and immediate starting baseline.
Qualitative examples in \Cref{tab:qualitative_examples} illustrate this potential: pretrained structured probes successfully flag several hallucinated claims and assign high confidence to correct generations, while the included failure cases show that confident errors remain. %
A natural next step is to adapt such probes with limited target-domain supervision as deployment data become available.
The surrogate-model results further suggest that the uncertainty module need not share the generator's architecture, opening a path toward factuality estimation for black-box LLMs.%

\section{Conclusion}

In this work, we examine the influence of feature representation, training data construction, and transfer on probe-based UE. Our results indicate that raw features are surprisingly strong in-domain. However, under distribution shift, structured features are more robust, suggesting that in-domain discriminability alone is insufficient for assessing progress.
Based on these findings, we distil practical design recommendations to train probes on benchmark data. Even without task-specific supervision, these pretrained probes demonstrate reasonable performance when transferred to long-form open-ended factual generation, providing a useful off-the-shelf baseline. We hope this work encourages the field to move beyond matched benchmark comparison and towards probe configurations that remain useful under realistic distribution shifts. Future work can build on this direction by adapting pretrained probes, enabling stronger performance in deployment.

\section*{Limitations}

Although our work has derived design recommendations to train uncertainty/factuality probes and has shown how these can support deployment-oriented open-ended generation, it has the following limitations:

\paragraph{Constrained Open-Ended Generation.}
We simulate a more realistic setting by considering long-form generation involving an entity. However, there are many other uncovered use cases, such as unconstrained dialogue and multi-turn interaction. Thus, our conclusions about pretrained probe transfer should be interpreted as evidence for entity-centric factual generation, rather than a universal claim of robustness across all open-ended settings.

\paragraph{Model and Feature Coverage.}
Our study covers a broad set of LLMs, probe features, and design factors, but it is not exhaustive. We focus on popular open-weight LLMs and internal signals.
In particular, we conduct our detailed analysis with base Qwen-3-8B and Gemma-3-12B-Instruct.
Future work can assess how broadly the observed best practices apply to larger instruction-tuned LLMs and other feature representations.

\paragraph{Hallucination Mitigation.}
We focus on hallucination detection using probes. We leave probe-guided hallucination mitigation to future work. They can investigate whether the probe configurations identified here also support effective mitigation strategies.

\paragraph{Transfer Direction.}
We study transfer from short-form QA benchmarks to open-ended factual
generation. This reflects our intended practical setting, where benchmark supervision is readily available, but target annotation can be costly. The reverse direction has not been studied and can additional limitations of supervised methods.

\paragraph{LLM-as-a-judge Labels.} In our experiments, we regard LLM annotations as ground truth labels. Human validation across multiple benchmark and open-ended domains show high, \emph{but imperfect}, agreement with these annotations. Reported performance should therefore be interpreted as agreement with the chosen annotation protocol.

\bibliography{custom}

\clearpage

\appendix

\section{Extended Related Work}
\label{sec:extended_related_work}

Here, we provide additional background on related work that we omit from the main text for clearer presentation.

\subsection{Probe-based Uncertainty Estimation} 

Probe-based UE leverages LLM internal states to predict whether generated responses are correct or factual, with the predicted probability reflecting model confidence, or inversely, uncertainty.
This paradigm is attractive because it typically requires only a single LLM forward pass, unlike expensive sampling-based methods, and often achieves strong in-domain performance.
Early work showed that truth-related information can be extracted from hidden activations, often from the final layer and final token, using simple linear classifiers \citep{azaria-mitchell-2023-internal, burns2022discovering, marks2023geometry}. %
Subsequent methods extend this paradigm by deriving more informative hidden state representations in several ways.
Some works steer the generation procedure via prompting to elicit responses, and hence internal states, that are more discriminative for factual verification \citep{zhang-etal-2025-prompt, srey-etal-2025-unsupervised}.

Later work exposes the probe to richer hidden-state information, for example by pooling hidden states across all layers \citep{ch-wang-etal-2024-androids}, or by modelling hidden states from all generated tokens as sequential inputs \citep{shelmanov-etal-2025-head, srey-etal-2026-learning, zhu-etal-2024-pollmgraph}, or by integrating cross-model hidden states \citep{tan-etal-2025-consistent}.
Another approach transforms hidden states into structured summaries intended to capture uncertainty- and hallucination-relevant geometry, such as density-based features across layers \citep{vazhentsev-etal-2025-token}, or cross-layer dynamics \citep{srey-etal-2026-learning}.

A parallel line of work explores internal signals beyond hidden states alone. Attention-based methods, for example, use patterns such as the lookback ratio \citep{chuang-etal-2024-lookback}, the relative attention paid to source context compared to generated tokens.
Related works also incorporate probability-space information, such as token probabilities \citep{vazhentsev-etal-2025-token}, entropy \citep{srey-etal-2026-learning}, or logit-derived features, such as top-$k$ output indices \citep{he-etal-2024-llm}.
These recent hybrid methods combine hidden states, attention maps, and probability-based signals through direct concatenation \citep{shelmanov-etal-2025-head, srey-etal-2025-unsupervised, vazhentsev-etal-2025-token}, or through specialised submodules \citep{he-etal-2024-llm}.
Related calibration methods learn post-hoc mappings from heuristic UE scores to correctness-aligned estimates with a model-specific corrector \citep{li-etal-2025-towards, srey2026towards}, further extending internal-state probes.

Overall, these methods show that diverse information relevant to factuality is available within the generation trace.
As a result, probe-based UE has increasingly been framed as a search for richer features, better calibrated signals, and higher-capacity optimisation pipelines.
However, these methods often vary simultaneously in many design factors, making it unclear which choices actually account for the observed gains.
Our work clarifies when such engineering is necessary, when simple hidden state probes suffice, and which design choices remain robust under transfer.

\subsection{Benchmarking and Transfer Robustness}
LM-Polygraph \cite{fadeeva-etal-2023-lm} provides an extensive toolkit for comparing uncertainty estimators, while \citet{vashurin-etal-2025-benchmarking} benchmark a broad collection of UE and confidence-normalisation methods across numerous generation tasks. UQLM \citep{bouchard2026uqlm} similarly supplies off-the-shelf black-box, white-box, LLM-judge, and ensemble scorers. Other evaluation studies examine robustness to semantically equivalent prompts \citep{mahaut-etal-2024-factual}, in-domain and OOD question answering \citep{wang2025measuring}, uncertainty-aware LLM benchmarking \citep{ye-2024-benchmarking}, and real-time entity-level hallucination detection \citep{obeso2025real}. These efforts establish the importance of consistent multi-task evaluation, although they do not systematically factorise the training pipeline of supervised internal-state probes. LM-Polygraph's central contribution is broad, controllable UE benchmarking rather than a probe-specific analysis of training-test transfer.

Probe transfer has been considered in several forms. For example, \citet{ch-wang-etal-2024-androids} and \citet{chuang-etal-2024-lookback} evaluate transfer between benchmark datasets. More recently, this extends to long-form generations. \citet{han-etal-2025-simple} demonstrate that lightweight final-state probes can detect factual errors in long-form generations, and generalise to outputs from other models and generations of other entities.
Similarly, \citet{shelmanov-etal-2025-head} train reusable uncertainty heads and study their application when deployed to generations in other domains. These works motivate reusable probes but within relatively constrained settings, while our experiments specifically consider probes trained on short-form benchmark responses and transferred to claim-level open-ended generation. \citet{han-etal-2025-simple}'s principal finding is therefore complementary to our focus on factorised design and transfer across generation formats.

\subsection{Correctness Annotation and Prompting}

The evaluation of UE requires a correctness annotation protocol or function against which uncertainty scores are assessed. Lexical overlap metrics such as ROUGE \citep{lin-2004-rouge} are inexpensive but may not capture semantically correct paraphrases and are threshold-sensitive. \citet{janiak-etal-2025-illusion} show that ROUGE can align poorly with human factuality judgements and can substantially misguide conclusions on UE performance. \citet{santilli-etal-2025-revisiting} analyse how shared biases, namely response-length bias, between correctness functions and UE scores can distort AUROC rankings.
Their evaluation covers lexical, embedding-based, and LLM-judge correctness functions as well as learned final-token probes trained with ROUGE-L or LLM-judge supervision. They find LLM-judge correctness functions to be comparatively well aligned with human judgements and less susceptible to length-related confounding.

Our annotation study builds on these findings rather than claiming the general observation as new. We test whether the effect extends across hidden-state, attention, probability, dispersion, and hybrid probe representations while holding the evaluation labels fixed. This places annotation choice within the broader probe-training pipeline and allows its interaction with feature design to be examined under matched conditions.

Prompting can likewise alter both model confidence and internal representations used by uncertainty estimators. Prior work finds that reasoning can improve confidence expressed in the output token space in some settings \citep{yoon2026reasoning}, but not consistently \citep{mei-etal-2026-reasoning}, and may inflate probability-based confidence on multiple-choice tasks \citep{fu2025multiple}. Prompt-guided probe methods also intentionally elicit representations that are more separable for factuality prediction \citep{zhang-etal-2025-prompt}.
These findings motivate our controlled comparison of direct, concise-reasoning, and long-reasoning response formats.

\paragraph{Our novelty.}
While our work recovers several prior observations, our controlled comparison yields two novel insights. First, in-domain superiority and transfer robustness favour different feature classes, encouraging researchers to look beyond in-domain benchmark evaluation. Second, probes pretrained only on short-form benchmarks, adopting our findings, can retain meaningful factuality signals when transferred to open-ended generation, providing a useful zero-target-label baseline for when collecting annotated long-form data is costly. 
Collectively, these findings identify transfer robustness as a distinct design objective and demonstrate a practical path towards reusable uncertainty probes beyond matched benchmarks.

\section{Experimental Setup Details}

\subsection{Feature Representations}
\label{app:feat_details}

\Cref{tab:feat_details} summarises the feature representations used in our factorised study.
We follow the original formulation proposed in the respective papers and apply sequence mean aggregation as needed.
In particular, for embedding- and logit-based UE scores, we apply TAC \citep{srey2026towards} to better align raw confidence with truthfulness, and we adapt Internal Variance to a sequence-level internal-dispersion feature. 

\paragraph{Feature and Probe Coverage.} We compare representative features under a tractable design. On top of concatenation with MSP, as is standard in prior work \citep{vazhentsev-etal-2025-token}, the evaluated hybrid combinations include: Internal Variance combines with Entropy; Layer Top-m Probability projects intermediate-layer representations through the output head, combining embeddings with logits. Other combinations are worthwhile, but evaluating every pair would be infeasible and obscure the factorised comparison. 
Our evaluation also includes layer-wise information: Embedding (all) retains a representation from every layer, while Layer Top-\(m\) Probability and several structured features capture layer-wise trajectories. The study does not exhaustively cover alternative token- and layer-wise representations.
Subsequently, we limit the probe architectures to simple classifiers as these features were primarily designed as fixed-length, sequence-level features. Adapting them to variable-length sequences would require redefining their aggregation, introducing padding and masking choices, and substantially increasing storage and training costs. A systematic evaluation of sequential probes, such as Transformers, is valuable but constitutes a separate design axis beyond the current scope.

\begin{table*}[!ht]
\centering
\resizebox{\textwidth}{!}{
\begin{tabular}{c|c|l|c}
\toprule
\textbf{Type} & \textbf{Name} & \multicolumn{1}{c}{\textbf{Description}} & \textbf{Citation} \\
\midrule
\multirow{6}{*}{Hidden State} & \multicolumn{1}{c|}{Embedding (last)} & Last generated token, last layer embedding & \citep{azaria-mitchell-2023-internal} \\
& \multicolumn{1}{c|}{Embedding (mean)} & Mean embedding across generated tokens and layers & \citep{su-etal-2024-unsupervised} \\
& \multicolumn{1}{c|}{Embedding (all)} &  Token-mean embeddings across all layers & \citep{su-etal-2024-unsupervised} \\
& \multicolumn{1}{c|}{CoE-R, CoE-C} & Geometry of hidden state trajectory across successive layers & \citep{wang2025latent} \\
& \multicolumn{1}{c|}{Circular Variance, Cov. Determinant} & Cross-layer internal dispersion & \citep{srey-etal-2026-learning} \\
& \multicolumn{1}{c|}{SATMD} & Layer-wise token deviations from the distribution of correct generations & \citep{vazhentsev-etal-2025-token} \\
\midrule
\multirow{2}{*}{Logit} & \multicolumn{1}{c|}{MSP, Entropy, Perplexity, Energy} & Logit-derived uncertainty score & \citep{huang2025look, liu2020energy} \\
& \multicolumn{1}{c|}{Top-$m$} & $m$-highest predicted token probabilities. & \citep{he-etal-2024-llm} \\
\midrule
\multirow{2}{*}{Attention} & \multicolumn{1}{c|}{Attention} & Attention maps to previous tokens. & \citep{shelmanov-etal-2025-head} \\
& \multicolumn{1}{c|}{Lookback Ratio} & Relative attention paid to input context versus generated tokens. & \citep{chuang-etal-2024-lookback} \\
\midrule
\multirow{2}{*}{Combined} & \multicolumn{1}{c|}{Layer Top-$m$ Prob.} & Concatenation of Top-$m$ Prob. from all layers. & \citep{he-etal-2024-llm} \\
& \multicolumn{1}{c|}{Internal Variance, Attention + MSP, SATMD + MSP} & Combination via concatenation. & --- \\
\bottomrule
\end{tabular}
}
\caption{Feature representations used in our study.}
\label{tab:feat_details}
\end{table*}

\subsection{Feature Formulations}
\label{app:feature_formulations}

Here, we provide brief formulae of the feature representations.
More detailed formulations can be found in the respective papers.

\paragraph{Notation.}
Let $x = (x_1, \ldots, x_N)$ denote the input query and $y=(y_1, \ldots, y_T)$ the response.
The LM has $L$ transformer layers, hidden dimension $d$, $Q$ attention heads per layer, and vocabulary $\mathcal{V}$.
We write $h_{l,t} \in \mathbb{R}^d$ for the hidden state of generated token $t$ at layer $l\in\{0, \ldots, L\}$, including the embedding output at $l=0$. We define sequence-mean embedding, LM predicted probability, and logit, respectively, as
\begin{align*}
    \bar{h}_l &= \frac{1}{T} \sum_{t=1}^T h_{l,t}, \\
    p_t &= \operatorname{softmax}(z_t), \\
    z_t &= W_u h_{L,t} + b_U,
\end{align*}
where $W_U$ is the LM output head. Let $\operatorname{Top}_m(p) \in \mathbb{R}^{m}$ return the $m$ largest entries of $p$ in descending order, and let $\bigoplus$ denote the concatenation operator. Unless otherwise stated, token-level quantities are averaged over the generated response. We use $m=10$, energy temperature $\tau=0.7$, attention-window size $k=2$, and covariance regularisation $\eta=10^{-3}$.

\paragraph{Hidden-state embeddings.}
We evaluate three embedding aggregations: Embedding (last), Embedding (mean), and Embedding (all), which are defined, respectively, as
\begin{align*}
    f_{\text{last}} &= h_{L,T}, \\
    f_{\text{mean}} &= \frac{1}{L+1}\sum_{l=0}^L \bar{h}_l,\\
    f_{\text{all}} & = \bigoplus_{l=0}^{L} \bar{h}_l
\end{align*}
Thus, Embedding (last) uses the final token and final layer, Embedding (mean) averages across tokens and layers, and Embedding (all) averages across tokens while retaining the embedding from every layer.

\paragraph{Cross-layer evolution.}
Let $c(a,b)=(a^\top b)/(\|a\|_2\|b\|_2)$ , and for $l=1,\ldots,L$, define the successive-layer normalised angular and magnitude changes
\begin{equation*}
    a_l = \frac{\arccos\left(c(\bar{h}_l, \bar{h}_{l-1})\right)}{\arccos\left(c(\bar{h}_L, \bar{h}_{0})\right)}, \;
    d_l = \frac{\|\bar{h}_l - \bar{h}_{l-1}\|_2}{\|\bar{h}_L - \bar{h}_{0}\|_2},
\end{equation*}
The two cross-layer evolution summaries, CoE-R and CoE-C, are
\begin{align*}
    &f_{\text{CoE-R}} = \frac{1}{L}\sum_{l=1}^L (d_l-a_l) \\
    &f_{\text{CoE-C}} = \\ &\sqrt{\left(\frac{1}{L}\sum_{l=1}^L d_l \cos a_l\right)^2 + \left(\frac{1}{L}\sum_{l=1}^L d_l \sin a_l\right)^2}
\end{align*}

\paragraph{Cross-layer dispersion.}
Circular Variance measures directional disagreement among layer representations
\begin{equation*}
    f_{\text{CircVar}} = 1 - \left\lVert \frac{1}{L+1} \sum_{l=0}^{L} \frac{\bar{h}_l}{\|\bar{h}_l\|_2} \right\rVert_2
\end{equation*}
For Cov. Determinant, let $H\in \mathbb{R}^{(L+1)\times d}$ contain $\bar{h}_l^\top$ as its rows, and centre each row across its hidden dimensions
\begin{equation*}
    \tilde{H}_{l,r} = H_{l,r} - \frac{1}{d}\sum_{s=1}^{d} H_{l,s}
\end{equation*}
We compute the regularised layer covariance
\begin{equation*}
    C = \frac{\tilde{H}\tilde{H}^\top}{d-1} + \eta I
\end{equation*}
with $I$ being the identity matrix and $\eta=10^{-3}$ , and use its normalised log-determinant
\begin{equation*}
    f_{\text{CovDet}} = \frac{1}{L+1}\operatorname{logdet}C = \frac{1}{L+1}\sum_{j=1}^{L+1} \log \lambda_{j}(C),
\end{equation*}
where $\lambda_j$ is the $j$-th eigenvalue.

\paragraph{Density-based feature.}
SATMD fits a Gaussian reference distribution at each layer using token representations from correct training responses. Let $\mu_{l}$ and $\Sigma_l$ be the corresponding layer-wise mean and regularised covariance. The Mahalanobis distance of token $t$ is
\begin{equation*}
    D_{l,t} = \sqrt{(h_{l,t}-\mu_l)^\top \Sigma_l^{-1}(h_{l,t}-\mu_l)}
\end{equation*}
SATMD retains the mean distance from each layer
\begin{equation*}
    f_{\text{SATMD}} = \bigoplus_{l=0}^{L}\left(\frac{1}{T}\sum_{t=1}^T D_{l,t}\right)
\end{equation*}

\paragraph{Output-distribution features.}
The sequence-level Max. Seq. Prob. (MSP), Entropy, (log-)Perplexity, and Energy scores are
\begin{align*}
    f_\text{MSP} &= \frac{1}{T}\sum_{t=1}^T \max_{v\in\mathcal{V}} p_t(v), \\
    f_\text{Entropy} &= - \frac{1}{T} \sum_{t=1}^T \sum_{v\in\mathcal{V}} p_t(v) \log p_t(v), \\
    f_\text{Perplexity} &= -\frac{1}{T}\sum_{t=1}^{T}\log\max_{v\in\mathcal{V}} p_t(v), \\
    f_\text{Energy} &= -\frac{\tau}{T}\sum_{t=1}^{T}\log\sum_{v \in \mathcal{V}} \exp{\left(\frac{z_t(v)}{\tau}\right)}
\end{align*}
Top-$m$ Prob. retains more information from the output distribution
\begin{equation*}
    f_{\text{Top-}m} = \frac{1}{T}\sum_{t=1}^T \operatorname{Top}_m(p_t)
\end{equation*}
For Layer Top-$m$ Prob., every intermediate hidden state is projected through the output head
\begin{align*}
    p_{l,t} &= \operatorname{softmax}(W_U h_{l,t}+ b_U),\\
    f_{\text{LayerTop-}m} &= \bigoplus_{l=1}^L \left[\frac{1}{T}\sum_{t=1}^T \operatorname{Top}_m(p_{l,t})\right]
\end{align*}

\paragraph{Attention features.}

Let $\alpha_{t,j}^{l,q}$ denote the attention weight from token $t$ to $j$, at layer $l$, and head $q$. For Lookback Ratio, we compute the mean attention assigned to the input and preceding generated tokens
\begin{align*}
    A^{l,q}_{\text{ctx}}(t) &= \frac{1}{N}\sum_{j=1}^N \alpha_{t,j}^{l,q},\\
    A^{l,q}_{\text{gen}}(t) &= \frac{1}{\max(t-1,1)}\sum_{j=1}^{t-1} \alpha_{t,N+j}^{l,q},\\
    R_t^{l,q}&=\frac{A^{l,q}_{\text{ctx}}(t)}{A^{l,q}_{\text{ctx}}(t) + A^{l,q}_{\text{gen}}(t)}
\end{align*}
The fixed-length Lookback Ratio representation averages this quantity across generated tokens and concatenates all layers and head
\begin{equation*}
    f_\text{LookbackRatio} = \bigoplus_{l=1}^L\bigoplus_{q=1}^Q \left(\frac{1}{T}\sum_{t=1}^T R_t^{l,q}\right)
\end{equation*}
Recent-token Attention retains each head's attention to the $k$ immediately preceding generated tokens
\begin{equation*}
    f_\text{Attention} = \bigoplus_{l=1}^L\bigoplus_{q=1}^Q\bigoplus_{r=1}^k\left[\frac{1}{T}\sum_{t=1}^T \mathbbm{1}[t>r]\alpha_{t,N+t-r}^{l,q}\right]
\end{equation*}
Unavailable preceding positions at the beginning of a generation are zero-padded. Following the small-window setting \citep{chuang-etal-2024-lookback}, we fix $k=2$ for all settings.

\paragraph{Hybrid features.} These are constructed by direct concatenation
\begin{align*}
    f_{\text{Attention} + \text{MSP}} &= f_\text{Attention} \oplus f_{\text{MSP}},\\
    f_{\text{InternalVar}} &= f_{\text{CircVar}} \oplus f_{\text{CovDet}} \oplus f_\text{Entropy},\\
    f_{\text{SATMD} + \text{MSP}} &= f_\text{SATMD} \oplus f_\text{MSP}
\end{align*}

Finally, each representation is flattened into a fixed-length vector and standardised dimension-wise using statistics fitted only on the training split
\begin{equation*}
    \tilde{f}_j = \frac{f_j-\mu_j^\text{train}}{\sigma_j^\text{train}+\epsilon}
\end{equation*}
The linear probe then estimates correctness as
\begin{equation*}
    P(\text{$y$ is correct}|f)=\sigma(w^\top \tilde{f} + b)
\end{equation*}
\subsection{Probe Training}
\label{subsec:probe_training}

\paragraph{Data Construction.} For each LLM/dataset configuration, we train a separate uncertainty probe per feature with label 1 denoting a correct generated response.  TriviaQA, SciQ, and PopQA responses are labelled by Gemini-3.1-Flash-Lite LLM-Judge, whereas ARC, CSQA, BoolQ, and StrategyQA are labelled by comparing the parsed final answer against the reference answer. 
For each dataset, we use the full 1000-example SciQ test split and the full 1221-example CSQA validation split. For TriviaQA validation, PopQA test, BoolQ validation, StrategyQA train, and ARC Challenge split, we select a random subset of 1000 examples.
We run five-seed experiments for \Cref{fig:main_results,fig:main_add} while other experiments are single-seed. Then, we construct a stratified 80/20 train/test split using binary correctness labels. 
For the pooled seven-benchmark probes in \Cref{sec:pretrained} experiments, we combine the labelled responses, obtaining approximately 7000 instances in total.

\paragraph{Hyperparameter Specifications.} 
Each feature is normalised before probe fitting. Sequence- or layer-structured representations are flattened before normalisation. 
We employ the binary cross-entropy loss using the Adam optimiser with learning rate $10^{-3}$ and batch size 32.
We train the probes for at most 100 epochs, with early stopping of patience 20 using validation AUROC. The linear probe is a single affine map from the standardised feature vector to one logit. The MLP uses a 32-unit hidden layer with ReLU activation followed by a scalar output layer. The CNN treats the flattened feature vector as a one-channel one-dimensional sequence and applies two 1D convolutional layers (32 channels, kernel size 3, same-length padding), each followed by ReLU. An adaptive global average pool and a final linear layer produce the scalar logit.

\subsection{Prompt Templates}
\label{subsec:prompts}

\Cref{tab:benchmark_prompts,tab:reasoning_suffix,tab:section4_prompts,tab:decomposition_prompt,tab:llm_judge_prompt} report all prompts. For long reasoning, we follow CoT \citep{wei2022chain}, and we adapt it to elicit constrained medium reasoning output.
Thinking mode is disabled, and reasoning is elicited solely through prompt suffixes.
The open-ended generation prompts follow entity-centric templates of \citet{shelmanov-etal-2025-head}, with Biographies (ID) and claim-span setup aligned with \citet{han-etal-2025-simple}. For claim-level annotation, we adopt the established decompose-then-verify framework \citep{min-etal-2023-factscore, wei2024long}, extracting atomic, self-contained claims before assigning binary factuality labels. We request the LLM-judge for schema-constrained JSON outputs with verbatim source spans and offsets to support reliable parsing similar to \citet{li-etal-2025-rag}.

\begin{table}[htbp]
\renewcommand\arraystretch{1.2}
\centering
\small
\begin{tabular}{p{0.98\columnwidth}}
\toprule
\rowcolor{gray!20}\textbf{Shared Prefix} \\
You are a helpful assistant, providing accurate and concise information without overthinking. \\
\midrule
\rowcolor{gray!20}\textbf{TriviaQA, SciQ, PopQA} \\
\texttt{\{question\}} Be concise, and output only the final answer. \\
\midrule
\rowcolor{gray!20}\textbf{ARC} \\
Answer the following multiple choice science question. Your final answer should be of the following format: ``Answer: \$LABEL'' (without quotes), where LABEL is one of the provided choice labels. Be concise, and output only the final answer.

\texttt{Question: \{question\}}

\texttt{Choices:}

\texttt{\{label\}. \{choice\}} \\
\midrule
\rowcolor{gray!20}\textbf{BoolQ} \\
Read the passage and answer the yes/no question. Your final answer should be of the following format: ``Answer: yes'' or ``Answer: no'' (without quotes). Be concise, and output only the final answer.

\texttt{Passage: \{passage\}}

\texttt{Question: \{question\}} \\
\midrule
\rowcolor{gray!20}\textbf{CommonsenseQA} \\
Answer the following multiple choice common-sense reasoning question. Your final answer should be of the following format: ``Answer: \$LETTER'' (without quotes), where LETTER is one of ABCDE. Be concise, and output only the final answer.

\texttt{\{question-with-choices\}} \\
\midrule

\rowcolor{gray!20}\textbf{StrategyQA} \\
Answer the following yes/no question. Your final answer should be of the following format: ``Answer: yes'' or ``Answer: no'' (without quotes). Be concise, and output only the final answer.

\texttt{\{question\}} \\
\bottomrule
\end{tabular}
\caption{Generation prompts for the Section~3 datasets.}
\label{tab:benchmark_prompts}
\end{table}

\section{Supplementary Experiments}
\label{app:supp_expt}

\begin{table}[htbp]
\renewcommand\arraystretch{1.2}
\centering
\small
\begin{tabular}{p{0.13\columnwidth}p{0.77\columnwidth}}
\toprule
\textbf{Mode} & \textbf{Suffix} \\
\midrule
\textbf{NR} & Be concise, and output only the final answer. \\
\textbf{MR} & Output a single sentence of reasoning before the final answer. \\
\textbf{LR} & Think step by step and output the reasoning process before answering. \\
\bottomrule
\end{tabular}
\caption{Reasoning-control suffixes for the NR, MR, and LR prompts for ARC and CommonsenseQA.}
\label{tab:reasoning_suffix}
\end{table}

\begin{table}[htbp]
\renewcommand\arraystretch{1.4}
\centering
\small
\begin{tabular}{p{0.32\columnwidth}p{0.60\columnwidth}}
\toprule
\textbf{Domain} & \textbf{Prompt} \\
\midrule
Biographies (ID) & Tell me a paragraph bio of \texttt{\{person\}}. \\
Biographies (OOD) & Tell me a bio of \texttt{\{person\}}. \\
Artworks & Tell me about the artwork \texttt{\{artwork\}} and its artist. \\
Books & Tell me about the book \texttt{\{book\}} and its author. \\
Cities & Tell me the history of \texttt{\{city\}}. \\
Events & Tell me about \texttt{\{event\}} event. \\
Inventions & Tell me about the invention of \texttt{\{invention\}} and its inventor. \\
Landmarks & Tell me about the landmark \texttt{\{landmark\}}. \\
Movies & Tell me about the movie \texttt{\{movie\}} and its cast. \\
\bottomrule
\end{tabular}
\caption{Prompts for open-ended entity-centric generation in Section 4.}
\label{tab:section4_prompts}
\end{table}

\begin{table*}[htbp]
\renewcommand\arraystretch{1.1}
\centering
\small
\begin{tabular}{p{0.15\textwidth}p{0.79\textwidth}}
\toprule
\rowcolor{gray!20}\textbf{Component} & \textbf{Prompt} \\
System & You decompose long-form passages into atomic factual claims. Return JSON only. \\
\midrule
User & \textbf{Task:} Break the passage into atomic factual claims.

\textbf{Rules:} (1) Each claim should contain exactly one checkable piece of information. (2) Claims should be self-contained; include the relevant person/entity name when needed. (3) Do not include opinions, formatting, or non-factual filler. (4) For every claim, copy the shortest exact contiguous source quote from the passage that supports where the claim came from. (5) The source quote must be a verbatim substring of the passage. (6) Return \texttt{source\_start} and \texttt{source\_end} as zero-based Python character offsets into the passage for \texttt{source\_quote}; \texttt{source\_end} is exclusive. (7) Return claims in passage order. Source spans may overlap when one sentence supports multiple atomic claims.

Return JSON only:
\texttt{\{"claims":[\{"claim":"...","source\_quote":"...",}
\texttt{"source\_start":0,"source\_end":10\}]\}}

Return JSON matching this schema: \texttt{\{decomposition-schema\}}

\texttt{Entity: \{entity\}} \emph{(included when available)}

\textbf{Passage}

\texttt{\{generated-passage\}} \\
\bottomrule
\end{tabular}
\caption{System and user prompts used to decompose each long-form generation into atomic claims. The API is run with temperature zero and JSON output enabled.}
\label{tab:decomposition_prompt}
\end{table*}

\begin{table*}[htbp]
\renewcommand\arraystretch{1.1}
\centering
\small
\begin{tabular}{p{0.15\textwidth}p{0.79\textwidth}}
\toprule
\rowcolor{gray!20}\textbf{Component} & \textbf{Prompt} \\
System & You are a strict factuality evaluation assistant. Return JSON only, with exactly one result per input ID: \texttt{\{"results":[\{"id":"...","label":1\}]\}}. \\
\midrule
User & You are evaluating one atomic factual claim using the supplied reference evidence. Label only whether every factual detail in the claim is directly and precisely supported by that evidence; do not use the generated answer as evidence.

0 means contradicted, unsupported, unverifiable, vague, approximate, wrong entity/topic, or only partially supported. 1 means fully and precisely supported. Be strict: if any part is uncertain or unsupported, return 0.

Return JSON matching this schema:
\texttt{\{label-schema\}}

Claims to evaluate:
\texttt{[\{"id":\{id\}, "entity":\{entity\}, "prompt":\{generation-prompt\},}
\texttt{"claim\_to\_label":\{atomic-claim\}, "reference\_source":}
\texttt{\{"title":\{title\}, "url":\{url\}, "evidence":\{evidence\}\}\}]} \\
\bottomrule
\end{tabular}
\caption{System and user prompts for claim-level annotation by LLM-Judge. Annotations for Biographies (ID) are evidence-grounded while other domains are not.}
\label{tab:llm_judge_prompt}
\end{table*}

\begin{table}[htbp]
\renewcommand\arraystretch{1.1}
\centering
\resizebox{\linewidth}{!}{
\begin{tabular}{lccc|ccc}
\toprule
\multirow{2}{*}{Domain} & \multicolumn{3}{c}{\# Claims} & \multicolumn{3}{c}{Correct Claims (\%)} \\
\cmidrule(lr){2-4}\cmidrule(lr){5-7}
& Qwen & Gemma & Llama & Qwen & Gemma & Llama \\
\midrule
Bio (ID) & 822 & 907 & 800 & 33.09 & 40.46 & 58.13 \\
Bio (OOD) & 682 & 703 & 983 & 66.86 & 78.09 & 80.37 \\
Artworks & 603 & 595 & 623 & 85.57 & 84.54 & 82.99 \\
Books & 746 & 637 & 784 & 88.87 & 95.45 & 87.37 \\
Cities & 714 & 542 & 714 & 71.57 & 75.65 & 72.97 \\
Events & 675 & 550 & 711 & 88.74 & 93.09 & 85.65 \\
Inventions & 692 & 545 & 715 & 86.99 & 86.97 & 76.64 \\
Landmarks & 803 & 675 & 820 & 83.94 & 89.78 & 81.10 \\
Movies & 722 & 611 & 905 & 76.73 & 92.31 & 87.62 \\
\bottomrule
\end{tabular}
}
\caption{Statistics for open-ended generation datasets.}
\label{tab:section4_dataset_statistics}
\end{table}

Here, we provide supplementary experiments.

\begin{figure*}
    \centering
    \includegraphics[width=\linewidth]{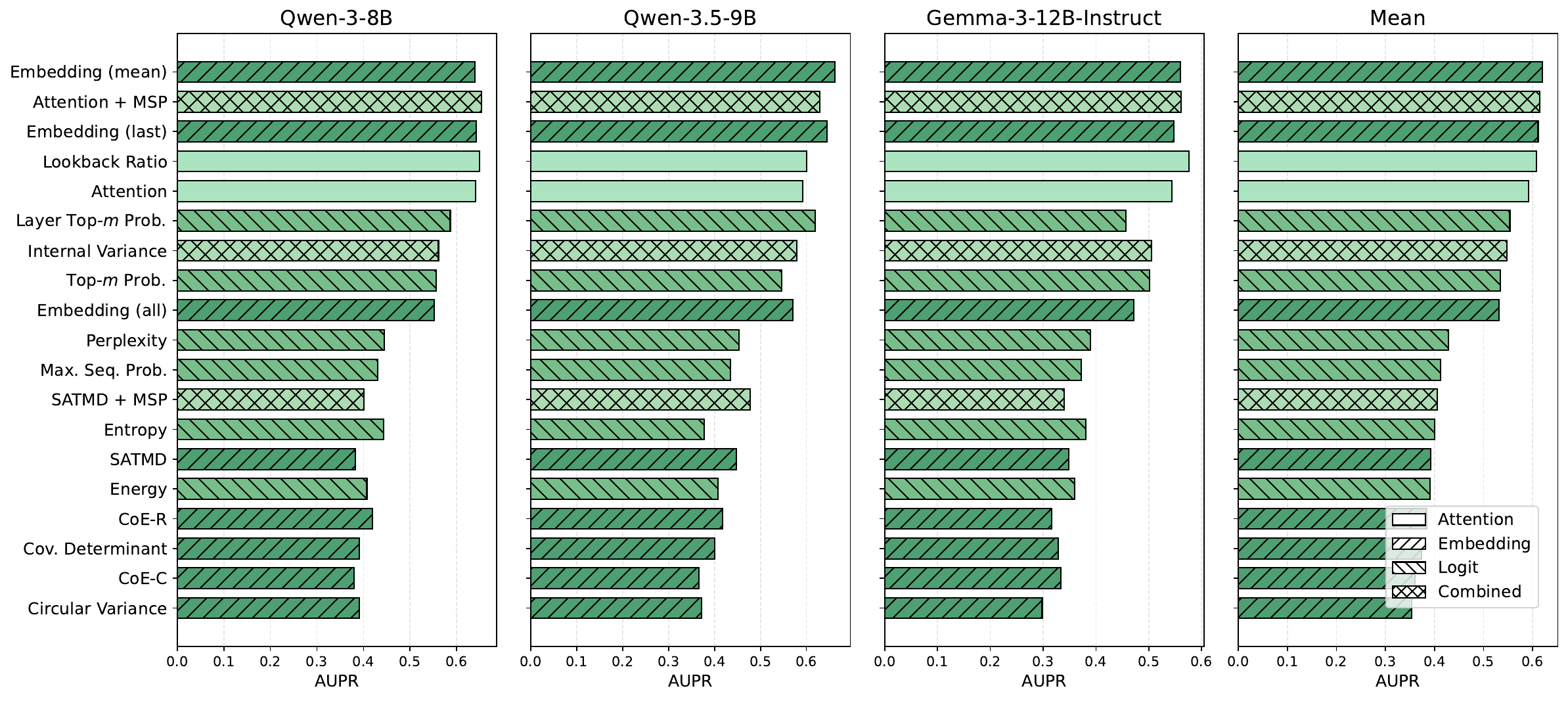}
    \caption{AUPR for the main models.}
    \label{fig:aupr}
\end{figure*}

\paragraph{AUPR Results.} We report the five-seed AUPR in \Cref{fig:aupr}, where incorrect responses represent the positive class, and summarise the rankings in \Cref{tab:ranking}. These results support our main conclusions.

\begin{figure*}
    \centering
    \begin{subfigure}[b]{0.49\textwidth}
        \centering
        \includegraphics[width=\linewidth]{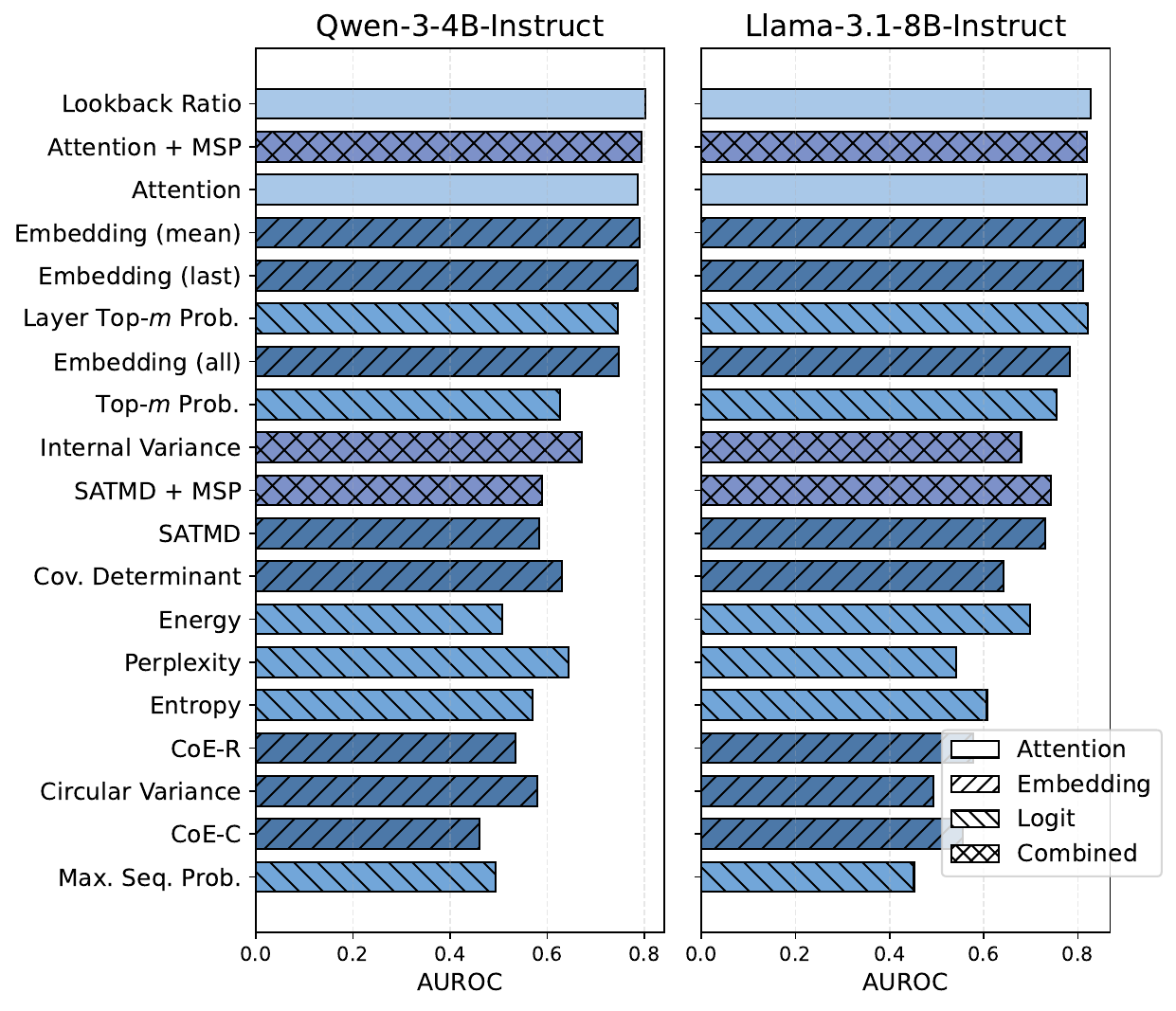}
    \end{subfigure}
    \hfill
    \begin{subfigure}[b]{0.49\textwidth}
        \centering
        \includegraphics[width=\linewidth]{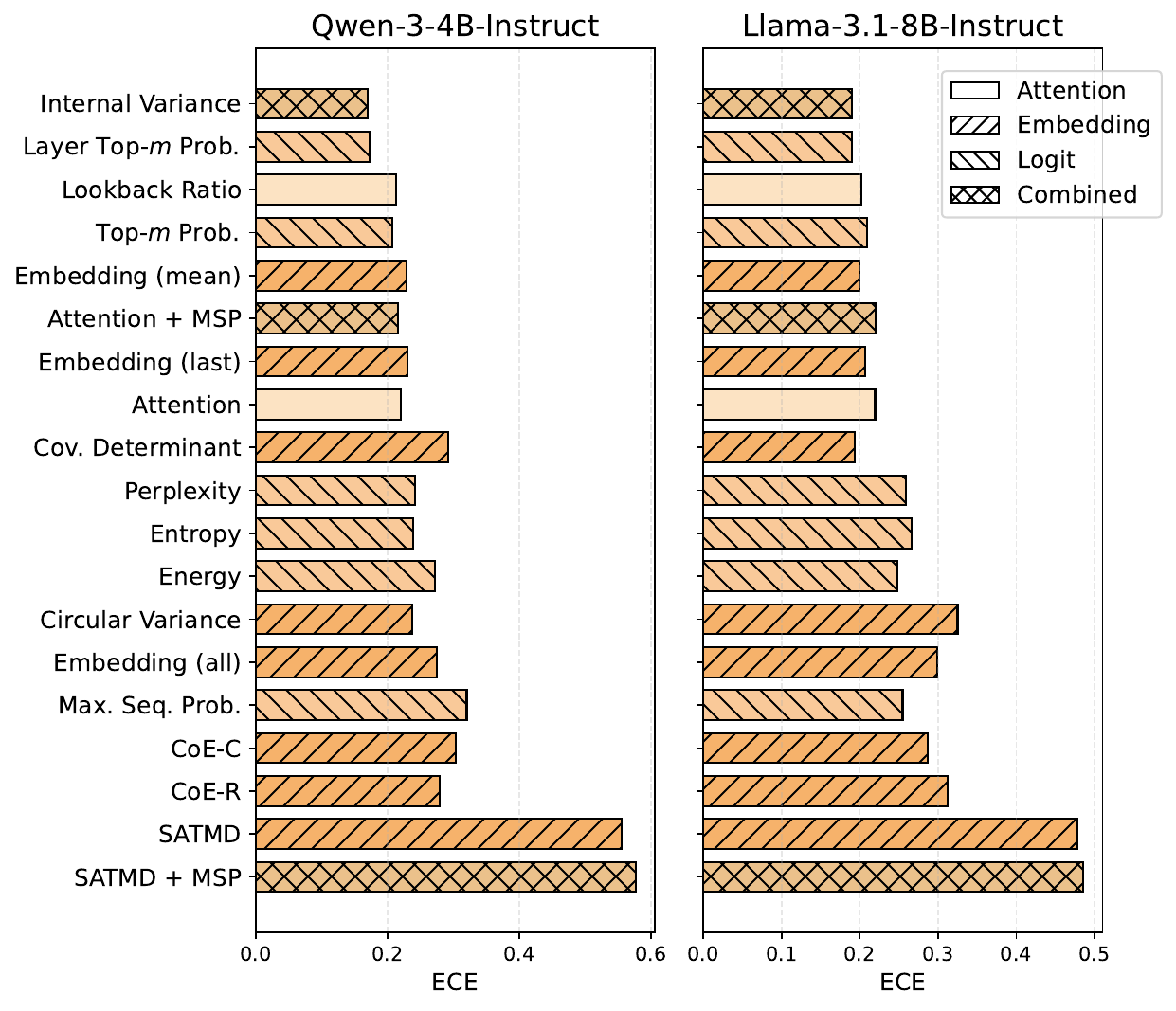}
    \end{subfigure}
    \caption{AUROC and ECE of Llama-3.1-8B-Instruct and Qwen-3-4B-Instruct.}
    \label{fig:main_add}
\end{figure*}

\begin{figure*}
    \centering
    \includegraphics[width=\linewidth]{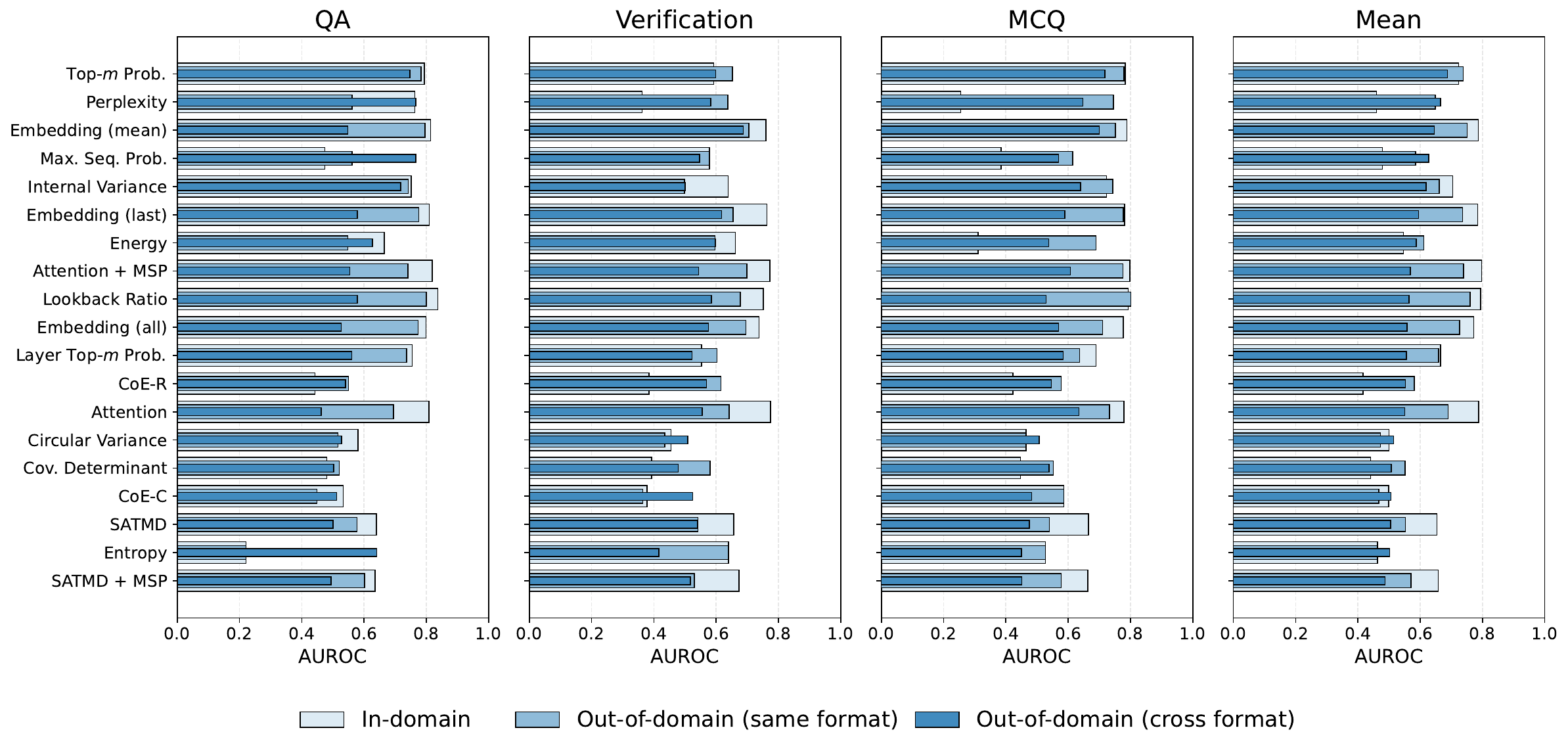}
    \caption{Benchmark-transfer performance with Gemma-3-12B-Instruct. Average AUROC across In-domain, Out-of-domain (same task), and Out-of-domain (cross task) configurations.}
    \label{fig:task_transfer_auroc_gemma}
\end{figure*}

\paragraph{Additional Model Results.}
\Cref{fig:main_add} reports in-domain AUROC and ECE for Qwen-3-4B-Instruct and Llama-3.1-8B-Instruct.
The overall trends are consistent with the main results: hidden state and attention-based features remain strong in-domain.
\Cref{fig:task_transfer_auroc_gemma} replicates task transfer analysis in \Cref{subsec:transfer} with Gemma-3-12B-Instruct.
The broad conclusion remains consistent: hidden state and attention features lose much of their in-domain advantage, particularly under cross-task shift, whereas several structured probability and dispersion features retain performance more consistently. However, the ordering of individual representations changes substantially, indicating that transfer robustness is model-dependent, with no dominant feature.

\begin{table*}[!ht]
\centering
\renewcommand{\arraystretch}{1.5}
\setlength{\tabcolsep}{3pt}
\resizebox{\textwidth}{!}{
\begin{tabular}{l c c | l c c | l c c}
\toprule
\multicolumn{3}{c}{\cellcolor{blue!10} \textbf{Discriminability}} & \multicolumn{3}{c}{\cellcolor{orange!10} \textbf{Calibration}} & \multicolumn{3}{c} {\cellcolor{green!10} \textbf{Precision}}\\
\midrule
\textbf{Feature} & \textbf{AUROC} & \textbf{Mean Rank (SE)} & \textbf{Feature} & \textbf{ECE} & \textbf{Mean Rank (SE)} & \textbf{Feature} & \textbf{AUPR} & \textbf{Mean Rank (SE)}\\
\midrule
Embedding (Mean) & 82.77 & 3.68 (0.33) & Internal Variance & 15.42 & 4.78 (0.59) & Embedding (Mean) & 62.05 & 3.56 (0.36) \\
Lookback Ratio & 82.38 & 4.14 (0.51) & Top-$m$ Prob. & 18.53 & 6.30 (0.53) & Attention + MSP & 61.45 & 4.08 (0.32) \\
Attention + MSP & 82.03 & 4.32 (0.36) & Lookback Ratio & 20.84 & 6.98 (0.50) & Lookback Ratio & 60.86 & 4.15 (0.47) \\
Embedding (Last) & 81.88 & 4.41 (0.40) & Layer Top-$m$ Prob. & 20.04 & 7.10 (0.56) & Embedding (Last) & 61.14 & 4.22 (0.50) \\
Attention & 81.18 & 5.23 (0.56) & Embedding (Last) & 21.55 & 7.43 (0.41) & Attention & 59.22 & 5.85 (0.56) \\
Embedding (All) & 80.91 & 5.60 (0.36) & Attention + MSP & 21.48 & 7.64 (0.44) & Layer Top-$m$ Prob. & 53.48 & 7.26 (0.65) \\
Internal Variance & 75.56 & 8.17 (0.58) & Attention & 21.63 & 8.02 (0.50) & Internal Variance & 54.84 & 7.65 (0.52) \\
Layer Top-$m$ Prob. & 76.46 & 8.28 (0.72) & Embedding (Mean) & 22.67 & 8.46 (0.42) & Top-$m$ Prob. & 53.48 & 7.96 (0.63) \\
Top-$m$ Prob. & 75.28 & 8.59 (0.60) & Max. Seq. Prob. & 28.22 & 10.12 (0.73) & Embedding (All) & 53.15 & 8.58 (0.47) \\
SATMD + MSP & 65.98 & 11.75 (0.63) & CoE-R & 28.72 & 10.56 (0.48) & Perplexity & 42.87 & 12.17 (0.49) \\
SATMD & 64.77 & 12.36 (0.65) & Embedding (All) & 26.80 & 10.62 (0.92) & Entropy & 40.07 & 12.26 (0.82) \\
Perplexity & 52.84 & 12.70 (0.79) & Circular Variance & 29.67 & 11.17 (0.62) & SATMD + MSP & 40.60 & 12.95 (0.55) \\
Entropy & 58.33 & 12.86 (0.51) & Energy & 30.08 & 11.20 (0.41) & Max. Seq. Prob. & 41.27 & 13.10 (0.51) \\
Max. Seq. Prob. & 57.17 & 13.72 (0.46) & Cov. Determinant & 29.77 & 11.34 (0.50) & Energy & 39.13 & 13.73 (0.47) \\
Energy & 55.95 & 13.96 (0.44) & Perplexity & 31.13 & 11.41 (0.51) & SATMD & 39.28 & 13.82 (0.51) \\
Cov. Determinant & 52.91 & 14.78 (0.36) & CoE-C & 30.94 & 11.64 (0.55) & CoE-R & 38.42 & 14.20 (0.43) \\
CoE-R & 55.06 & 14.81 (0.31) & Entropy & 32.56 & 11.98 (0.64) & Cov. Determinant & 37.30 & 14.50 (0.39) \\
Circular Variance & 51.54 & 15.27 (0.30) & SATMD + MSP & 59.48 & 16.59 (0.72) & CoE-C & 35.99 & 14.95 (0.39) \\
CoE-C & 53.48 & 15.35 (0.32) & SATMD & 62.36 & 16.65 (0.89) & Circular Variance & 35.37 & 15.00 (0.33) \\
\bottomrule
\end{tabular}
}
\caption{Summary ranking statistics.}
\label{tab:ranking}
\end{table*}

\paragraph{Aggregate Ranking Statistics.}
\Cref{tab:ranking} lists the explicit ranking statistics for the main results in \Cref{fig:main_results,fig:main_add} aggregated across all datasets, models, and five seeds.
\Cref{tab:transfer_ranking} summarises the ranking for the benchmark transfer analysis in \Cref{fig:task_transfer_auroc}. To obtain the SE's, we bootstrap resamples of target datasets within each task; thus, these values describe uncertainty in method rankings across these benchmarks.

\begin{figure*}
    \centering
    \includegraphics[width=\linewidth]{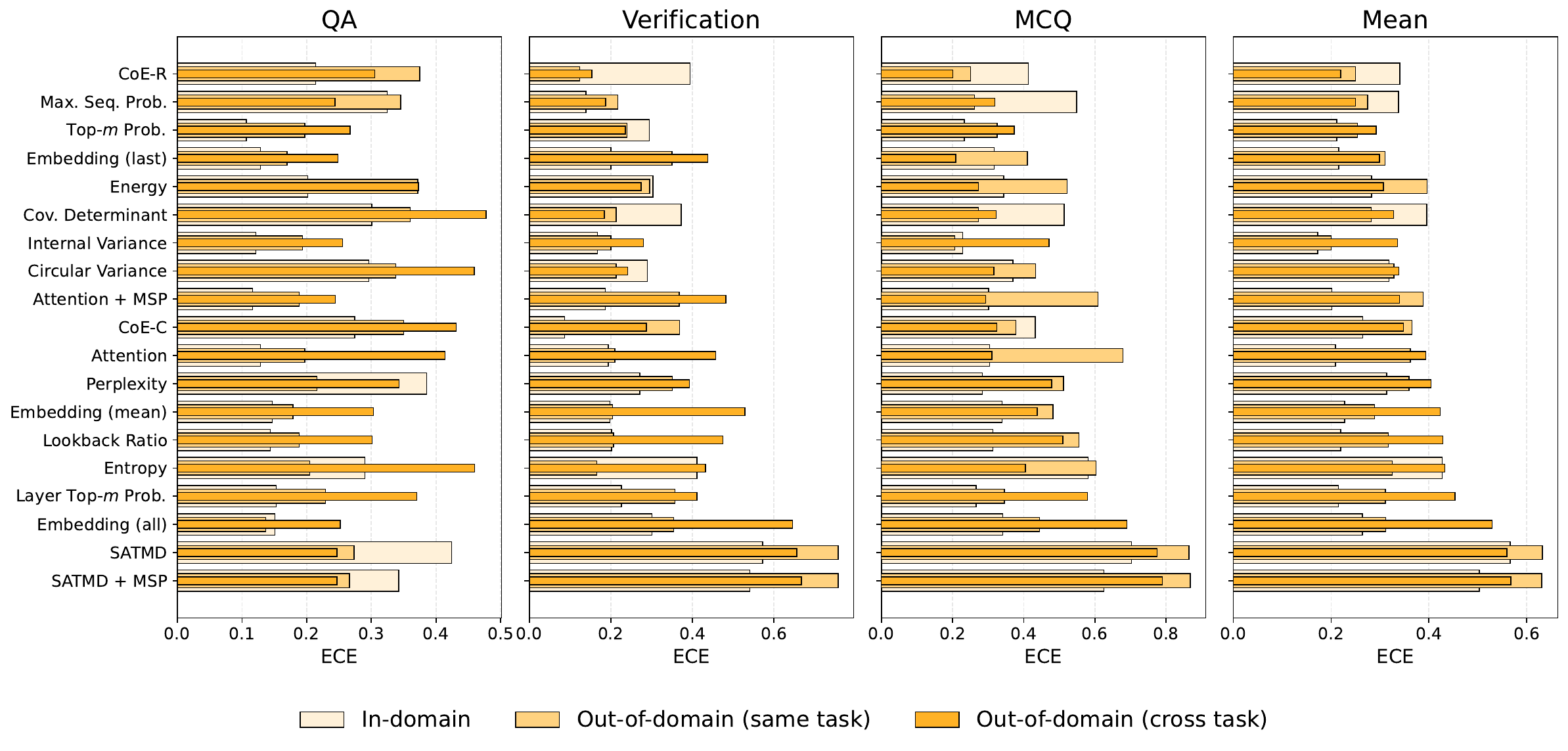}
    \caption{Benchmark-transfer performance. Average ECE across In-domain, Out-of-domain (same task), and Out-of-domain (cross task) configurations.}
    \label{fig:task_transfer_ece}
\end{figure*}

\paragraph{Calibration under Transfer.}
\Cref{fig:task_transfer_ece} reports ECE under the same in-domain, OOD same-task, and OOD cross-task settings. This shows that discriminability and calibration do not always move together. Features with strong AUROC are not necessarily the best calibrated, and some lower-dimensional structured features remain comparatively stable under shift.

\begin{figure*}
    \centering
    \includegraphics[width=\linewidth]{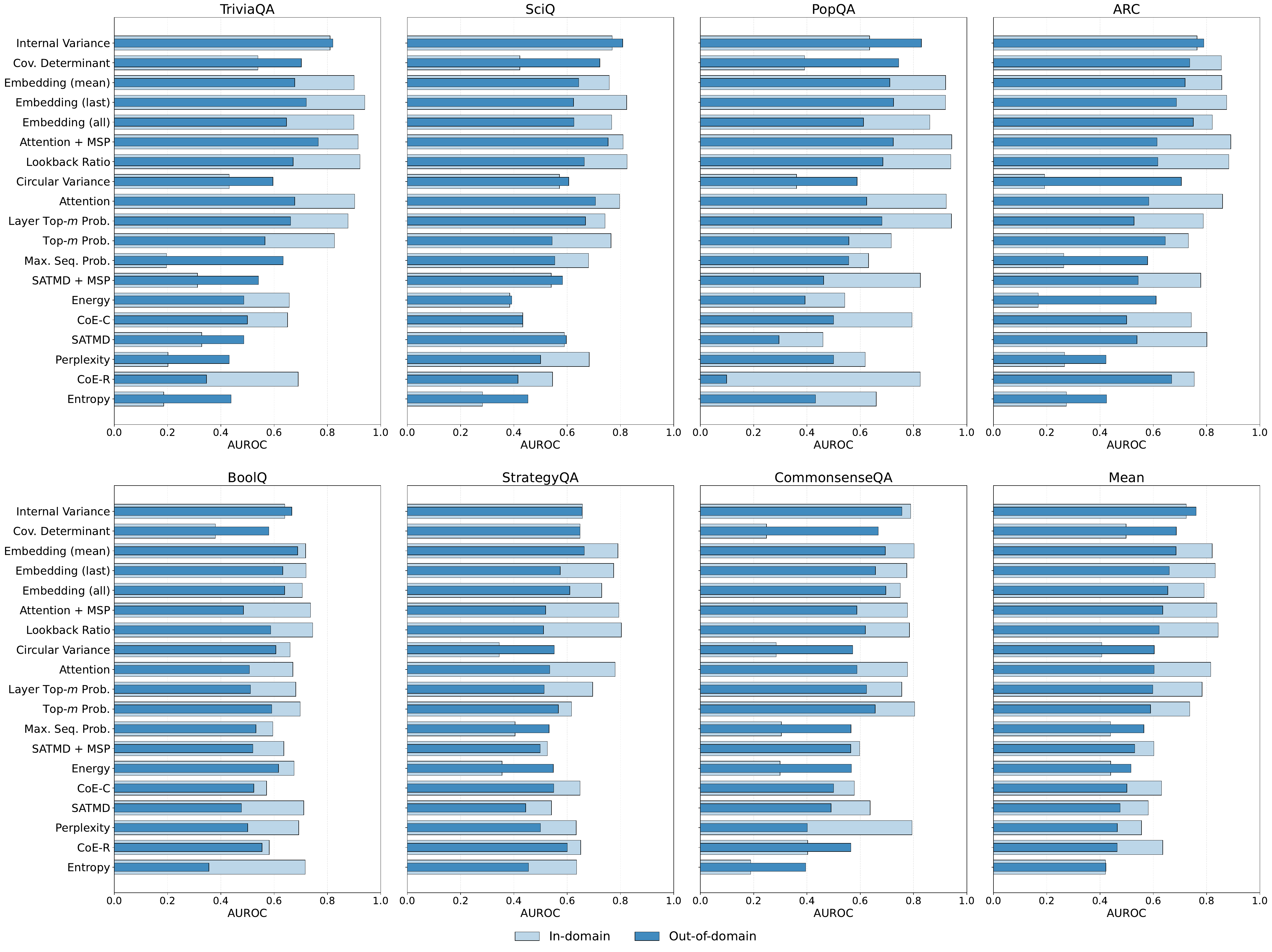}
    \caption{In-domain AUROC against OOD setting for individual benchmarks with Qwen-3-8B.}
    \label{fig:transfer_auroc}
\end{figure*}

\begin{figure*}
    \centering
    \includegraphics[width=\linewidth]{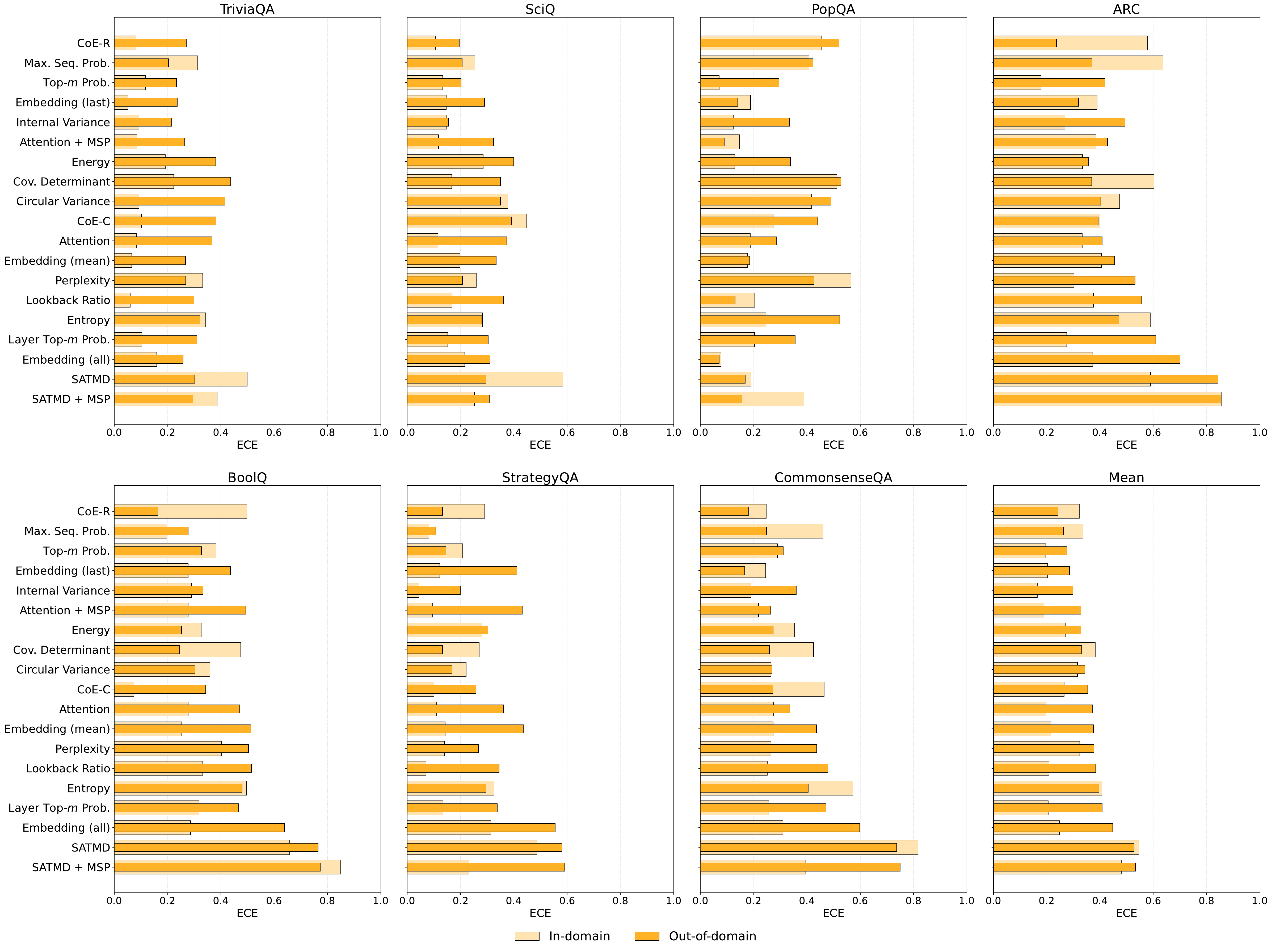}
    \caption{In-domain ECE against OOD setting for individual benchmarks with Qwen-3-8B.}
    \label{fig:transfer_ece}
\end{figure*}

\paragraph{Per-dataset Transfer.}
\Cref{fig:transfer_auroc} and \Cref{fig:transfer_ece} provide per-dataset transfer results for AUROC and ECE, respectively. While the degree of transfer degradation varies across datasets, hidden state features generally remain strong in-domain, whereas structured features tend to retain performance more consistently under OOD evaluation.
In particular, some features occasionally demonstrate significant OOD improvements, namely Cov. Determinant and Circular Variance.
Both features reduce cross-layer dispersion into a single scalar, so the linear probe primarily determines the direction and threshold of its association with correctness. Our per-dataset results show that this direction can be unstable: on the same target, probes trained on different sources sometimes produce approximately reversed rankings. An ID probe can therefore perform poorly when its training split estimates the direction unreliably, while an OOD source may learn the direction that happens to align better with the target test split. 
The apparent improvement should thus not be interpreted as distribution shift improving the feature. Rather, it exposes the instability of fitting individual scalar dispersion measures. \Cref{fig:task_transfer_auroc} further accentuates the pattern by averaging multiple unstable source-target transfers.

\begin{figure}
    \begin{subfigure}[b]{0.49\textwidth}
        \centering
        \includegraphics[width=\linewidth]{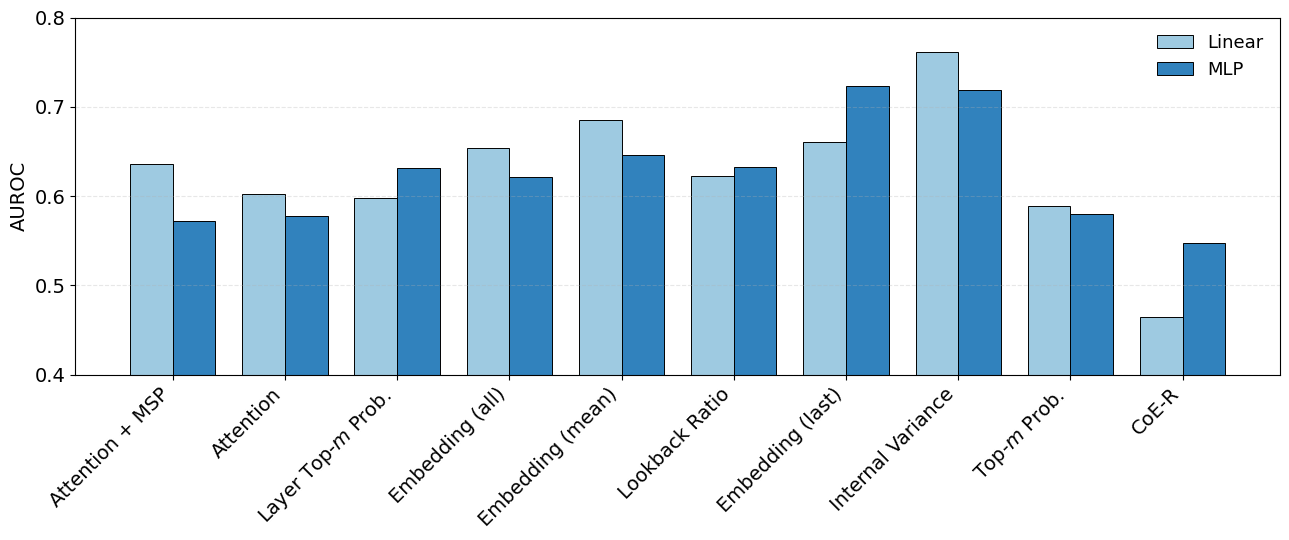}
    \end{subfigure}
    \hfill
    \begin{subfigure}[b]{0.49\textwidth}
        \centering
        \includegraphics[width=\linewidth]{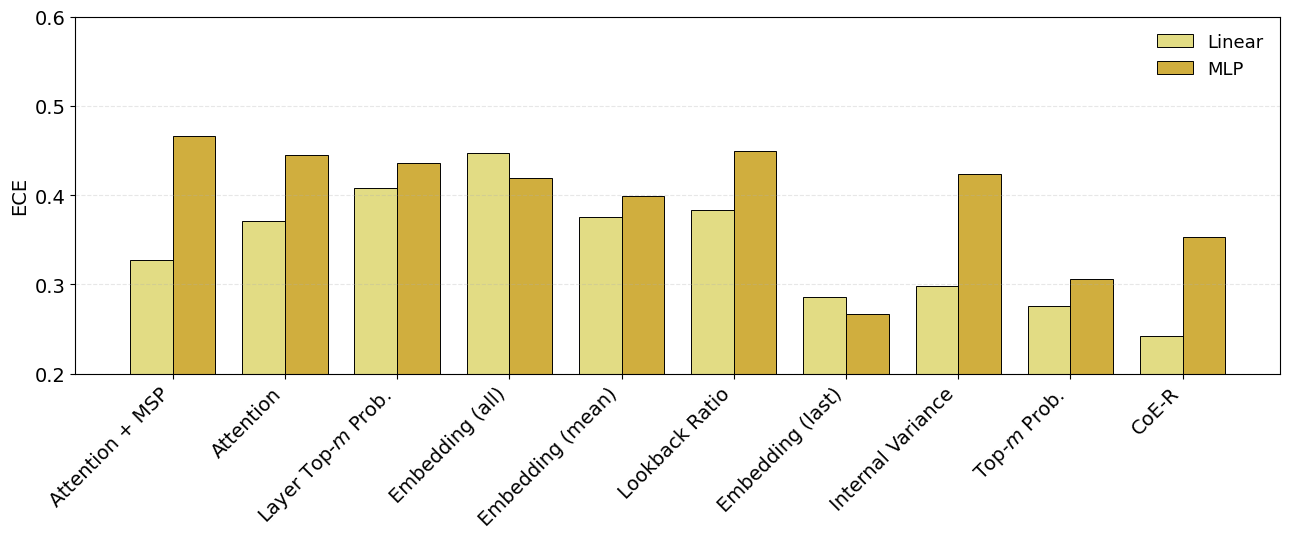}
    \end{subfigure}
    \caption{Average OOD performance with different probe architectures.}
    \label{fig:probe_transfer}
\end{figure}

\begin{figure}[htbp]
    \centering
    \includegraphics[width=\linewidth]{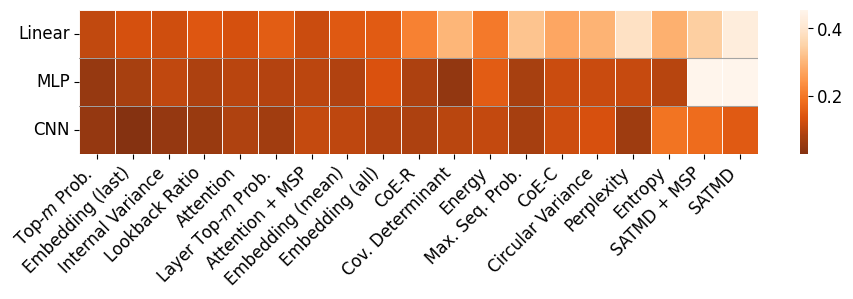}
    \caption{Effect of probe architecture on ECE.}
    \label{fig:probe_ece}
\end{figure}

\begin{figure}
    \centering
    \includegraphics[width=\linewidth]{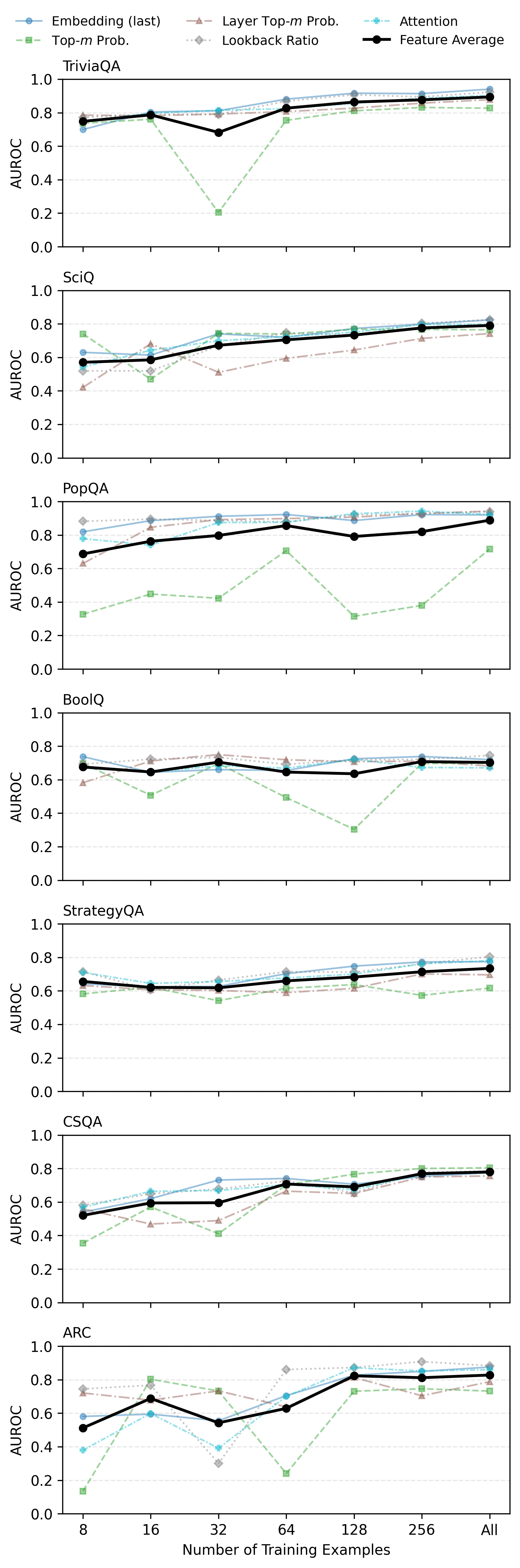}
    \caption{AUROC for individual dataset with varying training set size.}
    \label{fig:trsize_indiv}
\end{figure}

\paragraph{Transfer Ranking.}
\Cref{tab:transfer_ranking} presents a summary ranking for the main transfer AUROC results.

\begin{table}
\centering
\renewcommand{\arraystretch}{1.25}
\setlength{\tabcolsep}{3pt}
\resizebox{\linewidth}{!}{
\begin{tabular}{lrr|lrr}
\toprule
\multicolumn{3}{c|}{OOD (same task)} & \multicolumn{3}{c}{OOD (cross task)} \\ \midrule
Feature & AUROC & Mean rank (SE) & Feature & AUROC & Mean rank (SE) \\
\midrule
Lookback Ratio & 82.13 & 3.78 (0.53) & Internal Variance & 74.56 & 2.84 (0.33) \\
Embedding (mean) & 79.67 & 4.50 (0.79) & Cov. Determinant & 66.14 & 5.68 (0.51) \\
Embedding (last) & 79.19 & 4.72 (0.86) & Embedding (mean) & 63.95 & 7.22 (0.40) \\
Attention + MSP & 77.82 & 5.39 (1.33) & Circular Variance & 58.38 & 8.01 (0.43) \\
Attention & 77.39 & 6.28 (1.52) & Embedding (all) & 61.06 & 8.18 (0.39) \\
Internal Variance & 76.03 & 6.39 (0.59) & Top-$m$ Prob. & 58.75 & 8.92 (0.44) \\
Cov. Determinant & 71.74 & 7.94 (1.03) & Embedding (last) & 60.34 & 9.03 (0.68) \\
Embedding (all) & 76.04 & 7.97 (0.11) & Max. Seq. Prob. & 57.10 & 9.62 (0.29) \\
Energy & 57.58 & 10.33 (1.86) & Energy & 52.06 & 10.18 (0.33) \\
Circular Variance & 61.29 & 10.89 (2.30) & Attention + MSP & 56.65 & 10.71 (0.28) \\
Layer Top-$m$ Prob. & 67.55 & 11.14 (1.19) & Layer Top-$m$ Prob. & 55.38 & 10.98 (0.48) \\
Top-$m$ Prob. & 63.90 & 11.78 (1.08) & Lookback Ratio & 54.70 & 11.33 (0.39) \\
Max. Seq. Prob. & 58.49 & 12.39 (0.51) & CoE-C & 51.13 & 11.50 (0.50) \\
CoE-R & 55.26 & 12.94 (0.85) & Attention & 52.98 & 11.63 (0.31) \\
Perplexity & 51.07 & 13.33 (1.58) & CoE-R & 47.40 & 12.13 (0.76) \\
SATMD + MSP & 57.81 & 14.22 (0.70) & Perplexity & 45.81 & 12.13 (0.30) \\
SATMD & 56.08 & 14.44 (0.58) & SATMD + MSP & 51.48 & 12.34 (0.15) \\
CoE-C & 49.44 & 15.50 (0.91) & Entropy & 42.12 & 13.51 (0.16) \\
Entropy & 38.90 & 16.06 (1.07) & SATMD & 45.75 & 14.04 (0.51) \\
\bottomrule
\end{tabular}
}
\caption{Transfer summary ranking statistics.}
\label{tab:transfer_ranking}
\end{table}

\paragraph{Probe Architecture and Training Size.}
\Cref{fig:probe_transfer} and \Cref{fig:probe_ece} provide average OOD performance for different probe architectures and effect of probe architecture on ECE, respectively. Higher-capacity probes can improve some weak feature representations, but do not consistently improve transfer performance, supporting our use of linear probes as the default architecture. \Cref{fig:trsize_indiv} presents per-dataset performance with different training set sizes, showing that performance often improves rapidly with limited supervision and achieves a satisfactory level at around 128--256 examples for many feature representations.

\begin{table}[htbp]
    \centering
    \resizebox{\linewidth}{!}{
    \begin{tabular}{l|ccc}
    \toprule
    \textbf{Annotator} & \textbf{TriviaQA} & \textbf{SciQ} & \textbf{PopQA} \\
    \midrule
    Rouge-1 & .684 (85\%) & .237 (56\%) & .756 (95\%) \\
    Rouge-2 & .245 (66\%) & .069 (32\%) & .319 (89\%) \\
    Rouge-L & .684 (85\%) & .228 (55\%) & .756 (95\%) \\
    AlignScore & .728 (87\%) & .220 (60\%) & .756 (95\%) \\
    LLM-as-a-judge & .960 (98\%) & .820 (94\%) & .879 (97\%) \\
    \bottomrule
    \end{tabular}
    }
    \caption{Cohen's kappa (agreement \%) for automated labels and human judgement.}
    \label{tab:labels}
\end{table}

\begin{table}[htbp]
    \centering
    \setlength{\tabcolsep}{4.0pt}
    \resizebox{\linewidth}{!}{
    \begin{tabular}{l|ccc}
    \toprule
    \textbf{LLM-Judge} & \textbf{TriviaQA} & \textbf{SciQ} & \textbf{PopQA} \\
    \midrule
    Gemini-3.1-Flash-Lite & .960 (98\%) & .820 (94\%) & .879 (97\%) \\
    GPT-5.4-Mini & .940 (97\%) & .702 (90\%) & .917 (98\%) \\
    \midrule
    & \textbf{Bio (ID)} & \textbf{Artworks} & \textbf{Books} \\
    \midrule
    Gemini-3.1-Flash-Lite & .715 (89\%) & .153 (91\%) & .000 (91\%) \\
    GPT-5.4-Mini & .836 (93\%) & .627 (93\%) & .595 (92\%) \\
    \bottomrule
    \end{tabular}
    }
    \caption{Cohen's kappa (agreement \%) between LLM-judge and human labels.}
    \label{tab:labels_llm}
\end{table}

\paragraph{Lexical and LLM-as-a-judge Annotation Validation.} \label{par:llm_judge_val}

On TriviaQA, SciQ, and PopQA with Qwen-3-8B, %
we binarise Rouge and AlignScore using a generic decision threshold of 0.5, consistent with prior work \citep{duan-etal-2024-shifting, lei2023chain}.
To assess the fidelity of the automatically generated lexical labels, we randomly select 100 generations from each dataset, and measure agreement of correctness label between the automated scorer and a human annotator.
To obtain gold ground truth labels, we instruct the human annotator to compare the generated answer with the reference answer, and to use web search if a further check is required.
\Cref{tab:labels} presents Cohen's kappa and percentage of agreed labels between the automated and human scorer. 
Furthermore, to assess the reliability of the LLM-as-a-judge labels in \Cref{sec:pretrained}, we compare the same subsets from TriviaQA, SciQ, and PopQA, and additionally 105, 113, and 104 claims from Bio (ID), Artworks, and Books passages, against human annotations.
Consistent with findings by \citet{janiak-etal-2025-illusion}, LLM-as-a-judge aligns much more closely with human judgement than lexical labels, with AlignScore as second-best but still subpar by a large margin.
Due to its high agreement, we utilise LLM-as-a-judge labels as the ground truth.
Between the two LLM-judges, GPT-5.4-Mini shows stronger agreement on the open-ended generation setting, while Gemini-3.1-Flash-Lite remains comparable on the short-answer benchmark datasets.
Gemini's tendency to label claims as true results in lower Cohen's $\kappa$, falling to 0 for the Books subset as it labelled all instances as true.
We select the judge according to the annotation regime rather than enforcing the same judge across substantially different tasks.
Thus, we use Gemini labels for the benchmark experiments in \Cref{subsec:transfer} and GPT-5.4-Mini labels for the open-ended factual generation experiments in \Cref{sec:pretrained}.

\begin{figure}
    \centering
    \includegraphics[width=\linewidth]{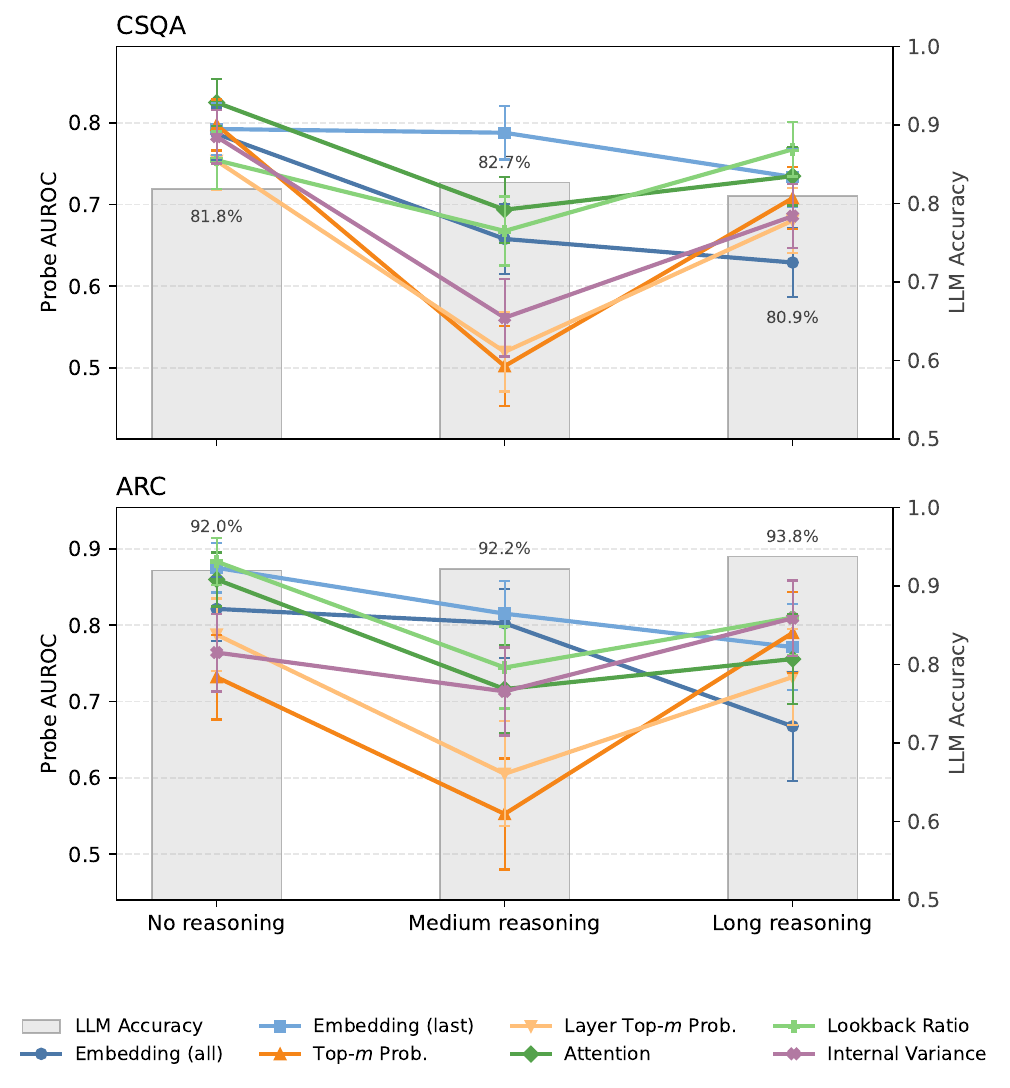}
    \caption{Effects of inducing reasoning (Qwen-3-8B).}
    \label{fig:prompt}
\end{figure}

\begin{figure}
    \centering
    \begin{subfigure}[b]{\linewidth}
        \centering
        \includegraphics[width=\linewidth]{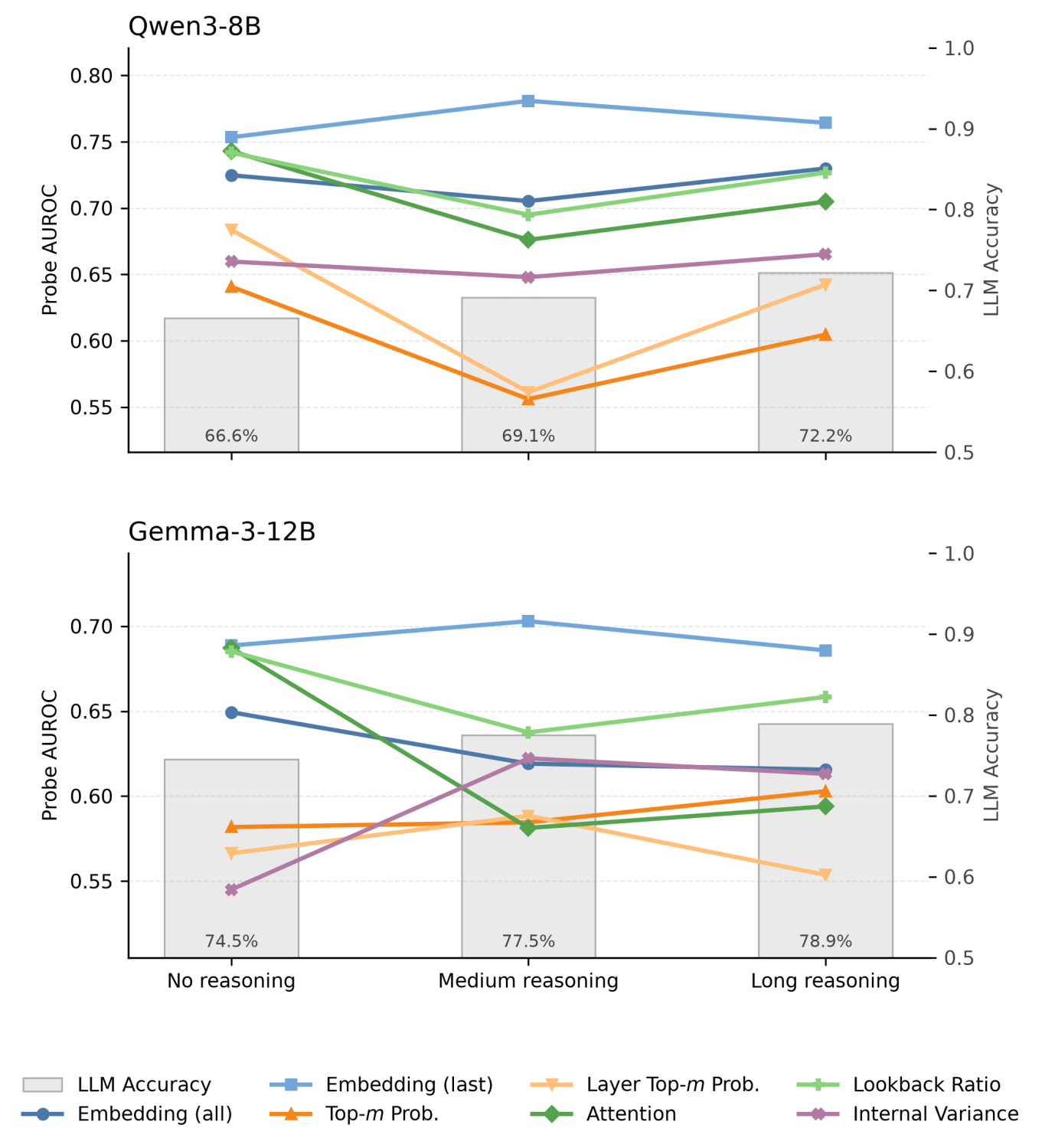}
    \end{subfigure}
    \begin{subfigure}[b]{\linewidth}
        \centering
        \includegraphics[width=\linewidth]{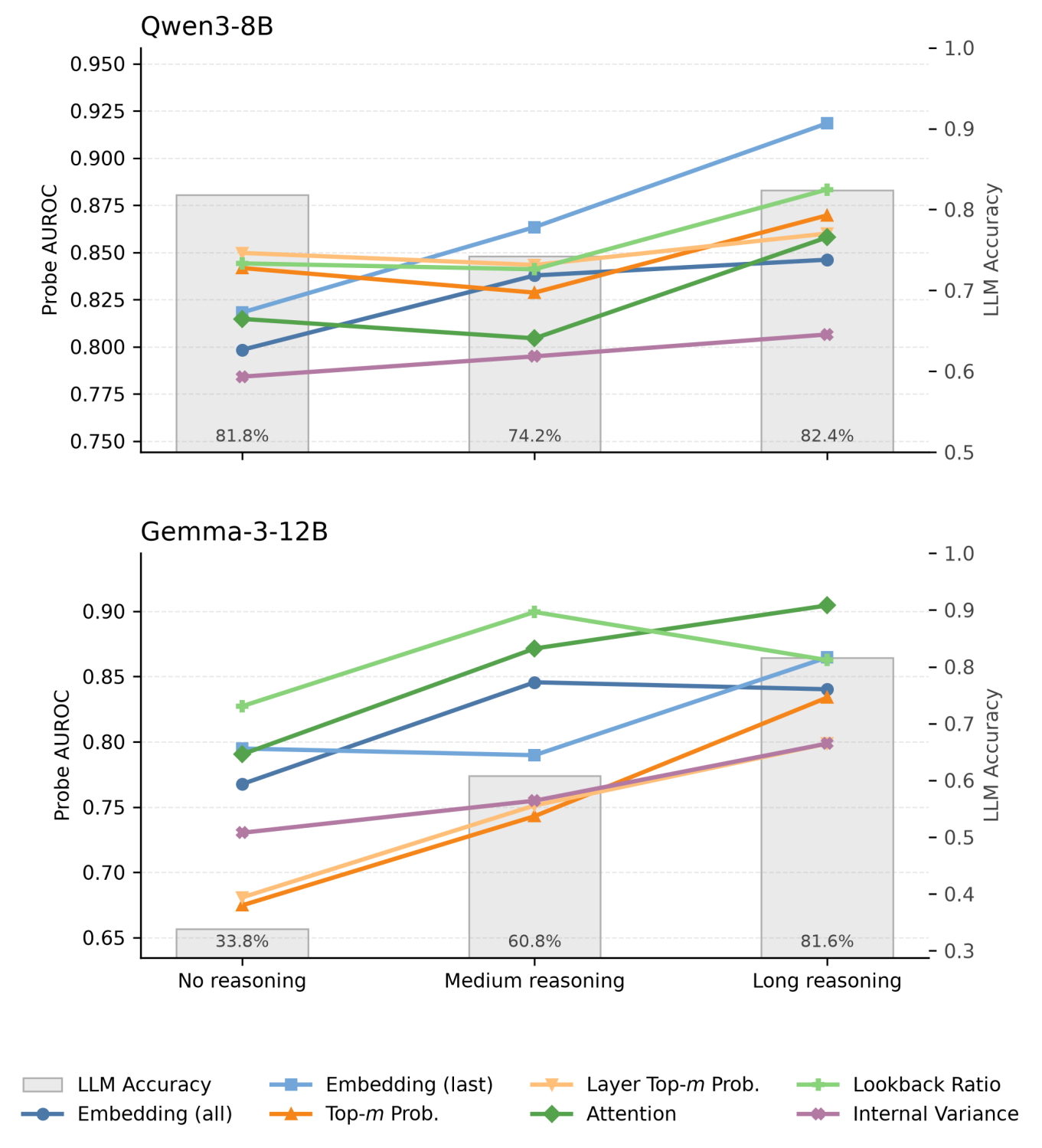}
    \end{subfigure}
    \caption{Effects of inducing reasoning on StrategyQA and MATH-500.}
    \label{fig:prompt_extra}
\end{figure}

\begin{figure}
    \centering
    \includegraphics[width=\linewidth]{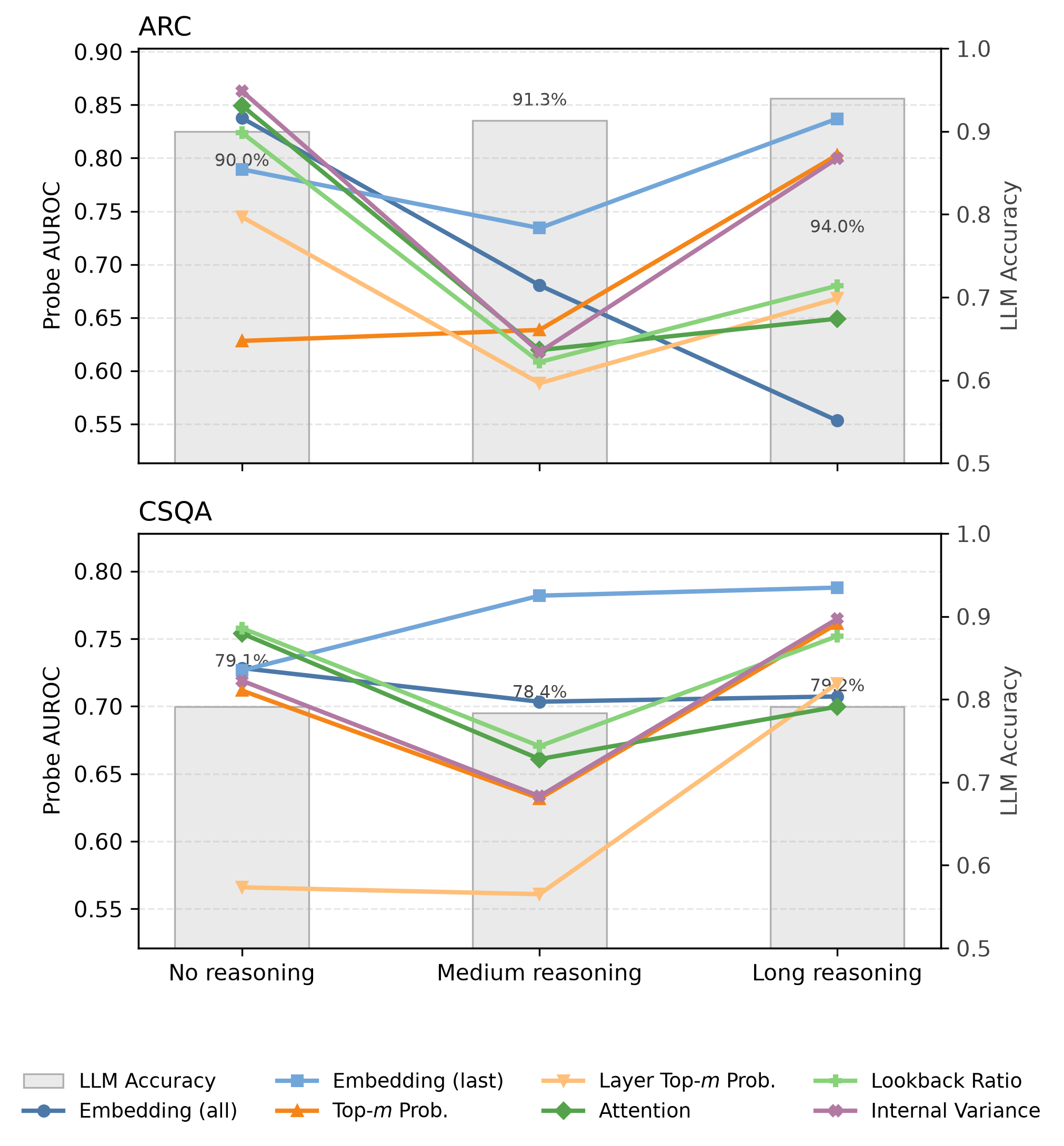}
    \caption{Effects of inducing reasoning with Gemma-3-12B-Instruct.}
    \label{fig:prompt_gemma}
\end{figure}

\paragraph{More Reasoning-Eliciting Experiments.} \label{par:reasoning_controls}
We extend the no-, medium-, and long-reasoning comparison to StrategyQA and MATH-500 using Qwen-3-8B and Gemma-3-12B-Instruct (\Cref{fig:prompt_extra,fig:prompt_gemma}). StrategyQA broadly reproduces the prompt sensitivity observed on ARC and CSQA. On MATH-500, however, reasoning can substantially improve answer accuracy and probe discriminability, in particular, for Gemma, showing that probe performance is prompt-sensitive but not necessarily that reasoning elicitation is intrinsically harmful.

\begin{figure}
    \centering
    \includegraphics[width=\linewidth]{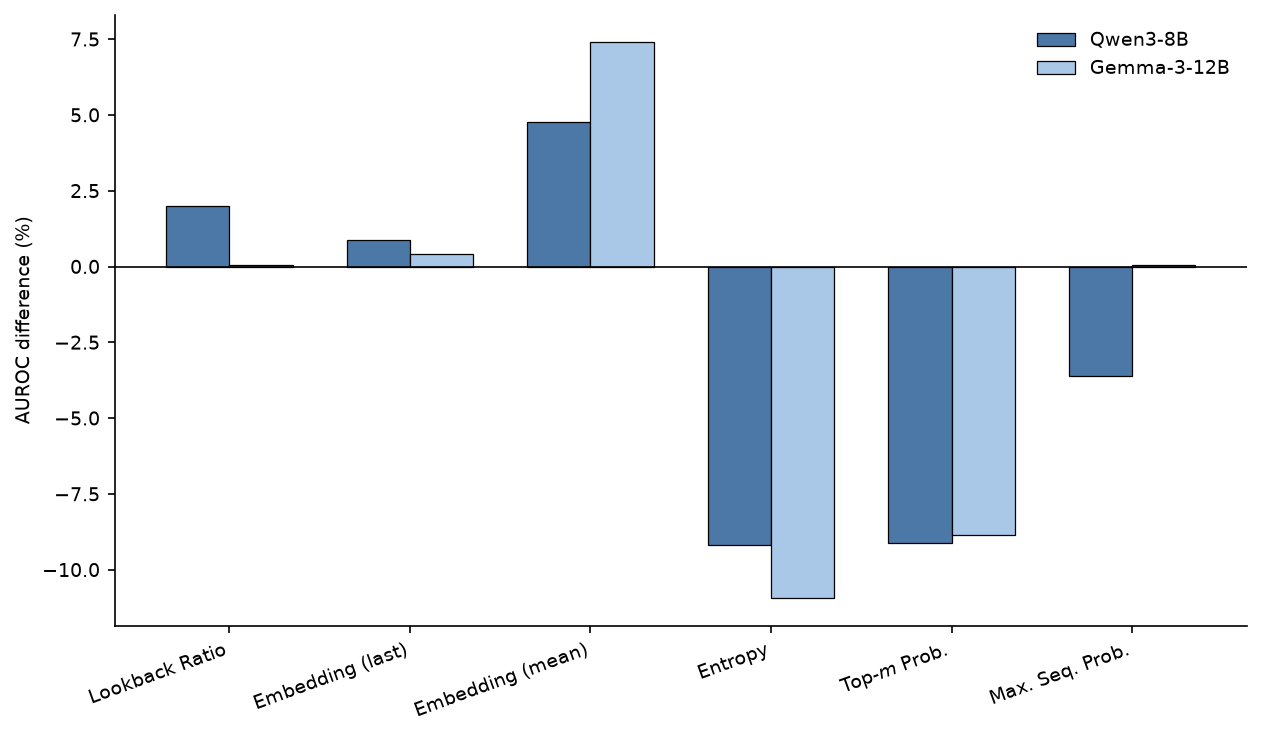}
    \caption{Effect of answer-token-only aggregation. Values report AUROC difference between answer-token-only and full-trace aggregation.}
    \label{fig:aggregation_sqa}
\end{figure}

\begin{figure}
\centering
    \begin{subfigure}[b]{\linewidth}
        \centering
        \includegraphics[width=\linewidth]{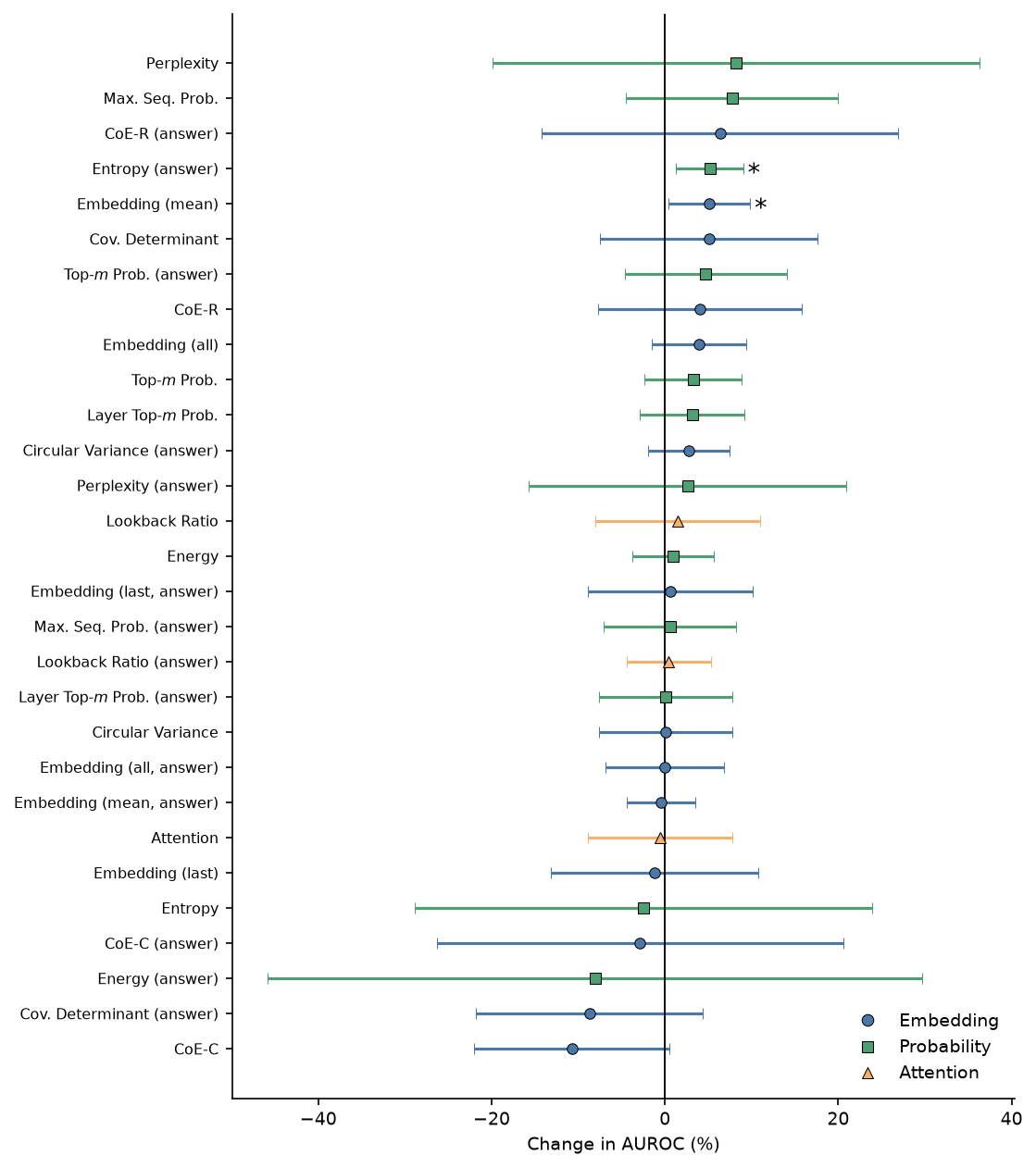}
        \caption{Effect of trimming and re-encoding only reasoning or answer spans. Values report AUROC change relative to the original trace.}
        \label{fig:trimmed_sqa}
    \end{subfigure}
    
    \begin{subfigure}[b]{\linewidth}
        \centering
        \includegraphics[width=\linewidth]{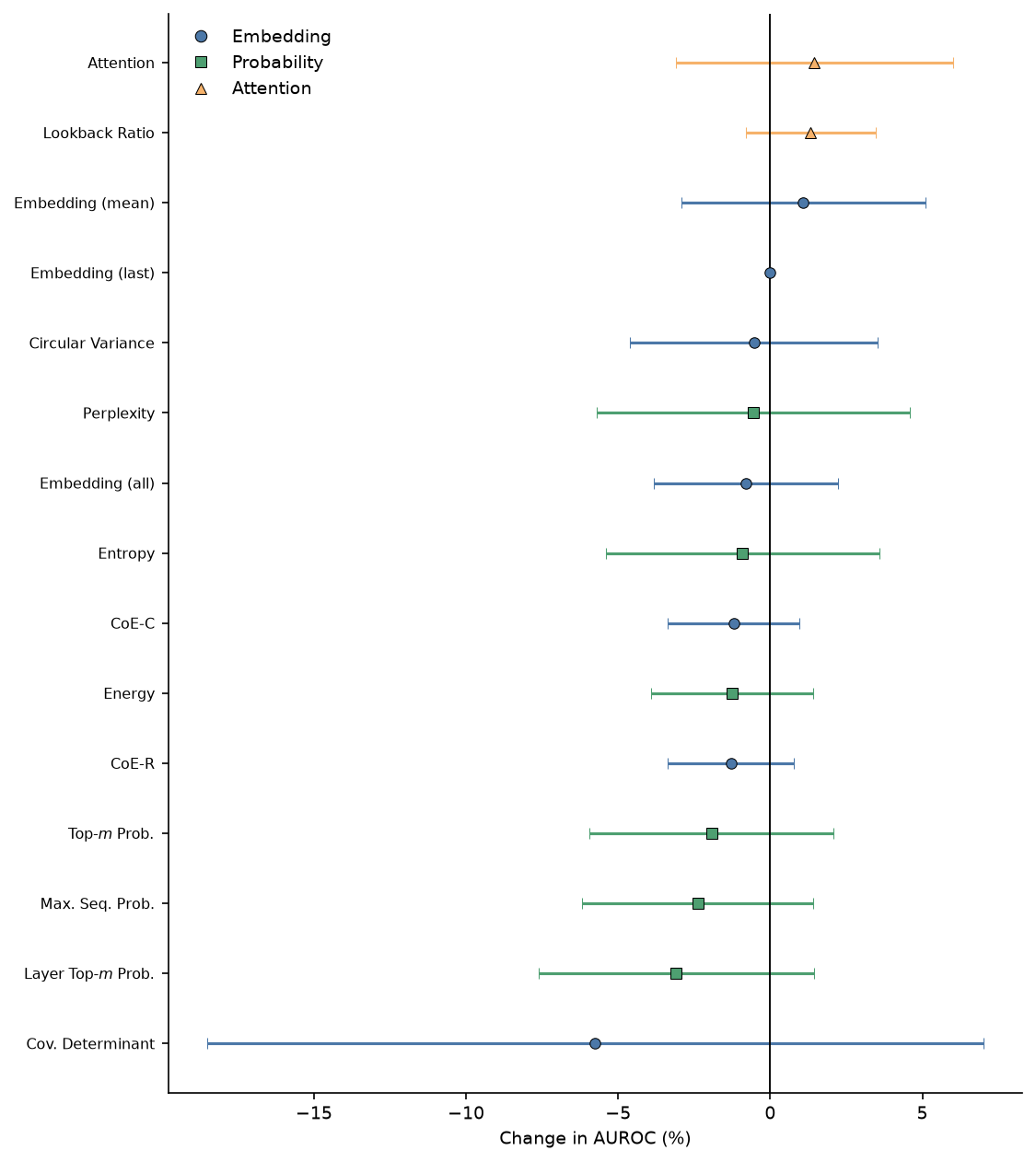}
        \caption{Effect of reasoning token aggregation. Values report AUROC difference between reasoning-token-only and full-trace aggregation.}
        \label{fig:reasoning_agg_sqa}
    \end{subfigure}
    \caption{Reasoning-content and aggregation results.}
        
\end{figure}

\paragraph{Reasoning-Content and Aggregation Controls.}
We further analyse StrategyQA with Gemma-3-12B-Instruct using two complementary controls. First, Gemini-3.1-Flash-Lite identifies sentences that do not contribute to the reasoning. We exclude these spans and re-encode the prompt and shortened trace before recomputing the features; we also consider an answer-only condition that retains only the final answer (\Cref{fig:trimmed_sqa}). This control captures both non-reasoning-content removal and the resulting changes to token context and position.
Second, we retain the original complete trace and its token  states, but aggregate features only over the identified reasoning tokens (\Cref{fig:reasoning_agg_sqa}) and answer tokens (\Cref{fig:aggregation_sqa}). This isolates the aggregation choice without changing the encoded sequence.

Both controls produce mixed, feature-dependent effects. Re-encoding the trimmed trace leads to larger changes for several representations, whereas reasoning-token-only aggregation generally has more modest effects and does occasionally improve performance but not uniformly and is not stable.
Interestingly, answer-token-only aggregation benefits some embedding features but may severely degrade several probability-based features (\Cref{fig:aggregation_sqa}).
These results partially support the dilution hypothesis, but indicate that aggregation alone is insufficient: non-reasoning content may also alter the contextual representations from which uncertainty features are extracted.

We caution that these findings should not be over-interpreted. The reasoning-token spans are identified by an auxiliary LLM and may contain classification errors, and both analyses are post-hoc controls over fixed generations rather than causal interventions on the reasoning process. We therefore treat them as diagnostic evidence for reasoning-content- and aggregation-dependent effects, not as establishing a causal mechanism.

\begin{table*}[t]

\centering
\renewcommand{\arraystretch}{1.25}
\setlength{\tabcolsep}{4.0pt}
\resizebox{\textwidth}{!}{
\begin{tabular}{ll|ccccccccc|c|c}
\toprule
\multicolumn{2}{c|}{\textbf{Method}} & \multirowcell{\textbf{Biographies} \\ \textbf{(ID)}} & \multirowcell{\textbf{Biographies} \\ 
\textbf{(OOD)}} & \textbf{Artworks} & \textbf{Books} & \textbf{Cities} & \textbf{Events} & \textbf{Inventions} & \textbf{Landmarks} & \textbf{Movies} & \multirowcell{\textbf{OOD} \\ \textbf{Mean}} & \multirowcell{\textbf{OOD} \\ \textbf{Rank}} \\
\midrule
\multirow{7}{*}{\rotatebox{90}{Bio-ID-trained}} & \multicolumn{1}{|l|}{Linear (\#64)} & 64.83 \scriptsize ($\pm$ 2.04) & 60.19 \scriptsize ($\pm$ 2.16) & 59.57 \scriptsize ($\pm$ 3.11) & 56.87 \scriptsize ($\pm$ 3.57) & 54.51 \scriptsize ($\pm$ 2.25) & 53.75 \scriptsize ($\pm$ 3.67) & 59.56 \scriptsize ($\pm$ 3.03) & 58.27 \scriptsize ($\pm$ 2.72) & 62.44 \scriptsize ($\pm$2.63) & 58.14 \scriptsize ($\pm$ 1.04) & 16.75 \\
& \multicolumn{1}{|l|}{Linear (\#128)} & 72.43 \scriptsize ($\pm$ 2.01) & 62.61 \scriptsize ($\pm$ 2.10) & 64.89 \scriptsize ($\pm$ 3.01) & 53.53 \scriptsize ($\pm$ 3.74) & 61.8 \scriptsize ($\pm$ 2.18) & 57.22 \scriptsize ($\pm$ 3.91) & 60.81 \scriptsize ($\pm$ 3.1) & 61.24 \scriptsize ($\pm$ 2.82) & 64.45 \scriptsize ($\pm$ 2.47) & 60.82 \scriptsize ($\pm$ 1.05) & 13.38 \\
& \multicolumn{1}{|l|}{MLP (\#128)} & 73.69 \scriptsize ($\pm$  1.91) & 58.65 \scriptsize ($\pm$  2.14) & 62.15 \scriptsize ($\pm$  2.94) & 48.06 \scriptsize ($\pm$  3.48) & 58.97 \scriptsize ($\pm$  2.24) & 54.40 \scriptsize ($\pm$  3.55) & 58.03 \scriptsize ($\pm$  3.15) & 57.51 \scriptsize ($\pm$  2.59) & 60.06 \scriptsize ($\pm$  2.51) & 57.23 \scriptsize ($\pm$ 1.01) & 17.13 \\
& \multicolumn{1}{|l|}{Linear (\#256)} & 78.15 \scriptsize ($\pm$  2.06) & 64.72 \scriptsize ($\pm$  2.04) & 64.72 \scriptsize ($\pm$  2.81) & 56.30 \scriptsize ($\pm$  3.55) & 64.37 \scriptsize ($\pm$  2.13) & 57.30 \scriptsize ($\pm$  3.42) & 61.63 \scriptsize ($\pm$  3.19) & 61.28 \scriptsize ($\pm$  2.68) & 65.55 \scriptsize ($\pm$  2.41) & 61.98 \scriptsize ($\pm$ 1.00) & 12.25 \\
& \multicolumn{1}{|l|}{Lookback Ratio} & 77.06 \scriptsize ($\pm$ 2.15) & 63.72 \scriptsize ($\pm$ 2.26) & 67.14 \scriptsize ($\pm$ 3.05) & 62.53 \scriptsize ($\pm$ 2.97) & 67.08 \scriptsize ($\pm$ 2.18) & 60.60 \scriptsize ($\pm$ 3.40) & 59.73 \scriptsize ($\pm$ 3.11) & 63.28 \scriptsize ($\pm$ 2.60) & 66.65 \scriptsize ($\pm$ 2.30) & 63.84 \scriptsize ($\pm$ 0.98) & 9.25 \\
& \multicolumn{1}{|l|}{Layer Top-$m$ Prob.} & 63.91 \scriptsize ($\pm$ 2.42) & 64.10 \scriptsize ($\pm$ 2.22) & 57.90 \scriptsize ($\pm$ 3.31) & 57.71 \scriptsize ($\pm$ 3.31) & 65.62 \scriptsize ($\pm$ 2.20) & 52.12 \scriptsize ($\pm$ 3.50) & 64.05 \scriptsize ($\pm$ 3.01) &64.39 \scriptsize ($\pm$ 2.64) & 72.25 \scriptsize ($\pm$ 2.39) & 62.27 \scriptsize ($\pm$ 1.01) & 12.63 \\ 
& \multicolumn{1}{|l|}{Internal Variance} & 71.40 \scriptsize ($\pm$ 2.38) & 68.43 \scriptsize ($\pm$ 2.10) & 62.31 \scriptsize ($\pm$ 3.39) & 60.58 \scriptsize ($\pm$ 3.24) & 67.24 \scriptsize ($\pm$ 2.25) & 50.96 \scriptsize ($\pm$ 3.67) & 65.39 \scriptsize ($\pm$ 3.17) & 66.23 \scriptsize ($\pm$ 2.66) & 71.35 \scriptsize ($\pm$ 2.41) & 64.06 \scriptsize ($\pm$ 1.03) & 10.44 \\
\midrule
\multirow{6}{*}{\rotatebox{90}{Training-free}} & \multicolumn{1}{|l|}{Max. Seq. Prob.} & 65.71 \scriptsize ($\pm$ 2.50) & 67.70 \scriptsize ($\pm$ 2.12) & 66.36 \scriptsize ($\pm$ 3.46) & 61.03 \scriptsize ($\pm$ 3.40) & 72.07 \scriptsize ($\pm$ 1.99) & 56.32 \scriptsize ($\pm$ 3.10) & 68.98 \scriptsize ($\pm$ 2.94) & 68.54 \scriptsize ($\pm$ 2.49) & 73.63 \scriptsize ($\pm$ 2.25) & 66.83 \scriptsize ($\pm$ 0.98) & 7.25 \\
& \multicolumn{1}{|l|}{Entropy} & 66.91 \scriptsize ($\pm$ 2.48) & 68.72 \scriptsize ($\pm$ 2.12) & 68.61 \scriptsize ($\pm$ 3.38) & 61.68 \scriptsize ($\pm$ 3.38) & 72.73 \scriptsize ($\pm$ 1.98) & 57.08 \scriptsize ($\pm$ 3.14) & 71.02 \scriptsize ($\pm$ 2.86) & 69.22 \scriptsize ($\pm$ 2.53) & 76.29 \scriptsize ($\pm$ 2.23) & 68.17 \scriptsize ($\pm$ 0.97) & 4.25 \\
& \multicolumn{1}{|l|}{Perplexity} & 66.12 \scriptsize ($\pm$ 2.48) & 68.48 \scriptsize ($\pm$ 2.11) & 67.15 \scriptsize ($\pm$ 3.47) & 61.26 \scriptsize ($\pm$ 3.41) & 72.45 \scriptsize ($\pm$ 2.00) & 55.92 \scriptsize ($\pm$ 3.11) & 69.14 \scriptsize ($\pm$ 2.94) & 68.65 \scriptsize ($\pm$ 2.48) & 74.63 \scriptsize ($\pm$ 2.25) & 67.21 \scriptsize ($\pm$ 0.98) & 6.13 \\
& \multicolumn{1}{|l|}{Energy} & 54.16 \scriptsize ($\pm$ 2.46) & 65.59 \scriptsize ($\pm$ 2.23 & 70.29 \scriptsize ($\pm$ 3.23) & 60.74 \scriptsize ($\pm$ 3.29) & 73.07 \scriptsize ($\pm$ 1.96) & 61.49 \scriptsize ($\pm$ 3.34) & 74.06 \scriptsize ($\pm$ 2.70) & 65.14 \scriptsize ($\pm$ 2.67) & 76.83 \scriptsize ($\pm$ 2.12) & 68.40 \scriptsize ($\pm$ 0.97) & 4.63 \\
& \multicolumn{1}{|l|}{Semantic Entropy} & 71.85 \scriptsize ($\pm$ 2.27) & 61.84 \scriptsize ($\pm$ 2.17) & 64.91 \scriptsize ($\pm$ 2.98) & 67.44 \scriptsize ($\pm$ 2.96) & 68.25 \scriptsize ($\pm$ 2.09) & 62.23 \scriptsize ($\pm$ 3.15) & 66.78 \scriptsize ($\pm$ 2.90) & 72.77 \scriptsize ($\pm$ 2.30) & 70.09 \scriptsize ($\pm$ 2.21) & 66.79 \scriptsize ($\pm$ 0.93) & 6.50 \\
& \multicolumn{1}{|l|}{Eigenscore} & 62.25 \scriptsize ($\pm$ 2.53) & 57.73 \scriptsize ($\pm$ 2.25) & 58.37 \scriptsize ($\pm$ 3.20) & 58.33 \scriptsize ($\pm$ 3.25) & 57.24 \scriptsize ($\pm$ 2.37) & 48.77 \scriptsize ($\pm$ 3.42) & 58.94 \scriptsize ($\pm$ 3.03) & 59.92 \scriptsize ($\pm$ 2.73) & 68.31 \scriptsize ($\pm$ 2.31) & 58.45 \scriptsize ($\pm$ 1.01) & 16.50 \\
\midrule
\multirow{6}{*}{\rotatebox{90}{Pretrained}} & \multicolumn{1}{|l|}{Embedding (last)} & 54.85 \scriptsize ($\pm$  2.42) & 60.51 \scriptsize ($\pm$  2.12) & 52.11 \scriptsize ($\pm$  2.97) & 57.17 \scriptsize ($\pm$  3.39) & 54.68 \scriptsize ($\pm$  2.49) & 50.93 \scriptsize ($\pm$  3.68) & 59.29 \scriptsize ($\pm$  3.08) & 61.01 \scriptsize ($\pm$  2.59) & 66.25 \scriptsize ($\pm$  2.41) & 57.74 \scriptsize ($\pm$ 1.02) & 16.63 \\
& \multicolumn{1}{|l|}{Lookback Ratio} & 64.02 \scriptsize ($\pm$  2.32) & 61.68 \scriptsize ($\pm$  2.14) & 63.12 \scriptsize ($\pm$  2.81) & 59.87 \scriptsize ($\pm$  2.97) & 58.23 \scriptsize ($\pm$  2.30) & 56.60 \scriptsize ($\pm$  3.80) & 57.08 \scriptsize ($\pm$  3.20) & 62.72 \scriptsize ($\pm$  2.44) & 71.97 \scriptsize ($\pm$  2.31) & 61.41 \scriptsize ($\pm$ 0.99) & 13.38 \\
& \multicolumn{1}{|l|}{Layer Top-$m$ Prob.} & 66.16 \scriptsize ($\pm$  2.26) & 68.60 \scriptsize ($\pm$  1.99) & 66.44 \scriptsize ($\pm$  3.07) & 64.11 \scriptsize ($\pm$  3.56) & 66.48 \scriptsize ($\pm$  2.14) & 61.77 \scriptsize ($\pm$  3.29) & 72.96 \scriptsize ($\pm$  3.03) & 69.31 \scriptsize ($\pm$  2.48) & 80.14 \scriptsize ($\pm$  1.93) & 68.73 \scriptsize ($\pm$ 0.97) & 3.88 \\
& \multicolumn{1}{|l|}{Internal Variance} &  67.27 \scriptsize ($\pm$  2.27) & 70.19 \scriptsize ($\pm$  1.96) & 68.49 \scriptsize ($\pm$  2.96) & 64.46 \scriptsize ($\pm$  3.34) & 67.24 \scriptsize ($\pm$  2.09) & 59.28 \scriptsize ($\pm$  3.48) & 69.90 \scriptsize ($\pm$  3.12) & 67.74 \scriptsize ($\pm$  2.57) & 77.80 \scriptsize ($\pm$  2.05) & 68.14 \scriptsize ($\pm$ 0.97) & 4.31 \\
& \multicolumn{1}{|l|}{Ensemble (probe)} & 73.76 \scriptsize ($\pm$  2.03) & 70.18 \scriptsize ($\pm$  2.00) & 63.72 \scriptsize ($\pm$  3.20) & 58.88 \scriptsize ($\pm$  3.42) & 65.67 \scriptsize ($\pm$  2.26) & 55.00 \scriptsize ($\pm$  3.57) & 68.08 \scriptsize ($\pm$  3.36) & 64.46 \scriptsize ($\pm$  2.76) & 73.17 \scriptsize ($\pm$  2.43) & 64.90 \scriptsize ($\pm$ 1.04) & 9.50 \\ 
& \multicolumn{1}{|l|}{Ensemble (task)} & 68.38 \scriptsize ($\pm$  2.27) & 70.85 \scriptsize ($\pm$  1.86) & 65.38 \scriptsize ($\pm$  3.02) & 64.53 \scriptsize ($\pm$  3.25) & 63.44 \scriptsize ($\pm$  2.24) & 56.79 \scriptsize ($\pm$  3.64) & 70.23 \scriptsize ($\pm$  3.06) & 69.25 \scriptsize ($\pm$  2.46) & 77.93 \scriptsize ($\pm$  2.03) & 67.30 \scriptsize ($\pm$ 0.98) & 5.25 \\
\bottomrule
\end{tabular}

}
\caption{AUROC with bootstrapped standard errors (Qwen-3-8B). Baselines are indicated with probe architecture and number of training examples from Biographies (ID).}
\label{tab:pretrained}
\end{table*}

\begin{table*}[t]

\centering
\renewcommand{\arraystretch}{1.25}
\setlength{\tabcolsep}{4.0pt}
\resizebox{\textwidth}{!}{
\begin{tabular}{ll|ccccccccc|c|c}
\toprule
\multicolumn{2}{c|}{\textbf{Method}}
& \multirowcell{\textbf{Biographies} \\ \textbf{(ID)}}
& \multirowcell{\textbf{Biographies} \\ \textbf{(OOD)}}
& \textbf{Artworks}
& \textbf{Books}
& \textbf{Cities}
& \textbf{Events}
& \textbf{Inventions}
& \textbf{Landmarks}
& \textbf{Movies}
& \multirowcell{\textbf{OOD} \\ \textbf{Mean}}
& \multirowcell{\textbf{OOD} \\ \textbf{Rank}} \\
\midrule

\multirow{4}{*}{\rotatebox{90}{Bio-ID-trained}}
& \multicolumn{1}{|l|}{Embedding (last)}
& 74.08 \scriptsize ($\pm$ 1.86)
& 55.92 \scriptsize ($\pm$ 2.64)
& 55.47 \scriptsize ($\pm$ 3.14)
& 68.40 \scriptsize ($\pm$ 4.95)
& 60.60 \scriptsize ($\pm$ 2.74)
& 52.78 \scriptsize ($\pm$ 4.77)
& 64.38 \scriptsize ($\pm$ 3.43)
& 53.97 \scriptsize ($\pm$ 3.75)
& 63.47 \scriptsize ($\pm$ 4.32)
& 59.37 \scriptsize ($\pm$ 1.35)
& 9.50 \\

& \multicolumn{1}{|l|}{Lookback Ratio}
& 75.62 \scriptsize ($\pm$ 1.90)
& 53.41 \scriptsize ($\pm$ 2.63)
& 61.96 \scriptsize ($\pm$ 3.21)
& 46.69 \scriptsize ($\pm$ 5.31)
& 56.53 \scriptsize ($\pm$ 3.10)
& 53.51 \scriptsize ($\pm$ 5.36)
& 51.92 \scriptsize ($\pm$ 4.18)
& 52.39 \scriptsize ($\pm$ 3.91)
& 61.94 \scriptsize ($\pm$ 4.17)
& 54.79 \scriptsize ($\pm$ 1.45)
& 12.13 \\

& \multicolumn{1}{|l|}{Layer Top-$m$ Prob.}
& 63.80 \scriptsize ($\pm$ 2.23)
& 60.19 \scriptsize ($\pm$ 2.45)
& 51.70 \scriptsize ($\pm$ 3.34)
& 43.25 \scriptsize ($\pm$ 5.56)
& 57.28 \scriptsize ($\pm$ 2.81)
& 55.81 \scriptsize ($\pm$ 4.72)
& 72.62 \scriptsize ($\pm$ 3.02)
& 49.44 \scriptsize ($\pm$ 3.80)
& 57.43 \scriptsize ($\pm$ 4.01)
& 55.97 \scriptsize ($\pm$ 1.36)
& 11.88 \\

& \multicolumn{1}{|l|}{Internal Variance}
& 69.30 \scriptsize ($\pm$ 2.06)
& 61.34 \scriptsize ($\pm$ 2.58)
& 55.97 \scriptsize ($\pm$ 3.44)
& 51.30 \scriptsize ($\pm$ 4.68)
& 60.13 \scriptsize ($\pm$ 2.58)
& 54.44 \scriptsize ($\pm$ 4.87)
& 68.96 \scriptsize ($\pm$ 3.29)
& 54.36 \scriptsize ($\pm$ 3.54)
& 60.65 \scriptsize ($\pm$ 4.02)
& 58.39 \scriptsize ($\pm$ 1.31)
& 10.25 \\

\midrule

\multirow{6}{*}{\rotatebox{90}{Training-free}}
& \multicolumn{1}{|l|}{Max. Seq. Prob.}
& 67.24 \scriptsize ($\pm$ 2.15)
& 62.09 \scriptsize ($\pm$ 2.56)
& 56.38 \scriptsize ($\pm$ 3.36)
& 52.12 \scriptsize ($\pm$ 4.55)
& 60.34 \scriptsize ($\pm$ 2.58)
& 58.05 \scriptsize ($\pm$ 4.16)
& 69.53 \scriptsize ($\pm$ 3.24)
& 56.50 \scriptsize ($\pm$ 3.66)
& 62.38 \scriptsize ($\pm$ 3.49)
& 59.67 \scriptsize ($\pm$ 1.24)
& 7.50 \\

& \multicolumn{1}{|l|}{Entropy}
& 69.09 \scriptsize ($\pm$ 2.09)
& 62.67 \scriptsize ($\pm$ 2.59)
& 57.31 \scriptsize ($\pm$ 3.40)
& 52.35 \scriptsize ($\pm$ 4.54)
& 60.72 \scriptsize ($\pm$ 2.57)
& 59.00 \scriptsize ($\pm$ 4.38)
& 72.40 \scriptsize ($\pm$ 3.05)
& 56.44 \scriptsize ($\pm$ 3.69)
& 63.84 \scriptsize ($\pm$ 3.55)
& 60.59 \scriptsize ($\pm$ 1.25)
& 5.00 \\

& \multicolumn{1}{|l|}{Perplexity}
& 67.31 \scriptsize ($\pm$ 2.15)
& 61.84 \scriptsize ($\pm$ 2.56)
& 56.18 \scriptsize ($\pm$ 3.36)
& 52.40 \scriptsize ($\pm$ 4.66)
& 60.46 \scriptsize ($\pm$ 2.61)
& 58.49 \scriptsize ($\pm$ 4.22)
& 69.59 \scriptsize ($\pm$ 3.24)
& 56.42 \scriptsize ($\pm$ 3.67)
& 62.27 \scriptsize ($\pm$ 3.52)
& 59.71 \scriptsize ($\pm$ 1.25)
& 7.50 \\

& \multicolumn{1}{|l|}{Energy}
& 66.73 \scriptsize ($\pm$ 2.17)
& 57.64 \scriptsize ($\pm$ 2.52)
& 57.46 \scriptsize ($\pm$ 3.21)
& 66.46 \scriptsize ($\pm$ 4.76)
& 62.10 \scriptsize ($\pm$ 2.80)
& 59.66 \scriptsize ($\pm$ 4.67)
& 67.12 \scriptsize ($\pm$ 3.58)
& 66.85 \scriptsize ($\pm$ 3.54)
& 63.98 \scriptsize ($\pm$ 3.88)
& 62.66 \scriptsize ($\pm$ 1.31)
& 4.75 \\

& \multicolumn{1}{|l|}{Semantic Entropy}
& 72.55 \scriptsize ($\pm$ 1.96)
& 59.32 \scriptsize ($\pm$ 2.40)
& 56.73 \scriptsize ($\pm$ 3.17)
& 58.34 \scriptsize ($\pm$ 5.00)
& 66.70 \scriptsize ($\pm$ 2.45)
& 58.41 \scriptsize ($\pm$ 4.47)
& 67.33 \scriptsize ($\pm$ 3.42)
& 61.21 \scriptsize ($\pm$ 3.43)
& 72.71 \scriptsize ($\pm$ 3.55)
& 62.59 \scriptsize ($\pm$ 1.27)
& 5.25 \\

& \multicolumn{1}{|l|}{Eigenscore}
& 65.21 \scriptsize ($\pm$ 1.94)
& 58.04 \scriptsize ($\pm$ 2.39)
& 55.02 \scriptsize ($\pm$ 3.00)
& 50.66 \scriptsize ($\pm$ 4.79)
& 57.47 \scriptsize ($\pm$ 2.71)
& 49.79 \scriptsize ($\pm$ 4.51)
& 59.56 \scriptsize ($\pm$ 3.44)
& 52.60 \scriptsize ($\pm$ 3.39)
& 60.00 \scriptsize ($\pm$ 4.19)
& 55.39 \scriptsize ($\pm$ 1.29)
& 13.25 \\

\midrule

\multirow{6}{*}{\rotatebox{90}{Pretrained}}
& \multicolumn{1}{|l|}{Embedding (last)}
& 55.83 \scriptsize ($\pm$ 2.04)
& 51.99 \scriptsize ($\pm$ 2.43)
& 60.80 \scriptsize ($\pm$ 3.26)
& 56.11 \scriptsize ($\pm$ 5.51)
& 53.09 \scriptsize ($\pm$ 2.76)
& 58.09 \scriptsize ($\pm$ 4.78)
& 63.75 \scriptsize ($\pm$ 3.40)
& 58.28 \scriptsize ($\pm$ 3.77)
& 67.31 \scriptsize ($\pm$ 3.85)
& 58.68 \scriptsize ($\pm$ 1.36)
& 8.00 \\

& \multicolumn{1}{|l|}{Lookback Ratio}
& 60.30 \scriptsize ($\pm$ 1.89)
& 50.34 \scriptsize ($\pm$ 2.64)
& 49.78 \scriptsize ($\pm$ 3.29)
& 50.13 \scriptsize ($\pm$ 5.51)
& 49.29 \scriptsize ($\pm$ 2.84)
& 43.85 \scriptsize ($\pm$ 4.87)
& 48.75 \scriptsize ($\pm$ 3.77)
& 48.65 \scriptsize ($\pm$ 3.77)
& 48.07 \scriptsize ($\pm$ 3.86)
& 48.61 \scriptsize ($\pm$ 1.39)
& 15.75 \\

& \multicolumn{1}{|l|}{Layer Top-$m$ Prob.}
& 71.13 \scriptsize ($\pm$ 1.71)
& 61.37 \scriptsize ($\pm$ 2.55)
& 56.92 \scriptsize ($\pm$ 3.40)
& 53.81 \scriptsize ($\pm$ 4.45)
& 59.90 \scriptsize ($\pm$ 2.60)
& 61.34 \scriptsize ($\pm$ 4.33)
& 71.74 \scriptsize ($\pm$ 2.95)
& 56.72 \scriptsize ($\pm$ 3.63)
& 63.05 \scriptsize ($\pm$ 3.69)
& 60.61 \scriptsize ($\pm$ 1.24)
& 5.88 \\

& \multicolumn{1}{|l|}{Internal Variance}
& 70.95 \scriptsize ($\pm$ 1.70)
& 61.87 \scriptsize ($\pm$ 2.59)
& 56.27 \scriptsize ($\pm$ 3.41)
& 51.28 \scriptsize ($\pm$ 4.72)
& 60.76 \scriptsize ($\pm$ 2.53)
& 55.04 \scriptsize ($\pm$ 4.69)
& 69.79 \scriptsize ($\pm$ 3.22)
& 56.14 \scriptsize ($\pm$ 3.48)
& 61.52 \scriptsize ($\pm$ 3.90)
& 59.08 \scriptsize ($\pm$ 1.29)
& 8.25 \\

& \multicolumn{1}{|l|}{Ensemble (Probe)}
& 70.20 \scriptsize ($\pm$ 1.72)
& 61.45 \scriptsize ($\pm$ 2.61)
& 57.45 \scriptsize ($\pm$ 3.41)
& 52.42 \scriptsize ($\pm$ 4.23)
& 59.92 \scriptsize ($\pm$ 2.64)
& 59.07 \scriptsize ($\pm$ 4.65)
& 71.76 \scriptsize ($\pm$ 3.08)
& 54.00 \scriptsize ($\pm$ 3.68)
& 63.10 \scriptsize ($\pm$ 3.78)
& 59.90 \scriptsize ($\pm$ 1.26)
& 6.50 \\

& \multicolumn{1}{|l|}{Ensemble (Task)}
& 70.31 \scriptsize ($\pm$ 1.81)
& 60.50 \scriptsize ($\pm$ 2.56)
& 61.38 \scriptsize ($\pm$ 3.30)
& 53.53 \scriptsize ($\pm$ 4.78)
& 57.77 \scriptsize ($\pm$ 2.59)
& 59.65 \scriptsize ($\pm$ 4.34)
& 72.67 \scriptsize ($\pm$ 3.11)
& 59.90 \scriptsize ($\pm$ 3.48)
& 66.36 \scriptsize ($\pm$ 3.45)
& 61.47 \scriptsize ($\pm$ 1.25)
& 4.63 \\

\bottomrule
\end{tabular}

}
\caption{AUROC with bootstrapped standard errors (Gemma-3-12B-Instruct).}
\label{tab:pretrained_gemma}
\end{table*}

\begin{table*}[t]

\centering
\renewcommand{\arraystretch}{1.25}
\setlength{\tabcolsep}{4.0pt}
\resizebox{\textwidth}{!}{
\begin{tabular}{ll|ccccccccc|c|c}
\toprule
\multicolumn{2}{c|}{\textbf{Method}}
& \multirowcell{\textbf{Biographies} \\ \textbf{(ID)}}
& \multirowcell{\textbf{Biographies} \\ \textbf{(OOD)}}
& \textbf{Artworks}
& \textbf{Books}
& \textbf{Cities}
& \textbf{Events}
& \textbf{Inventions}
& \textbf{Landmarks}
& \textbf{Movies}
& \multirowcell{\textbf{OOD} \\ \textbf{Mean}}
& \multirowcell{\textbf{OOD} \\ \textbf{Rank}} \\
\midrule

\multirow{4}{*}{\rotatebox{90}{Bio-ID-trained}}
& \multicolumn{1}{|l|}{Embedding (last)}
& 66.11 \scriptsize ($\pm$ 2.38)
& 55.66 \scriptsize ($\pm$ 2.31)
& 59.63 \scriptsize ($\pm$ 2.83)
& 54.12 \scriptsize ($\pm$ 3.11)
& 59.57 \scriptsize ($\pm$ 2.41)
& 53.85 \scriptsize ($\pm$ 3.03)
& 55.53 \scriptsize ($\pm$ 2.47)
& 54.20 \scriptsize ($\pm$ 2.55)
& 57.02 \scriptsize ($\pm$ 2.76)
& 56.20 \scriptsize ($\pm$ 0.95)
& 14.44 \\

& \multicolumn{1}{|l|}{Lookback Ratio}
& 70.26 \scriptsize ($\pm$ 2.23)
& 63.62 \scriptsize ($\pm$ 2.16)
& 69.13 \scriptsize ($\pm$ 2.73)
& 58.16 \scriptsize ($\pm$ 3.02)
& 66.48 \scriptsize ($\pm$ 2.19)
& 55.99 \scriptsize ($\pm$ 2.83)
& 62.93 \scriptsize ($\pm$ 2.44)
& 64.58 \scriptsize ($\pm$ 2.22)
& 70.47 \scriptsize ($\pm$ 2.37)
& 63.92 \scriptsize ($\pm$ 0.89)
& 8.12 \\

& \multicolumn{1}{|l|}{Layer Top-$m$ Prob.}
& 64.45 \scriptsize ($\pm$ 2.50)
& 56.17 \scriptsize ($\pm$ 2.33)
& 51.26 \scriptsize ($\pm$ 3.32)
& 54.04 \scriptsize ($\pm$ 3.19)
& 51.44 \scriptsize ($\pm$ 2.35)
& 46.80 \scriptsize ($\pm$ 3.14)
& 59.20 \scriptsize ($\pm$ 2.55)
& 51.70 \scriptsize ($\pm$ 2.86)
& 62.97 \scriptsize ($\pm$ 3.07)
& 54.20 \scriptsize ($\pm$ 1.02)
& 14.88 \\

& \multicolumn{1}{|l|}{Internal Variance}
& 77.40 \scriptsize ($\pm$ 2.02)
& 64.87 \scriptsize ($\pm$ 2.08)
& 61.51 \scriptsize ($\pm$ 3.15)
& 61.18 \scriptsize ($\pm$ 3.10)
& 69.38 \scriptsize ($\pm$ 2.22)
& 59.85 \scriptsize ($\pm$ 3.32)
& 69.01 \scriptsize ($\pm$ 2.37)
& 65.99 \scriptsize ($\pm$ 2.40)
& 76.86 \scriptsize ($\pm$ 2.41)
& 66.08 \scriptsize ($\pm$ 0.94)
& 5.12 \\

\midrule

\multirow{6}{*}{\rotatebox{90}{Training-free}}
& \multicolumn{1}{|l|}{Max. Seq. Prob.}
& 75.39 \scriptsize ($\pm$ 2.04)
& 63.53 \scriptsize ($\pm$ 2.00)
& 64.52 \scriptsize ($\pm$ 2.68)
& 59.12 \scriptsize ($\pm$ 3.21)
& 67.34 \scriptsize ($\pm$ 2.19)
& 59.49 \scriptsize ($\pm$ 3.17)
& 69.25 \scriptsize ($\pm$ 2.21)
& 63.25 \scriptsize ($\pm$ 2.47)
& 70.43 \scriptsize ($\pm$ 2.72)
& 64.62 \scriptsize ($\pm$ 0.92)
& 7.25 \\

& \multicolumn{1}{|l|}{Entropy}
& 75.58 \scriptsize ($\pm$ 2.04)
& 65.10 \scriptsize ($\pm$ 2.01)
& 66.75 \scriptsize ($\pm$ 2.66)
& 60.90 \scriptsize ($\pm$ 3.27)
& 68.73 \scriptsize ($\pm$ 2.19)
& 59.24 \scriptsize ($\pm$ 3.23)
& 70.70 \scriptsize ($\pm$ 2.22)
& 65.14 \scriptsize ($\pm$ 2.48)
& 72.73 \scriptsize ($\pm$ 2.65)
& 66.16 \scriptsize ($\pm$ 0.93)
& 4.50 \\

& \multicolumn{1}{|l|}{Perplexity}
& 75.25 \scriptsize ($\pm$ 2.03)
& 64.40 \scriptsize ($\pm$ 2.01)
& 64.43 \scriptsize ($\pm$ 2.73)
& 59.51 \scriptsize ($\pm$ 3.18)
& 67.55 \scriptsize ($\pm$ 2.20)
& 60.20 \scriptsize ($\pm$ 3.20)
& 69.21 \scriptsize ($\pm$ 2.21)
& 63.21 \scriptsize ($\pm$ 2.48)
& 70.98 \scriptsize ($\pm$ 2.70)
& 64.94 \scriptsize ($\pm$ 0.93)
& 6.44 \\

& \multicolumn{1}{|l|}{Energy}
& 72.03 \scriptsize ($\pm$ 2.22)
& 60.01 \scriptsize ($\pm$ 2.13)
& 62.36 \scriptsize ($\pm$ 2.97)
& 56.42 \scriptsize ($\pm$ 3.23)
& 67.30 \scriptsize ($\pm$ 2.21)
& 54.86 \scriptsize ($\pm$ 3.23)
& 70.07 \scriptsize ($\pm$ 2.25)
& 54.20 \scriptsize ($\pm$ 2.63)
& 67.49 \scriptsize ($\pm$ 2.88)
& 61.59 \scriptsize ($\pm$ 0.96)
& 10.69 \\

& \multicolumn{1}{|l|}{Semantic Entropy}
& 69.17 \scriptsize ($\pm$ 2.27)
& 68.16 \scriptsize ($\pm$ 2.02)
& 63.79 \scriptsize ($\pm$ 2.82)
& 59.35 \scriptsize ($\pm$ 2.72)
& 66.40 \scriptsize ($\pm$ 2.24)
& 59.09 \scriptsize ($\pm$ 2.82)
& 60.67 \scriptsize ($\pm$ 2.43)
& 59.83 \scriptsize ($\pm$ 2.56)
& 69.81 \scriptsize ($\pm$ 2.45)
& 63.39 \scriptsize ($\pm$ 0.89)
& 8.62 \\

& \multicolumn{1}{|l|}{Eigenscore}
& 66.81 \scriptsize ($\pm$ 2.25)
& 65.57 \scriptsize ($\pm$ 2.15)
& 63.96 \scriptsize ($\pm$ 2.77)
& 58.58 \scriptsize ($\pm$ 2.93)
& 63.73 \scriptsize ($\pm$ 2.26)
& 52.32 \scriptsize ($\pm$ 2.97)
& 62.65 \scriptsize ($\pm$ 2.53)
& 55.50 \scriptsize ($\pm$ 2.48)
& 68.97 \scriptsize ($\pm$ 2.42)
& 61.41 \scriptsize ($\pm$ 0.91)
& 10.50 \\

\midrule

\multirow{6}{*}{\rotatebox{90}{Pretrained}}
& \multicolumn{1}{|l|}{Embedding (last)}
& 55.09 \scriptsize ($\pm$ 2.54)
& 52.95 \scriptsize ($\pm$ 2.32)
& 46.70 \scriptsize ($\pm$ 3.22)
& 52.82 \scriptsize ($\pm$ 2.93)
& 60.08 \scriptsize ($\pm$ 2.32)
& 54.01 \scriptsize ($\pm$ 3.29)
& 56.94 \scriptsize ($\pm$ 2.55)
& 58.07 \scriptsize ($\pm$ 2.55)
& 59.02 \scriptsize ($\pm$ 2.77)
& 55.07 \scriptsize ($\pm$ 0.98)
& 14.38 \\

& \multicolumn{1}{|l|}{Lookback Ratio}
& 62.63 \scriptsize ($\pm$ 2.38)
& 67.40 \scriptsize ($\pm$ 1.91)
& 66.13 \scriptsize ($\pm$ 2.75)
& 62.03 \scriptsize ($\pm$ 3.01)
& 54.74 \scriptsize ($\pm$ 2.34)
& 55.45 \scriptsize ($\pm$ 2.86)
& 50.63 \scriptsize ($\pm$ 2.51)
& 65.61 \scriptsize ($\pm$ 2.25)
& 75.23 \scriptsize ($\pm$ 2.49)
& 62.15 \scriptsize ($\pm$ 0.90)
& 7.25 \\

& \multicolumn{1}{|l|}{Layer Top-$m$ Prob.}
& 71.06 \scriptsize ($\pm$ 2.21)
& 60.45 \scriptsize ($\pm$ 2.17)
& 58.02 \scriptsize ($\pm$ 3.10)
& 56.55 \scriptsize ($\pm$ 3.24)
& 64.87 \scriptsize ($\pm$ 2.19)
& 57.23 \scriptsize ($\pm$ 2.93)
& 62.75 \scriptsize ($\pm$ 2.40)
& 60.83 \scriptsize ($\pm$ 2.60)
& 66.09 \scriptsize ($\pm$ 3.00)
& 60.85 \scriptsize ($\pm$ 0.97)
& 11.38 \\

& \multicolumn{1}{|l|}{Internal Variance}
& 77.50 \scriptsize ($\pm$ 2.02)
& 64.92 \scriptsize ($\pm$ 2.08)
& 60.94 \scriptsize ($\pm$ 3.13)
& 61.50 \scriptsize ($\pm$ 3.07)
& 69.20 \scriptsize ($\pm$ 2.24)
& 60.38 \scriptsize ($\pm$ 3.31)
& 69.04 \scriptsize ($\pm$ 2.38)
& 65.86 \scriptsize ($\pm$ 2.40)
& 75.84 \scriptsize ($\pm$ 2.53)
& 65.96 \scriptsize ($\pm$ 0.95)
& 4.88 \\

& \multicolumn{1}{|l|}{Ensemble (probe)}
& 77.16 \scriptsize ($\pm$ 1.99)
& 64.99 \scriptsize ($\pm$ 2.02)
& 64.23 \scriptsize ($\pm$ 2.93)
& 62.18 \scriptsize ($\pm$ 3.13)
& 69.55 \scriptsize ($\pm$ 2.20)
& 60.20 \scriptsize ($\pm$ 3.26)
& 70.21 \scriptsize ($\pm$ 2.29)
& 66.05 \scriptsize ($\pm$ 2.45)
& 73.99 \scriptsize ($\pm$ 2.62)
& 66.43 \scriptsize ($\pm$ 0.94)
& 3.31 \\

& \multicolumn{1}{|l|}{Ensemble (task)}
& 75.64 \scriptsize ($\pm$ 2.06)
& 65.74 \scriptsize ($\pm$ 1.96)
& 64.88 \scriptsize ($\pm$ 2.74)
& 62.84 \scriptsize ($\pm$ 2.96)
& 68.71 \scriptsize ($\pm$ 2.14)
& 59.17 \scriptsize ($\pm$ 3.05)
& 66.75 \scriptsize ($\pm$ 2.33)
& 67.29 \scriptsize ($\pm$ 2.30)
& 72.75 \scriptsize ($\pm$ 2.57)
& 66.02 \scriptsize ($\pm$ 0.90)
& 4.25 \\

\bottomrule
\end{tabular}

}
\caption{AUROC with bootstrapped standard errors (Llama-3.1-8B-Instruct).}
\label{tab:pretrained_llama}
\end{table*}

\begin{table*}[t]
\centering
\setlength{\tabcolsep}{5pt}
\renewcommand{\arraystretch}{1.25}
\resizebox{\textwidth}{!}{%
\begin{tabular}{ll|ccccccccc|c}
\toprule
\multicolumn{2}{c}{\textbf{Probe}} &
\multirowcell{\textbf{Biographies}\\ \textbf{(ID)}} &
\multirowcell{\textbf{Biographies}\\ \textbf{(OOD)}} &
\textbf{Artworks} &
\textbf{Books} &
\textbf{Cities} &
\textbf{Events} &
\textbf{Inventions} &
\textbf{Landmarks} &
\textbf{Movies} &
\multirowcell{\textbf{OOD}\\\textbf{Mean}}  \\

\midrule
\rowcolor{black!8}
\multicolumn{12}{c}{\textbf{Surrogate: Qwen3-4B-Instruct}} \\

\multirow{4}{*}{\rotatebox{90}{Bio-ID-trained}}
& \multicolumn{1}{|l|}{Embedding (Last)}
& 75.48 {\scriptsize($\pm$2.19)}
& 59.14 {\scriptsize($\pm$2.33)}
& 56.75 {\scriptsize($\pm$3.27)}
& 60.36 {\scriptsize($\pm$3.29)}
& 57.86 {\scriptsize($\pm$2.23)}
& 54.91 {\scriptsize($\pm$3.30)}
& 57.33 {\scriptsize($\pm$2.89)}
& 66.08 {\scriptsize($\pm$2.75)}
& 68.05 {\scriptsize($\pm$2.14)}
& 60.06 {\scriptsize($\pm$0.99)} \\

& \multicolumn{1}{|l|}{Lookback Ratio}
& 77.65 {\scriptsize($\pm$2.17)}
& 65.14 {\scriptsize($\pm$2.17)}
& 61.50 {\scriptsize($\pm$2.97)}
& 62.05 {\scriptsize($\pm$3.15)}
& 69.03 {\scriptsize($\pm$1.98)}
& 57.75 {\scriptsize($\pm$3.44)}
& 62.57 {\scriptsize($\pm$3.10)}
& 65.88 {\scriptsize($\pm$2.42)}
& 68.36 {\scriptsize($\pm$2.31)}
& 64.03 {\scriptsize($\pm$0.97)} \\

& \multicolumn{1}{|l|}{Layer Top-$m$ Prob.}
& 66.40 {\scriptsize($\pm$2.50)}
& 64.40 {\scriptsize($\pm$2.21)}
& 59.53 {\scriptsize($\pm$3.26)}
& 49.50 {\scriptsize($\pm$3.32)}
& 60.34 {\scriptsize($\pm$2.21)}
& 57.92 {\scriptsize($\pm$3.43)}
& 63.15 {\scriptsize($\pm$3.03)}
& 62.33 {\scriptsize($\pm$2.63)}
& 70.18 {\scriptsize($\pm$2.23)}
& 60.92 {\scriptsize($\pm$1.00)} \\

& \multicolumn{1}{|l|}{Internal Variance}
& 70.36 {\scriptsize($\pm$2.39)}
& 66.72 {\scriptsize($\pm$2.15)}
& 61.88 {\scriptsize($\pm$3.37)}
& 61.20 {\scriptsize($\pm$3.28)}
& 68.77 {\scriptsize($\pm$2.07)}
& 58.49 {\scriptsize($\pm$3.29)}
& 72.69 {\scriptsize($\pm$2.84)}
& 67.70 {\scriptsize($\pm$2.59)}
& 72.64 {\scriptsize($\pm$2.42)}
& 66.26 {\scriptsize($\pm$0.99)} \\

\midrule
\multirow{6}{*}{\rotatebox{90}{Pretrained}}
& \multicolumn{1}{|l|}{Embedding (Last)}
& 52.63 {\scriptsize($\pm$2.73)}
& 57.41 {\scriptsize($\pm$2.30)}
& 60.79 {\scriptsize($\pm$3.16)}
& 62.35 {\scriptsize($\pm$3.53)}
& 58.39 {\scriptsize($\pm$2.35)}
& 52.43 {\scriptsize($\pm$3.67)}
& 64.18 {\scriptsize($\pm$2.91)}
& 57.77 {\scriptsize($\pm$2.89)}
& 64.84 {\scriptsize($\pm$2.39)}
& 59.77 {\scriptsize($\pm$1.04)} \\

& \multicolumn{1}{|l|}{Lookback Ratio}
& 71.13 {\scriptsize($\pm$2.45)}
& 63.87 {\scriptsize($\pm$2.20)}
& 70.96 {\scriptsize($\pm$3.00)}
& 58.71 {\scriptsize($\pm$2.96)}
& 65.59 {\scriptsize($\pm$2.14)}
& 55.74 {\scriptsize($\pm$3.49)}
& 70.03 {\scriptsize($\pm$2.80)}
& 68.45 {\scriptsize($\pm$2.35)}
& 73.70 {\scriptsize($\pm$2.23)}
& 65.88 {\scriptsize($\pm$0.95)} \\

& \multicolumn{1}{|l|}{Layer Top-$m$ Prob.}
& 64.18 {\scriptsize($\pm$2.54)}
& 63.92 {\scriptsize($\pm$2.23)}
& 71.85 {\scriptsize($\pm$2.90)}
& 62.70 {\scriptsize($\pm$3.08)}
& 68.40 {\scriptsize($\pm$2.09)}
& 58.23 {\scriptsize($\pm$3.31)}
& 72.48 {\scriptsize($\pm$3.02)}
& 67.81 {\scriptsize($\pm$2.57)}
& 76.68 {\scriptsize($\pm$2.15)}
& 67.76 {\scriptsize($\pm$0.96)} \\

& \multicolumn{1}{|l|}{Internal Variance}
& 65.60 {\scriptsize($\pm$2.49)}
& 66.94 {\scriptsize($\pm$2.16)}
& 68.01 {\scriptsize($\pm$3.04)}
& 62.57 {\scriptsize($\pm$3.26)}
& 70.74 {\scriptsize($\pm$2.02)}
& 58.77 {\scriptsize($\pm$3.30)}
& 73.21 {\scriptsize($\pm$3.00)}
& 69.65 {\scriptsize($\pm$2.55)}
& 79.31 {\scriptsize($\pm$1.99)}
& 68.65 {\scriptsize($\pm$0.96)} \\

& \multicolumn{1}{|l|}{Ensemble (Probe)}
& 67.34 {\scriptsize($\pm$2.47)}
& 67.75 {\scriptsize($\pm$2.13)}
& 65.81 {\scriptsize($\pm$3.32)}
& 63.46 {\scriptsize($\pm$3.31)}
& 70.92 {\scriptsize($\pm$2.00)}
& 62.30 {\scriptsize($\pm$3.17)}
& 74.24 {\scriptsize($\pm$2.92)}
& 68.80 {\scriptsize($\pm$2.58)}
& 76.68 {\scriptsize($\pm$2.34)}
& 68.75 {\scriptsize($\pm$0.98)} \\

& \multicolumn{1}{|l|}{Ensemble (Task)}
& 67.16 {\scriptsize($\pm$2.52)}
& 67.57 {\scriptsize($\pm$2.12)}
& 71.35 {\scriptsize($\pm$2.75)}
& 64.00 {\scriptsize($\pm$3.11)}
& 70.49 {\scriptsize($\pm$2.02)}
& 58.46 {\scriptsize($\pm$3.36)}
& 73.92 {\scriptsize($\pm$2.94)}
& 70.82 {\scriptsize($\pm$2.38)}
& 78.80 {\scriptsize($\pm$2.07)}
& 69.43 {\scriptsize($\pm$0.93)} \\

\midrule
\rowcolor{black!8}
\multicolumn{12}{c}{\textbf{Surrogate: Gemma-3-12B-Instruct}} \\

\multirow{4}{*}{\rotatebox{90}{Bio-ID-trained}}
& \multicolumn{1}{|l|}{Embedding (Last)}
& 75.13 {\scriptsize($\pm$2.29)}
& 61.31 {\scriptsize($\pm$2.21)}
& 58.32 {\scriptsize($\pm$3.24)}
& 59.24 {\scriptsize($\pm$3.19)}
& 62.95 {\scriptsize($\pm$2.20)}
& 57.34 {\scriptsize($\pm$3.46)}
& 64.27 {\scriptsize($\pm$3.13)}
& 66.64 {\scriptsize($\pm$2.55)}
& 70.02 {\scriptsize($\pm$2.25)}
& 62.51 {\scriptsize($\pm$1.00)} \\

& \multicolumn{1}{|l|}{Lookback Ratio}
& 75.41 {\scriptsize($\pm$2.15)}
& 57.79 {\scriptsize($\pm$2.32)}
& 52.11 {\scriptsize($\pm$3.67)}
& 55.67 {\scriptsize($\pm$3.02)}
& 61.85 {\scriptsize($\pm$2.32)}
& 49.10 {\scriptsize($\pm$3.60)}
& 52.00 {\scriptsize($\pm$3.69)}
& 58.73 {\scriptsize($\pm$2.76)}
& 50.83 {\scriptsize($\pm$2.76)}
& 54.76 {\scriptsize($\pm$1.08)} \\

& \multicolumn{1}{|l|}{Layer Top-$m$ Prob.}
& 64.25 {\scriptsize($\pm$2.49)}
& 64.66 {\scriptsize($\pm$2.25)}
& 55.01 {\scriptsize($\pm$3.51)}
& 55.81 {\scriptsize($\pm$3.15)}
& 63.97 {\scriptsize($\pm$2.22)}
& 59.48 {\scriptsize($\pm$3.37)}
& 63.79 {\scriptsize($\pm$2.98)}
& 61.63 {\scriptsize($\pm$2.80)}
& 61.60 {\scriptsize($\pm$2.54)}
& 60.74 {\scriptsize($\pm$1.02)} \\

& \multicolumn{1}{|l|}{Internal Variance}
& 69.92 {\scriptsize($\pm$2.37)}
& 64.91 {\scriptsize($\pm$2.21)}
& 62.26 {\scriptsize($\pm$3.33)}
& 58.38 {\scriptsize($\pm$3.31)}
& 69.57 {\scriptsize($\pm$2.06)}
& 54.61 {\scriptsize($\pm$3.29)}
& 65.86 {\scriptsize($\pm$2.97)}
& 63.49 {\scriptsize($\pm$2.83)}
& 70.17 {\scriptsize($\pm$2.27)}
& 63.66 {\scriptsize($\pm$1.00)} \\

\midrule
\multirow{6}{*}{\rotatebox{90}{Pretrained}}
& \multicolumn{1}{|l|}{Embedding (Last)}
& 56.85 {\scriptsize($\pm$2.51)}
& 55.24 {\scriptsize($\pm$2.34)}
& 53.55 {\scriptsize($\pm$3.44)}
& 55.15 {\scriptsize($\pm$3.39)}
& 57.70 {\scriptsize($\pm$2.28)}
& 55.29 {\scriptsize($\pm$3.47)}
& 61.28 {\scriptsize($\pm$3.25)}
& 59.55 {\scriptsize($\pm$2.49)}
& 67.94 {\scriptsize($\pm$2.21)}
& 58.21 {\scriptsize($\pm$1.03)} \\

& \multicolumn{1}{|l|}{Lookback Ratio}
& 55.83 {\scriptsize($\pm$2.60)}
& 47.61 {\scriptsize($\pm$2.37)}
& 42.83 {\scriptsize($\pm$3.58)}
& 48.28 {\scriptsize($\pm$3.00)}
& 48.43 {\scriptsize($\pm$2.24)}
& 46.46 {\scriptsize($\pm$3.46)}
& 40.30 {\scriptsize($\pm$3.16)}
& 43.27 {\scriptsize($\pm$2.69)}
& 35.21 {\scriptsize($\pm$2.61)}
& 44.05 {\scriptsize($\pm$1.03)} \\

& \multicolumn{1}{|l|}{Layer Top-$m$ Prob.}
& 68.65 {\scriptsize($\pm$2.42)}
& 65.59 {\scriptsize($\pm$2.17)}
& 62.49 {\scriptsize($\pm$3.47)}
& 59.44 {\scriptsize($\pm$3.34)}
& 70.21 {\scriptsize($\pm$2.06)}
& 55.43 {\scriptsize($\pm$3.35)}
& 66.97 {\scriptsize($\pm$2.88)}
& 64.23 {\scriptsize($\pm$2.74)}
& 69.57 {\scriptsize($\pm$2.27)}
& 64.24 {\scriptsize($\pm$1.00)} \\

& \multicolumn{1}{|l|}{Internal Variance}
& 70.23 {\scriptsize($\pm$2.36)}
& 65.50 {\scriptsize($\pm$2.16)}
& 62.43 {\scriptsize($\pm$3.37)}
& 58.90 {\scriptsize($\pm$3.27)}
& 70.03 {\scriptsize($\pm$2.07)}
& 55.47 {\scriptsize($\pm$3.30)}
& 66.59 {\scriptsize($\pm$2.95)}
& 64.91 {\scriptsize($\pm$2.77)}
& 70.14 {\scriptsize($\pm$2.27)}
& 64.25 {\scriptsize($\pm$1.00)} \\

& \multicolumn{1}{|l|}{Ensemble (Probe)}
& 68.62 {\scriptsize($\pm$2.36)}
& 64.11 {\scriptsize($\pm$2.22)}
& 62.37 {\scriptsize($\pm$3.45)}
& 59.50 {\scriptsize($\pm$3.33)}
& 70.60 {\scriptsize($\pm$2.01)}
& 53.55 {\scriptsize($\pm$3.40)}
& 66.30 {\scriptsize($\pm$2.96)}
& 64.78 {\scriptsize($\pm$2.80)}
& 70.19 {\scriptsize($\pm$2.29)}
& 63.92 {\scriptsize($\pm$1.01)} \\

& \multicolumn{1}{|l|}{Ensemble (Task)}
& 67.33 {\scriptsize($\pm$2.35)}
& 64.09 {\scriptsize($\pm$2.18)}
& 62.25 {\scriptsize($\pm$3.38)}
& 56.75 {\scriptsize($\pm$3.32)}
& 67.75 {\scriptsize($\pm$1.97)}
& 55.31 {\scriptsize($\pm$3.25)}
& 63.63 {\scriptsize($\pm$2.81)}
& 62.75 {\scriptsize($\pm$2.75)}
& 65.49 {\scriptsize($\pm$2.22)}
& 62.25 {\scriptsize($\pm$0.98)} \\

\bottomrule
\end{tabular}%
}
\caption{Surrogate-model AUROC with bootstrapped standard errors. Text and atomic claims are generated by Qwen3-8B, while probe features are extracted
post hoc using the indicated model.}
\label{tab:surr_model_qwen3_8b}
\end{table*}

\begin{table*}[t]
\centering
\setlength{\tabcolsep}{5pt}
\renewcommand{\arraystretch}{1.25}
\resizebox{\textwidth}{!}{%
\begin{tabular}{ll|ccccccccc|c}
\toprule
\multicolumn{2}{c}{\textbf{Probe}} &
\multirowcell{\textbf{Biographies}\\ \textbf{(ID)}} &
\multirowcell{\textbf{Biographies}\\ \textbf{(OOD)}} &
\textbf{Artworks} &
\textbf{Books} &
\textbf{Cities} &
\textbf{Events} &
\textbf{Inventions} &
\textbf{Landmarks} &
\textbf{Movies} &
\multirowcell{\textbf{OOD}\\\textbf{Mean}} \\
\midrule

\rowcolor{black!8}
\multicolumn{12}{c}{\textbf{Surrogate: Qwen3-4B-Instruct}} \\

\multirow{4}{*}{\rotatebox{90}{Bio-ID-trained}}
& \multicolumn{1}{|l|}{Embedding (Last)}
& 70.65 {\scriptsize($\pm$1.95)}
& 57.88 {\scriptsize($\pm$2.63)}
& 56.35 {\scriptsize($\pm$3.22)}
& 71.45 {\scriptsize($\pm$3.65)}
& 55.23 {\scriptsize($\pm$2.75)}
& 51.53 {\scriptsize($\pm$5.10)}
& 69.13 {\scriptsize($\pm$3.28)}
& 59.39 {\scriptsize($\pm$3.60)}
& 62.36 {\scriptsize($\pm$3.62)}
& 60.41 {\scriptsize($\pm$1.26)} \\

& \multicolumn{1}{|l|}{Lookback Ratio}
& 76.09 {\scriptsize($\pm$1.88)}
& 53.64 {\scriptsize($\pm$2.58)}
& 59.48 {\scriptsize($\pm$3.03)}
& 49.65 {\scriptsize($\pm$5.15)}
& 56.58 {\scriptsize($\pm$2.73)}
& 56.61 {\scriptsize($\pm$4.79)}
& 67.66 {\scriptsize($\pm$3.39)}
& 54.17 {\scriptsize($\pm$3.72)}
& 56.53 {\scriptsize($\pm$4.19)}
& 56.79 {\scriptsize($\pm$1.34)} \\

& \multicolumn{1}{|l|}{Layer Top-$m$ Prob.}
& 62.73 {\scriptsize($\pm$2.19)}
& 58.30 {\scriptsize($\pm$2.51)}
& 47.35 {\scriptsize($\pm$3.16)}
& 54.58 {\scriptsize($\pm$5.00)}
& 50.51 {\scriptsize($\pm$2.82)}
& 54.09 {\scriptsize($\pm$4.53)}
& 58.27 {\scriptsize($\pm$3.51)}
& 54.22 {\scriptsize($\pm$3.90)}
& 52.84 {\scriptsize($\pm$3.83)}
& 53.77 {\scriptsize($\pm$1.32)} \\

& \multicolumn{1}{|l|}{Internal Variance}
& 70.21 {\scriptsize($\pm$2.07)}
& 62.53 {\scriptsize($\pm$2.70)}
& 56.19 {\scriptsize($\pm$3.34)}
& 56.87 {\scriptsize($\pm$5.18)}
& 61.60 {\scriptsize($\pm$2.61)}
& 60.01 {\scriptsize($\pm$4.45)}
& 74.69 {\scriptsize($\pm$3.27)}
& 58.52 {\scriptsize($\pm$3.57)}
& 67.91 {\scriptsize($\pm$3.46)}
& 62.29 {\scriptsize($\pm$1.29)} \\

\midrule
\multirow{6}{*}{\rotatebox{90}{Pretrained}}
& \multicolumn{1}{|l|}{Embedding (Last)}
& 53.55 {\scriptsize($\pm$2.32)}
& 52.93 {\scriptsize($\pm$2.76)}
& 58.21 {\scriptsize($\pm$3.00)}
& 59.29 {\scriptsize($\pm$5.60)}
& 60.15 {\scriptsize($\pm$2.89)}
& 53.91 {\scriptsize($\pm$4.35)}
& 56.28 {\scriptsize($\pm$3.91)}
& 59.53 {\scriptsize($\pm$3.32)}
& 58.61 {\scriptsize($\pm$3.64)}
& 57.36 {\scriptsize($\pm$1.34)} \\

& \multicolumn{1}{|l|}{Lookback Ratio}
& 65.37 {\scriptsize($\pm$2.18)}
& 55.02 {\scriptsize($\pm$2.61)}
& 63.03 {\scriptsize($\pm$3.02)}
& 62.14 {\scriptsize($\pm$5.84)}
& 61.92 {\scriptsize($\pm$2.64)}
& 61.07 {\scriptsize($\pm$5.13)}
& 63.39 {\scriptsize($\pm$3.42)}
& 64.98 {\scriptsize($\pm$3.49)}
& 58.13 {\scriptsize($\pm$4.32)}
& 61.21 {\scriptsize($\pm$1.40)} \\

& \multicolumn{1}{|l|}{Layer Top-$m$ Prob.}
& 63.30 {\scriptsize($\pm$2.22)}
& 58.12 {\scriptsize($\pm$2.59)}
& 55.15 {\scriptsize($\pm$3.33)}
& 57.93 {\scriptsize($\pm$5.34)}
& 60.78 {\scriptsize($\pm$2.62)}
& 60.93 {\scriptsize($\pm$4.29)}
& 75.96 {\scriptsize($\pm$2.83)}
& 60.89 {\scriptsize($\pm$3.48)}
& 63.33 {\scriptsize($\pm$4.15)}
& 61.64 {\scriptsize($\pm$1.30)} \\

& \multicolumn{1}{|l|}{Internal Variance}
& 63.38 {\scriptsize($\pm$2.20)}
& 60.00 {\scriptsize($\pm$2.57)}
& 57.56 {\scriptsize($\pm$3.20)}
& 60.17 {\scriptsize($\pm$5.92)}
& 63.90 {\scriptsize($\pm$2.58)}
& 62.00 {\scriptsize($\pm$4.42)}
& 75.74 {\scriptsize($\pm$3.12)}
& 59.98 {\scriptsize($\pm$3.62)}
& 63.64 {\scriptsize($\pm$4.22)}
& 62.87 {\scriptsize($\pm$1.36)} \\

& \multicolumn{1}{|l|}{Ensemble (Probe)}
& 66.04 {\scriptsize($\pm$2.19)}
& 60.83 {\scriptsize($\pm$2.70)}
& 57.04 {\scriptsize($\pm$3.37)}
& 54.20 {\scriptsize($\pm$5.47)}
& 63.76 {\scriptsize($\pm$2.59)}
& 58.64 {\scriptsize($\pm$4.49)}
& 74.61 {\scriptsize($\pm$3.39)}
& 60.95 {\scriptsize($\pm$3.56)}
& 66.27 {\scriptsize($\pm$3.90)}
& 62.04 {\scriptsize($\pm$1.34)} \\

& \multicolumn{1}{|l|}{Ensemble (Task)}
& 64.23 {\scriptsize($\pm$2.21)}
& 59.93 {\scriptsize($\pm$2.61)}
& 57.95 {\scriptsize($\pm$3.12)}
& 60.41 {\scriptsize($\pm$5.13)}
& 63.34 {\scriptsize($\pm$2.56)}
& 61.94 {\scriptsize($\pm$4.43)}
& 76.03 {\scriptsize($\pm$2.96)}
& 60.78 {\scriptsize($\pm$3.52)}
& 63.07 {\scriptsize($\pm$4.38)}
& 62.93 {\scriptsize($\pm$1.31)} \\

\midrule
\rowcolor{black!8}
\multicolumn{12}{c}{\textbf{Surrogate: Qwen3-8B}} \\

\multirow{4}{*}{\rotatebox{90}{Bio-ID-trained}}
& \multicolumn{1}{|l|}{Embedding (Last)}
& 70.02 {\scriptsize($\pm$1.96)}
& 58.20 {\scriptsize($\pm$2.55)}
& 51.97 {\scriptsize($\pm$3.15)}
& 62.95 {\scriptsize($\pm$3.76)}
& 47.96 {\scriptsize($\pm$2.86)}
& 58.96 {\scriptsize($\pm$4.60)}
& 64.14 {\scriptsize($\pm$3.10)}
& 57.87 {\scriptsize($\pm$3.29)}
& 50.67 {\scriptsize($\pm$4.05)}
& 56.59 {\scriptsize($\pm$1.23)} \\

& \multicolumn{1}{|l|}{Lookback Ratio}
& 73.02 {\scriptsize($\pm$2.03)}
& 52.29 {\scriptsize($\pm$2.63)}
& 54.91 {\scriptsize($\pm$3.34)}
& 44.80 {\scriptsize($\pm$4.73)}
& 60.68 {\scriptsize($\pm$2.79)}
& 53.91 {\scriptsize($\pm$5.08)}
& 57.54 {\scriptsize($\pm$3.40)}
& 51.76 {\scriptsize($\pm$3.59)}
& 57.40 {\scriptsize($\pm$3.82)}
& 54.16 {\scriptsize($\pm$1.33)} \\

& \multicolumn{1}{|l|}{Layer Top-$m$ Prob.}
& 56.63 {\scriptsize($\pm$2.28)}
& 53.98 {\scriptsize($\pm$2.59)}
& 51.73 {\scriptsize($\pm$3.49)}
& 49.85 {\scriptsize($\pm$6.16)}
& 56.34 {\scriptsize($\pm$2.83)}
& 54.91 {\scriptsize($\pm$4.65)}
& 57.15 {\scriptsize($\pm$3.56)}
& 48.80 {\scriptsize($\pm$3.81)}
& 52.04 {\scriptsize($\pm$4.00)}
& 53.10 {\scriptsize($\pm$1.42)} \\

& \multicolumn{1}{|l|}{Internal Variance}
& 71.08 {\scriptsize($\pm$2.07)}
& 61.75 {\scriptsize($\pm$2.84)}
& 56.61 {\scriptsize($\pm$3.36)}
& 54.57 {\scriptsize($\pm$5.24)}
& 59.07 {\scriptsize($\pm$2.72)}
& 59.26 {\scriptsize($\pm$4.52)}
& 70.83 {\scriptsize($\pm$3.47)}
& 63.08 {\scriptsize($\pm$3.34)}
& 68.34 {\scriptsize($\pm$3.74)}
& 61.69 {\scriptsize($\pm$1.32)} \\

\midrule
\multirow{6}{*}{\rotatebox{90}{Pretrained}}
& \multicolumn{1}{|l|}{Embedding (Last)}
& 49.35 {\scriptsize($\pm$2.44)}
& 57.13 {\scriptsize($\pm$2.47)}
& 47.42 {\scriptsize($\pm$2.98)}
& 53.61 {\scriptsize($\pm$4.76)}
& 56.63 {\scriptsize($\pm$2.83)}
& 63.10 {\scriptsize($\pm$4.78)}
& 57.60 {\scriptsize($\pm$3.64)}
& 55.46 {\scriptsize($\pm$3.56)}
& 55.96 {\scriptsize($\pm$4.32)}
& 55.86 {\scriptsize($\pm$1.33)} \\

& \multicolumn{1}{|l|}{Lookback Ratio}
& 60.62 {\scriptsize($\pm$2.30)}
& 56.31 {\scriptsize($\pm$2.62)}
& 61.71 {\scriptsize($\pm$2.93)}
& 57.91 {\scriptsize($\pm$5.86)}
& 57.56 {\scriptsize($\pm$2.87)}
& 54.48 {\scriptsize($\pm$5.21)}
& 51.98 {\scriptsize($\pm$3.63)}
& 58.69 {\scriptsize($\pm$3.59)}
& 58.90 {\scriptsize($\pm$4.25)}
& 57.19 {\scriptsize($\pm$1.42)} \\

& \multicolumn{1}{|l|}{Layer Top-$m$ Prob.}
& 59.74 {\scriptsize($\pm$2.24)}
& 60.58 {\scriptsize($\pm$2.49)}
& 58.28 {\scriptsize($\pm$3.29)}
& 58.83 {\scriptsize($\pm$4.59)}
& 61.96 {\scriptsize($\pm$2.56)}
& 54.90 {\scriptsize($\pm$5.23)}
& 77.27 {\scriptsize($\pm$2.86)}
& 65.90 {\scriptsize($\pm$3.28)}
& 66.17 {\scriptsize($\pm$4.02)}
& 62.99 {\scriptsize($\pm$1.29)} \\

& \multicolumn{1}{|l|}{Internal Variance}
& 65.96 {\scriptsize($\pm$2.15)}
& 59.25 {\scriptsize($\pm$2.67)}
& 58.05 {\scriptsize($\pm$3.35)}
& 55.34 {\scriptsize($\pm$5.37)}
& 61.29 {\scriptsize($\pm$2.64)}
& 57.60 {\scriptsize($\pm$4.65)}
& 71.22 {\scriptsize($\pm$3.35)}
& 63.26 {\scriptsize($\pm$3.53)}
& 67.54 {\scriptsize($\pm$3.70)}
& 61.69 {\scriptsize($\pm$1.33)} \\

& \multicolumn{1}{|l|}{Ensemble (Probe)}
& 70.37 {\scriptsize($\pm$2.11)}
& 61.54 {\scriptsize($\pm$2.88)}
& 56.95 {\scriptsize($\pm$3.40)}
& 53.39 {\scriptsize($\pm$4.88)}
& 60.00 {\scriptsize($\pm$2.71)}
& 57.22 {\scriptsize($\pm$4.49)}
& 71.04 {\scriptsize($\pm$3.61)}
& 63.78 {\scriptsize($\pm$3.33)}
& 67.10 {\scriptsize($\pm$3.53)}
& 61.38 {\scriptsize($\pm$1.30)} \\

& \multicolumn{1}{|l|}{Ensemble (Task)}
& 63.29 {\scriptsize($\pm$2.22)}
& 62.04 {\scriptsize($\pm$2.54)}
& 60.85 {\scriptsize($\pm$3.15)}
& 58.33 {\scriptsize($\pm$5.17)}
& 62.89 {\scriptsize($\pm$2.61)}
& 54.00 {\scriptsize($\pm$4.91)}
& 70.72 {\scriptsize($\pm$3.47)}
& 64.71 {\scriptsize($\pm$3.48)}
& 66.66 {\scriptsize($\pm$4.25)}
& 62.52 {\scriptsize($\pm$1.35)} \\

\bottomrule
\end{tabular}%
}
\caption{Surrogate-model AUROC with bootstrapped standard errors. Text and atomic claims are generated by Gemma-3-12B-Instruct, while probe features are extracted post hoc using the indicated surrogate model.}
\label{tab:surr_model_gemma_12b}
\end{table*}

\paragraph{Surrogate-Model Experiment: The Generator and Detector Can Be Decoupled.} \label{par:surrogate}
The surrogate-model experiments in \Cref{tab:surr_model_qwen3_8b,tab:surr_model_gemma_12b} show that features extracted post hoc from a separate model can remain competitive with generator-native features, and occasionally perform better. However, performance varies across surrogate--feature combinations, indicating that surrogate compatibility matters. These results suggest that probe-based factuality estimation can extend to black-box generators using an accessible white-box surrogate, though robustness of the surrogate and generator-surrogate compatibility remain future work.

\begin{table*}[ht]
\renewcommand\arraystretch{1.2}
\centering
\resizebox{\linewidth}{!}{%
\begin{tabular}{p{4cm}|p{12cm}}
\toprule
\multicolumn{1}{c}{\textbf{Query}} & \multicolumn{1}{c}{\textbf{Response}} \\
\midrule
\multicolumn{2}{c}{\cellcolor{black!8}\textbf{Qwen3-8B}} \\
\midrule
\textbf{Biographies (ID)}\par Tell me a paragraph bio of Gemma Arterton. & \ldots{} she gained international recognition for her role as the Bond girl in \correctclaim{Quantum of Solace (2008)}\nobreak\,{\bfseries\boldmath[\textcolor{predred}{\ensuremath{\mathrm{S}: 0.01}}, \textcolor{predgreen}{\ensuremath{\mathrm{L}: 0.75}}, \textcolor{predgreen}{\ensuremath{\mathcal{E}: 0.75}}]}, which launched her career in Hollywood. Over the years, she has appeared in a variety of films, including \hallucinatedclaim{The Golden Compass (2007)}\nobreak\,{\bfseries\boldmath[\textcolor{predred}{\ensuremath{\mathrm{S}: 0.01}}, \textcolor{predred}{\ensuremath{\mathrm{L}: 0.06}}, \textcolor{predred}{\ensuremath{\mathcal{E}: 0.36}}]}, \hallucinatedclaim{Tinker Tailor Soldier Spy (2011)}\nobreak\,{\bfseries\boldmath[\textcolor{predgreen}{\ensuremath{\mathrm{S}: 0.95}}, \textcolor{predred}{\ensuremath{\mathrm{L}: 0.22}}, \textcolor{predred}{\ensuremath{\mathcal{E}: 0.33}}]}, \ldots{} \\
\midrule
\textbf{Inventions}\par Tell me about the invention of WiFi and its inventor. & \ldots{} WiFi technology is based on \correctclaim{the IEEE 802.11 standards, which were developed by the Institute of Electrical and Electronics Engineers (IEEE)}\nobreak\,{\bfseries\boldmath[\textcolor{predred}{\ensuremath{\mathrm{S}: 0.04}}, \textcolor{predgreen}{\ensuremath{\mathrm{L}: 0.59}}, \textcolor{predgreen}{\ensuremath{\mathcal{E}: 0.57}}]} \ldots \par However, \hallucinatedclaim{the inventor most commonly associated with WiFi is Victor W. Grantham}\nobreak\,{\bfseries\boldmath[\textcolor{predred}{\ensuremath{\mathrm{S}: 0.30}}, \textcolor{predred}{\ensuremath{\mathrm{L}: 0.00}}, \textcolor{predred}{\ensuremath{\mathcal{E}: 0.09}}]}\hallucinatedclaim{, a researcher at Hewlett-Packard (HP)}\nobreak\,{\bfseries\boldmath[\textcolor{predgreen}{\ensuremath{\mathrm{S}: 1.00}}, \textcolor{predred}{\ensuremath{\mathrm{L}: 0.00}}, \textcolor{predred}{\ensuremath{\mathcal{E}: 0.04}}]}. \ldots{} \\
\midrule
\textbf{Inventions (Failure)}\par Tell me about the invention of Solar Power and its inventor. & \ldots{} \hallucinatedclaim{Augustin Mouchot (1828–1889) built the first solar-powered engine in 1860}\nobreak\,{\bfseries\boldmath[\textcolor{predred}{\ensuremath{\mathrm{S}: 0.07}}, \textcolor{predred}{\ensuremath{\mathrm{L}: 0.20}}, \textcolor{predred}{\ensuremath{\mathcal{E}: 0.31}}]}, which was used to pump water in Algeria.\par  The Birth of Modern Solar Power\par The breakthrough in modern solar power came \correctclaim{in the 1950s with the invention of the photovoltaic cell}\nobreak\,{\bfseries\boldmath[\textcolor{predred}{\ensuremath{\mathrm{S}: 0.13}}, \textcolor{predred}{\ensuremath{\mathrm{L}: 0.36}}, \textcolor{predred}{\ensuremath{\mathcal{E}: 0.30}}]}, which converts sunlight directly into electricity.\par - Inventors: \correctclaim{Daryl Chapin, Calvin Fuller, and Jerry Pearson at Bell Labs (now part of Lucent Technologies) developed the first practical silicon solar cell in 1954}\nobreak\,{\bfseries\boldmath[\textcolor{predred}{\ensuremath{\mathrm{S}: 0.21}}, \textcolor{predred}{\ensuremath{\mathrm{L}: 0.39}}, \textcolor{predred}{\ensuremath{\mathcal{E}: 0.26}}]}. \ldots{} \\
\midrule
\multicolumn{2}{c}{\cellcolor{black!8}\textbf{Gemma-3-12B-IT}} \\
\midrule
\textbf{Biographies (ID)}\par Tell me a paragraph bio of Radhika Apte. & \correctclaim{Radhika Apte is a highly acclaimed Indian actress}\nobreak\,{\bfseries\boldmath[\textcolor{predgreen}{\ensuremath{\mathrm{S}: 1.00}}, \textcolor{predgreen}{\ensuremath{\mathrm{L}: 0.75}}, \textcolor{predred}{\ensuremath{\mathcal{E}: 0.43}}]} known for her versatility and captivating performances across film, television, and theatre. \correctclaim{Beginning her career in theatre}\nobreak\,{\bfseries\boldmath[\textcolor{predred}{\ensuremath{\mathrm{S}: 0.14}}, \textcolor{predgreen}{\ensuremath{\mathrm{L}: 0.58}}, \textcolor{predgreen}{\ensuremath{\mathcal{E}: 0.56}}]}, she transitioned to Bollywood and quickly gained recognition for her roles in \hallucinatedclaim{films like "Vaastav: The Reality,"}\nobreak\,{\bfseries\boldmath[\textcolor{predgreen}{\ensuremath{\mathrm{S}: 0.99}}, \textcolor{predred}{\ensuremath{\mathrm{L}: 0.05}}, \textcolor{predred}{\ensuremath{\mathcal{E}: 0.14}}]} \ldots{} \\
\midrule
\textbf{Artworks}\par Tell me about the artwork "Still Life with Apples" by Paul Cézanne and its artist. & \ldots{} \correctclaim{The Artwork: "Still Life with Apples" (c. 1893)}\nobreak\,{\bfseries\boldmath[\textcolor{predred}{\ensuremath{\mathrm{S}: 0.46}}, \textcolor{predgreen}{\ensuremath{\mathrm{L}: 0.66}}, \textcolor{predgreen}{\ensuremath{\mathcal{E}: 0.63}}]} Description: \ldots \par \correctclaim{Medium: Oil on canvas.}\nobreak\,{\bfseries\boldmath[\textcolor{predred}{\ensuremath{\mathrm{S}: 0.15}}, \textcolor{predgreen}{\ensuremath{\mathrm{L}: 0.73}}, \textcolor{predgreen}{\ensuremath{\mathcal{E}: 0.66}}]} \par \hallucinatedclaim{Size: Approximately 38 cm × 51 cm (15 in × 20 in)} \par {\bfseries\boldmath[\textcolor{predgreen}{\ensuremath{\mathrm{S}: 0.70}}, \textcolor{predred}{\ensuremath{\mathrm{L}: 0.07}}, \textcolor{predred}{\ensuremath{\mathcal{E}: 0.29}}]}  \ldots{} \\
\midrule
\textbf{Landmarks (Failure)}\par Tell me about the landmark The Giza Pyramid Complex, Egypt. & \ldots{}  It's a vast complex of structures \correctclaim{built during the Fourth Dynasty of the Old Kingdom}\nobreak\,{\bfseries\boldmath[\textcolor{predred}{\ensuremath{\mathrm{S}: 0.36}}, \textcolor{predred}{\ensuremath{\mathrm{L}: 0.12}}, \textcolor{predred}{\ensuremath{\mathcal{E}: 0.25}}]} (\hallucinatedclaim{roughly 2589–2566 BC})\nobreak\,{\bfseries\boldmath[\textcolor{predred}{\ensuremath{\mathrm{S}: 0.42}}, \textcolor{predgreen}{\ensuremath{\mathrm{L}: 0.64}}, \textcolor{predgreen}{\ensuremath{\mathcal{E}: 0.64}}]}. \correctclaim{It served as a necropolis (city of the dead) for the pharaohs and their consorts}\nobreak\,{\bfseries\boldmath[\textcolor{predred}{\ensuremath{\mathrm{S}: 0.00}}, \textcolor{predred}{\ensuremath{\mathrm{L}: 0.16}}, \textcolor{predred}{\ensuremath{\mathcal{E}: 0.44}}]} \ldots{} \\
\bottomrule
\end{tabular}%
}
\caption{Qualitative claim-level predictions. Green and red spans are correct and hallucinated claims, respectively. After each span, $\mathrm{S}$, $L$, and $\mathcal{E}$ give $P(y=1)$ from the supervised embedding baseline, Layer Top-$m$, and Ensemble (Task). Green probability text denotes a prediction $\geq 0.5$; red denotes a prediction $<0.5$.}
\label{tab:qualitative_examples}
\end{table*}

\end{document}